\documentclass[12pt, TexShade, letterpaper]{report}
\usepackage[utf8]{inputenc}
\usepackage{palatino}
\usepackage{amsmath}
\usepackage{amssymb}
\usepackage{graphicx}
\usepackage[labelfont=bf]{caption}
\usepackage{subcaption}
\usepackage{setspace}
\usepackage[margin=1in, headsep=1cm, bottom=5cm]{geometry}
\usepackage{tabu}
\usepackage{nomencl}
\usepackage[nonumberlist,nogroupskip,xindy]{glossaries}
\usepackage{floatrow}
\usepackage{wrapfig}
\usepackage{fancyhdr}
\usepackage{lmodern}
\usepackage{nomencl}

\usepackage{todonotes}
\usepackage{lipsum}
\usepackage{natbib}
\usepackage{algorithm}
\usepackage[noend]{algpseudocode}
\usepackage{booktabs}

\usepackage[a-3u]{pdfx}

\hypersetup{
	hidelinks = true,
}

\makenomenclature

\renewcommand{\chaptermark}[1]{\markboth{#1}{}} 

\setglossarystyle{long}

\newglossarystyle{mystyle}{%
	{\begin{longtabu} to \linewidth {p{0.15\linewidth}p{0.85\linewidth}}}%
		{\end{longtabu}}%
}

\fancypagestyle{plain}{%
	\fancyhf{} 
	\fancyhead[R]{\textbf{\thepage}} 
}

\usepackage{xpatch}
\xpretocmd\headrule{\color{red}}{}{\PatchFailed}

\emergencystretch=\maxdimen
\newacronym{av}{AV}{autonomous vehicle}
\newacronym{nn}{NN}{neural network}
\newacronym{tsp}{TSP}{Travelling Salesman problem}
\newacronym{vrp}{VRP}{Vehicle Routing Problem}
\newacronym{cvrp}{CVRP}{Capacitated Vehicle Routing Problem}
\newacronym{rl}{RL}{reinforcement learning}
\newacronym{gnn}{GNN}{Graph Neural Net}
\newacronym{gat}{GAT}{Graph Attention Net}
\newacronym{rnn}{RNN}{Recurrent Neural Net}
\newacronym{drl}{DRL}{Deep Reinforcement Learning}
\newacronym{mlp}{MLP}{Multi-Layer Perceptron}

\newacronym{co}{CO}{Combinatorial Optimization}
\newacronym{ndp}{TNDP}{Transit Network Design Problem}
\newacronym{fsp}{FSP}{Frequency-Setting Problem}
\newacronym{dfsp}{DFSP}{Design and Frequency-Setting Problem}
\newacronym{sp}{SP}{Scheduling Problem}
\newacronym{tp}{TP}{Timetabling Problem}
\newacronym{ndsp}{NDSP}{Network Design and Scheduling Problem}

\newacronym{mod}{MoD}{Mobility on Demand}
\newacronym{amod}{AMoD}{Autonomous Mobility on Demand}
\newacronym{imodp}{IMoDP}{Intermodal Mobility-on-Demand Problem}
\newacronym{matsim}{MATSim}{Multi-Agent Transport Simulation}
\newacronym{od}{OD}{Origin-Destination}
\newacronym{csa}{CSA}{Connection Scan Algorithm}
\newacronym{mdp}{MDP}{Markov Decision Process}
\newacronym{dqn}{DQN}{Deep Q-Networks}
\newacronym{acer}{ACER}{Actor-Critic with Experience Replay}
\newacronym{ppo}{PPO}{Proximal Policy Optimization}
\newacronym{artm}{ARTM}{Metropolitan Regional Transportation Authority}
\newacronym{stl}{STL}{Soci\'et\'e de Transport de Laval}
\newacronym{cda}{CDA}{Census Dissemination Area}

\makeindex
\makeglossaries

\renewcommand{\thepage}{\roman{page}}

\author{\textcopyright Andrew Holliday, January, 2024}
\date{}

\AtBeginDocument{}%

\begin{document}

\begin{titlepage}
		\begin{center}
			\vspace*{0.5cm}

			\LARGE
			\textbf{Applications of deep reinforcement learning to urban transit network design}
			
			\vspace{1cm}
			
			\textit{Andrew Holliday}
			
			\vspace{1.2cm}
			
			\includegraphics[width=0.25\textwidth]{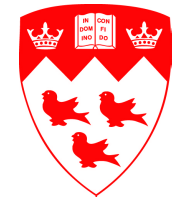}
			
			\Large
			School of Computer Science
			
			\vspace{-5mm}
			McGill University
			
			\vspace{-5mm}
			Montr\'eal, Qu\'ebec, Canada
			
			\vspace{5mm}
			\today
			\small
			\vspace{0.5cm}
			{\color{red} \hrule height 0.75mm}
			
			\vspace{0.2cm}
			
			A thesis presented for the degree of Doctor of Philosophy in Computer Science

			\copyright\hspace{0.5mm}2024 Andrew Holliday
			
		\end{center}
	\end{titlepage}
\setlength{\voffset}{2cm}
\renewcommand{\chaptermark}[1]{%
	\markboth{\thechapter.\ #1}{}}
\chapter*{Abstract}\markboth{Abstract}{}
	\label{chap:engAbstract}
	\addcontentsline{toc}{section}{\nameref{chap:engAbstract}}

    This thesis concerns the use of reinforcement learning to train neural networks to aid in the design of public transit networks.  The Transit Network Design Problem (TNDP) is an optimization problem of considerable practical importance.  Given a city with an existing road network and travel demands, the goal is to find a set of transit routes - each of which is a path through the graph - that collectively satisfy all demands, while minimizing a cost function that may depend both on passenger satisfaction and operating costs.
    
    The existing literature on this problem mainly considers metaheuristic optimization algorithms, such as genetic algorithms and ant-colony optimization.  By contrast, we begin by taking a reinforcement learning approach, formulating the construction of a set of transit routes as a Markov Decision Process (MDP) and training a neural net policy to act as the agent in this MDP.  We then show that, beyond using this policy to plan a transit network directly, it can be combined with existing metaheuristic algorithms, both to initialize the solution and to suggest promising moves at each step of a search through solution space.
    
 	We find that such hybrid algorithms, which use a neural policy trained via reinforcement learning as a core component within a classical metaheuristic framework, can plan transit networks that are superior to those planned by either the neural policy or the metaheuristic algorithm.  We demonstrate the utility of our approach by using it to redesign the transit network for the city of Laval, Quebec, and show that in simulation, the resulting transit network provides better service at lower cost than the existing transit network.

\chapter*{Abrégé}\markboth{Abrégé}{}
	\label{chap:frAbstract}
	\addcontentsline{toc}{section}{\nameref{chap:frAbstract}}
    Cette thèse concerne l'utilisation de l'apprentissage par renforcement pour entraîner des réseaux de neurones afin d'aider à la conception des réseaux de transport en commun. Le problème de conception des réseaux de transport en commun (TNDP) est un problème d'optimisation d'une importance pratique considérable. Étant donné une ville avec un réseau routier existant et des demandes de déplacement, l'objectif est de trouver un ensemble de lignes de transport en commun - chacune étant un chemin à travers le graphe - qui satisfont collectivement toutes les demandes, tout en minimisant une fonction de coût qui peut dépendre à la fois de la satisfaction des passagers et des coûts d'exploitation.
    
    La littérature existante sur ce problème considère principalement des algorithmes issus de la littérature sur l'optimisation par métaheuristiques, tels que les algorithmes génétiques et l'optimisation par colonies de fourmis. En revanche, nous commençons par adopter une approche d'apprentissage par renforcement, en formulant le TNDP comme un processus de décision markovien et en entraînant une politique de réseau de neurones avec une architecture novatrice pour assembler des réseaux de transport en commun. Nous montrons ensuite qu'au-delà de l'utilisation de cette politique pour planifier directement un réseau de transport en commun, elle peut être combinée avec des algorithmes métaheuristiques existants, à la fois pour l'initialisation et pour suggérer des mouvements prometteurs à chaque étape d'une recherche dans l'espace des solutions.
    
    Nous constatons que de tels algorithmes hybrides, qui utilisent une politique neuronale entraînée via l'apprentissage par renforcement comme composant central dans un cadre métaheuristique classique, peuvent planifier des réseaux de transport en commun supérieurs à ceux planifiés soit par la politique neuronale, soit par l'algorithme métaheuristique. Nous démontrons l'utilité de notre approche en l'utilisant pour redessiner le réseau de transport en commun de la ville de Laval, Québec, et montrons qu'en simulation, le réseau de transport en commun résultant offre un meilleur service à moindre coût que le réseau de transport en commun existant.
    
\chapter*{Acknowledgements}\markboth{Acknowledgements}{}
	\label{chap:acknowledgments}
	\addcontentsline{toc}{section}{\nameref{chap:acknowledgments}}
	
	The journey to the completion of this thesis was long and uncertain, and it would not have been possible without help from many quarters.  Credit goes first to my two supervisors, professors Gregory Dudek and Ahmed El-Geneidy.  With this thesis, both of them took a chance: Greg in accepting my proposal of a topic well outside of both of our expertise, and Ahmed, in taking me on as a student from far outside his own field of urban planning.  I am grateful to both of them for taking a chance on me, and I hope I have convinced them both that it was the right call.

	Professor Dudek guided me through the mists and mazes of graduate school and the ups and downs of the publishing cycle.  His boundless enthusiasm for his field has been a source of inspiration, as has his talent for bringing people together.  Because of Greg I have been privileged to meet many other fine teachers and colleagues who have made this journey not only possible but enjoyable.
	
	The data and contacts in the real world of transit planning that professor El-Geneidy provided were essential to this work.  Just as important was his guidance to me as a novice in his field, and his patience and good humour in delivering that guidance.
	
	My many colleagues at the Mobile Robotics Laboratory over the years gave me plenty of advice and encouragement, as well as the simple pleasure of their company.  Our lab is unique, not just in its large number of students, but also in the sense of community and camaraderie among them.  Belonging to this group of people has been a great privilege.
		
	Anqi Xu, Florian Shkurti, Jimmy Li, Sandeep Manjanna, Malika Meghjani, and Travis Manderson were senior students in the lab when I was just starting out. Their brilliance and friendliness made the lab a very welcoming place, and they were all inspiring examples of what an academic could be.  As a new student, I was fortunate to work with and learn from them on numerous robotics projects and trials out in the field.  These trips could be chaotic, but they were always great adventures. 

	I want to thank professor David Meger for his help and advice, and for being so tolerant of my constant second-guessing.  Arnold Kalmbach deserves special thanks for giving me a spare room to live in for a crucial few months, and for being such a good roommate.  Nikhil Kakodkar has been a good friend to me from the beginning: a teammate in some challenging classes, a sounding board for research ideas and a link to a broader social world.  I owe him thanks not just for his friendship but for the many other friends in Montr\'eal that I've met because of him.
	
	Many other colleagues past and present helped make MRL such a great place to work; this includes Lucas Berry, Wilfred Mason, Wei-Di Chang, Scott Fujimoto and Edward Smith, Julie Al-Hosh, Jean-Francois Tremblay, Johanna Hansen, Chelsea Taylor, Juan Camilo Gamboa Higuera, Amin Abyaneh, Stanley Wu, Hanna Yurchyk, Khalil Virji, Faraz Lotfi and Farnoosh Faraji, Charlotte Morissette, Mariana Sosa, Karim Koreitem, Xiru Zhu, professor Hsiu-Chin Lin, and doubtless others that I've forgotten.  I am grateful to have spent this time with you all.
	
	My parents have been unwavering in their encouragement and support of me throughout my academic journey.  Beyond that, they deserve the credit for all of the opportunities they worked so hard to provide for me, all of which brought me to undertake this degree.  For that I am endlessly grateful, and I can only hope that I have justified their pride in me.
	
	I also give thanks to the reviewers of all the paper submissions I've made over the past seven years.  Their feedback, both favourable and not, has provided valuable lessons and an impetus to improve my craft.
	
	Lastly, I want to thank all of the bus drivers and train operators who've ever helped get me where I've been going.  Their work was and is an inspiration; without it, this thesis would never have existed.
	
	\tableofcontents\thispagestyle{plain}

	\listoffigures\thispagestyle{plain}
	\listoftables
	\glsaddall
	\setlength\LTleft{0pt}
	\setlength\LTright{0pt}
	\setlength\glsdescwidth{0.8\hsize}
	\printglossary[title={List of Acronyms}]
	\markright{List of Acronyms}

	
	\printnomenclature 

 	\clearpage
	\pagenumbering{arabic} 
	
	\glsresetall
\chapter{Introduction}\label{chap:intro}

\nomenclature{$i$}{A generic indexing variable.  Used variously to denote a node in a graph, an element in a sequence of integers, or in other ways as needed.}
\nomenclature{$j$}{A generic indexing variable (see entry for $i$).}
\nomenclature{$k$}{A generic indexing variable (see entry for $i$).}
\nomenclature{$l$}{A generic indexing variable (see entry for $i$).}


The \gls{ndp} is the problem of designing a network of public transit routes - such as metro lines and bus routes - for a city, so as to satisfy some objectives, such as meeting all travel demand and minimizing operating costs.  It is an important problem in urban planning, as public transit is essential for overcoming the inefficiencies inherent in car-based urban transportation.  Compared with point-to-point car transit, mass transit can significantly reduce traffic congestion, energy consumption, and emissions~\citep{roughgarden2002bad, oh2020evaluating, rich2023fixed}.  The spatial layout of a transit network can have a major impact on its ability to realize these benefits.  One city planner~\citep{spielerInterview2018} gave the city of Seattle, Washington as a good example: it has recently expanded its transit network in a way that takes into account the interactions between routes, and the city's transit ridership has consequently grown.  The same planner gave the city of Denver, Colorado as a bad example, where much transit has been built that did not reach places people want to go.  In each case, the spatial layout of the transit routes was the major reason for the transit system's quality, or lack thereof.

The \gls{ndp}'s importance has arguably grown in recent years, as the COVID-19 pandemic resulted in declines in transit ridership in cities worldwide~\citep{liu2020impacts}.  This has led to a financial crisis for many municipal transit agencies, putting them under pressure to reduce transit operating costs~\citep{kar2022public}.  Many agencies thus find themselves having to do more with less.  Improving the design of a transit network, such as by changing bus routes, can improve the quality of service an agency can provide while reducing operating costs.  

But the \gls{ndp} is very challenging problem.  It is NP-hard, and has commonalities with the \gls{tsp} and \gls{vrp}, but is much more complex than these, due to the need to plan multiple routes that interact non-linearly due to passengers' ability to make transfers between routes.  Since real-world cities typically have hundreds or even thousands of possible stop locations, analytical optimization approaches are infeasible.  Real-world transit networks are still commonly designed by hand~\citep{duran2022survey}, but there exists a substantial literature on computational approaches to the \gls{ndp}.  

The most successful approaches to-date have been metaheuristic algorithms, such as evolutionary algorithms, simulated annealing, and ant colony optimization.  Most such algorithms work by repeatedly applying one or more low-level heuristics that make a small random change to a network; over many iterations, the algorithms guide this random walk towards better networks by means of a metaheuristic such as natural selection (as in evolutionary algorithms) or metallic annealing (as in simulated annealing).

By comparison, only little cross-over exists between the literature on this problem and that on neural nets~\citep{guihaire2008transitReview, kepaptsoglou2009transitReview, duran2022survey}.  In this thesis, we address this gap, considering ways in which \glspl{gnn} and \gls{rl} may be used to address the \gls{ndp}, both in isolation and in combination with existing metaheuristic approaches.

\section{Background}

\subsection{Why Transit Matters: Cities, Cars, and Mass Transit}\label{subsec:motivation}

\begin{figure*}
    \centering
    \begin{subfigure}[b]{0.381\textwidth}
        \centering
        \includegraphics[width=\textwidth]{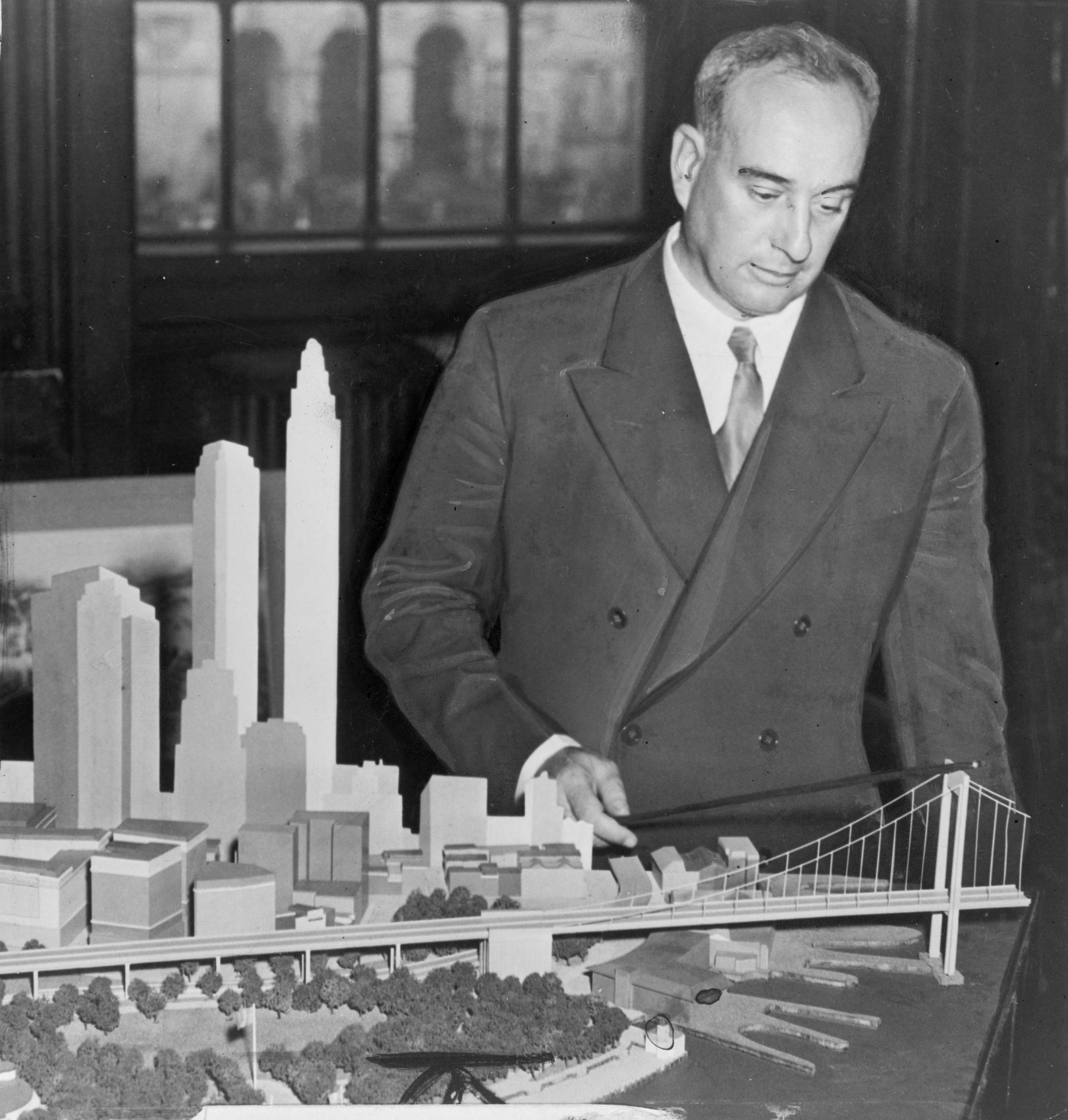}
        \label{fig:moses}
    \end{subfigure}
    \begin{subfigure}[b]{0.4\textwidth}
        \centering
        \includegraphics[width=\textwidth]{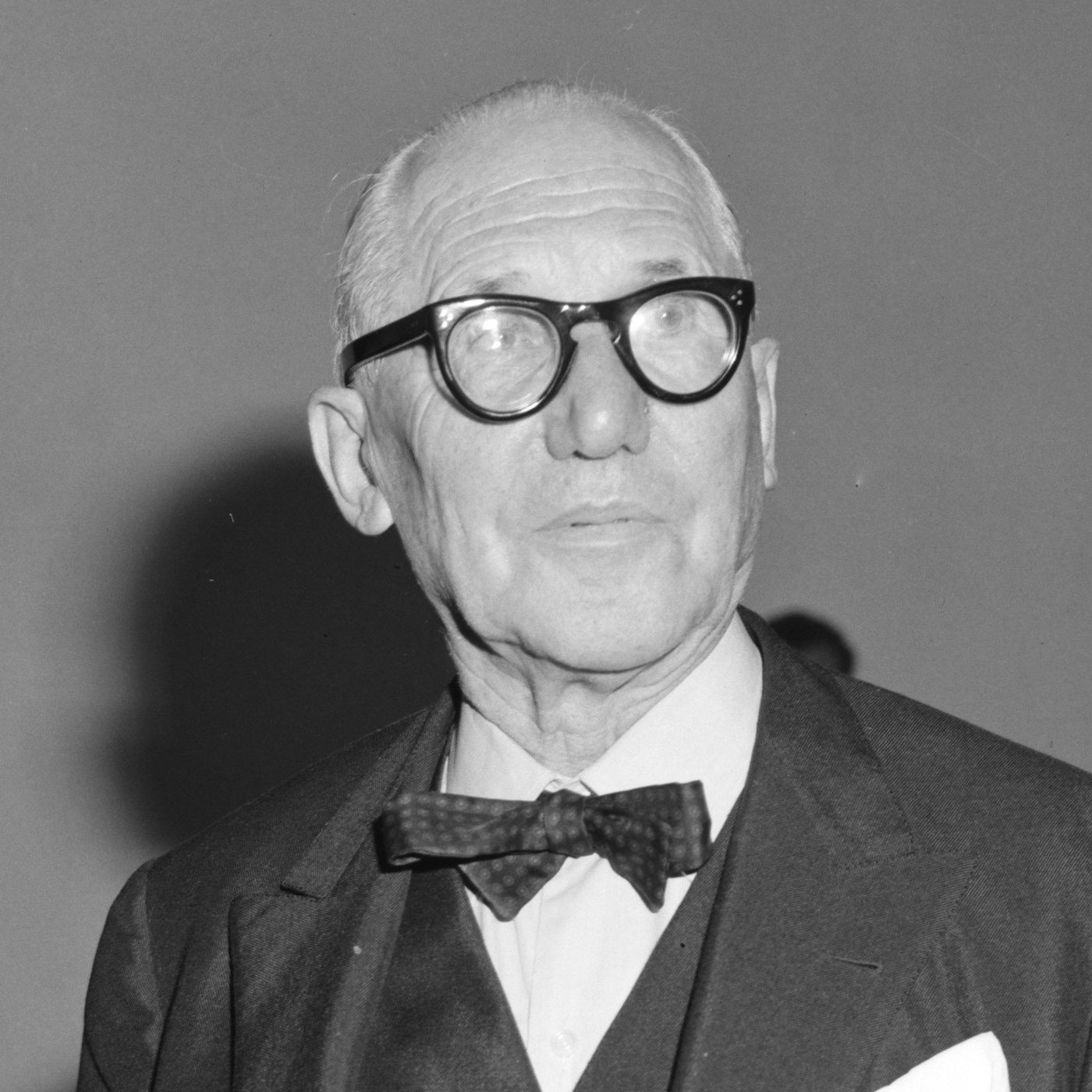}
        \label{fig:lecorbusier}
    \end{subfigure}
    \caption{Robert Moses (left \citep{moses_image}) and Charles-Édouard Jeanneret, also known as Le~Corbusier (right \citep{corbusier_image}), two influential figures in the development of automobile-centric cities.}
    \label{fig:lecorbusier_and_moses}
\end{figure*}

With the arrival of the Model T in 1908, automobiles became a mass-market product.  As more and more city dwellers became automobile owners, various factors conspired to shift the design of cities around the world away from walking and mass transit and towards the automobile as the chief mode of mobility for their residents.  In the 1920s, car manufacturers and drivers' associations engaged in concerted efforts to change public perception of what city streets were for (motorists, rather than pedestrians) and to shape safety laws governing the streets to favour motorists: efforts which proved successful~\cite[chapter~4]{happycity}.  Meanwhile, prominent figures in urban planning such as Le Corbusier and Robert Moses (\autoref{fig:lecorbusier_and_moses}), who saw the personal automobile as the future of urban transportation, shaped their urban designs around cars and influenced many others to do the same~\citetext{\citealp[Chapter~25]{powerbroker}; \citealp[Chapter~11]{leCorbusierCityOfTomorrow}}.

But the middle of the 20th century, the adverse effects of this move towards automobiles were beginning to be seen.  Congestion had revealed itself as a major problem, and evidence quickly mounted that, by inducing demand, adding more road capacity worsens congestion rather than relieving it~\citep{powerbroker, duranton2011fundamental}.  The automobile also was key to enabling urban sprawl: the growth of vast low-density residential developments outside of city centres, from which residents commute for work and other purposes.  Congestion and sprawl together have led to greatly increased pollution, along with negative effects on social cohesion and on commuter's well-being~\citep{sandow2011road, leyden2003social, freeman2001effects}.

The early 21st century has seen the advent of ride-hailing services such as Uber and Lyft, as well as major advancements in autonomous driving technology.  With these developments, the prospect has arisen of a new auto-mobility in which the commuter will not own a car, but will instead summon an \gls{av} when needed, and then dismiss it to be used by others.  It has been suggested that such mobility-as-a-service will reduce congestion and make transportation more efficient~\citep{selfDrivingReducesCongestion}, and that they may act as a useful supplement to public transit~\citep{selfDrivingCouldSupplementPublicTransit}.

Unfortunately, the evidence to date suggests the opposite.  \cite{SchallerConsulting2018} reports on data from a number of major US cities demonstrating that ride-sharing services such as Uber and Lyft have worsened congestion in major U.S. cities, because they primarily draw users away from non-car transit modes, especially public transit, increasing total vehicle miles driven by as much as 80\%.  \cite{harb2018projecting} designed a clever experiment that simulated self-driving personal cars for a sample of real people by giving them free access to human chauffeurs who would drive their cars for them.  This experiment showed that freeing car owners from the need to drive induces more demand for car travel - which will tend to exacerbate the problems of congestion.

\begin{figure}
	\centering\includegraphics[width=0.8\textwidth]{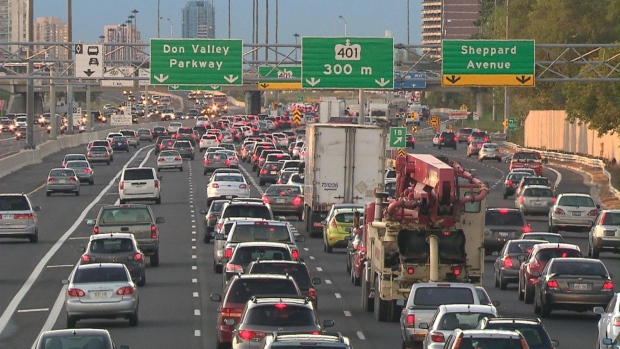}
	\caption{Automobile congestion causing travel delays in the city of Toronto, where it is a commonplace occurrence \citep{ctv2020congestion}.}
	\label{fig:congestion}
\end{figure}

In fact, even in an ideal world, the benefits to be gained from self-driving cars are limited.  A theoretical analysis by~\cite{roughgarden2002bad} compared globally-coordinated routing of cars over a congested road network with ``selfish'' routing.  They demonstrated that even in idealized theoretical scenarios, coordinated routing could improve on selfish routing by at most a factor of $4/3$ in terms of total transit time for all travellers.

Mass public transit systems, such as rail and bus networks, are far more space-efficient than private automobiles and have far less environmental impact~\cite{litman2012evaluating}.  A major challenge facing the architects of such systems, however, is persuading travellers in car-centric cities to use public transit.  In a study of Los Angeles, \cite{Chakrabarti2017getPeopleOutOfTheirCars} analyzed factors determining car or bus use by commuters, and found that frequent service, high speed, and reliability were the most important factors in rider's decisions to use public transit.  The design of transit networks - the spatial layout of the routes themselves - can have a major impact on passenger's travel times, and so can be instrumental in getting drivers out of their cars and onto public transit, reducing congestion and pollution and improving the mobility of all citizens.  

Applying \gls{av} technology to bus systems could help them better provide these qualities by allowing more buses to be operated with less downtime.  Critically, they may also reduce cost - according to~\cite[chapter 9]{cederBook}, driver wages and benefits form the single largest budget item for most transit agencies.  A growing number of cities are engaged in pilot projects with autonomous shuttles~\citep{bloombergJacksonville, smartcitiesdriveDetroit, malagaAutonomousBus}.  Autonomous buses would also be capable of navigating more complex routes, and changing routes more often, than human bus drivers are willing or able to do.  This could make feasible more complex but more efficient system designs, which computational network design methods may be better able to provide than human designers.

\subsection{Neural Nets for Optimization Problems}\label{subsec:nn_review}

\acrfull{co} problems are problems in which a solution comprises the arrangement of finite set of elements in some way, so as to accomplish an objective like minimizing the solution's cost or satisfying a set of constraints.  A well-known example is the \acrlong{tsp}, in which we are given a set of points in space and we seek the order in which to visit those points that minimizes the overall length of the journey.  The points may be visited in any order, so with $n$ points, there are $n!$ possible solutions.  
The \gls{tsp} belongs to the class of NP-Hard problems, meaning the fastest known algorithms for finding the optimal solution take time that grows exponentially with $n$.  Most \gls{co} problems of interest share this property, so there has been much research on algorithms that can find good sub-optimal solutions in feasible amounts of time, such as  \cite{christophides1976worst, kautz1992planning, concordeTspSolver} to name only a few.

Artificial neural nets (henceforth, neural nets) are a broad family of models used in machine learning.  A neural net composes a sequence of ``layers'' to map an $n$-dimensional vector input to an $m$-dimensional output.  The layers together have a set of learnable parameters $\theta$; by using an appropriate learning algorithm to find values for $\theta$, the neural net can be ``trained'' to closely approximate arbitrary functions $f: \mathbb{R}^n \rightarrow \mathbb{R}^m$.  Such a system is called a ``neural net'' because the hierarchical arrangement of layers and their internal structure are loosely inspired by the arrangement and function of biological neurons in animal nervous systems.  Many different types of neural net have been developed to operate on different types of input, including text, images, and graph-structured data.  Neural nets with many layers (typically, more than 3) are often referred to as deep neural nets, and their use in machine learning is referred to as ``deep learning''.  A comprehensive treatment of these models is presented by \cite{goodfellow2016}.

\Acrfull{rl} is a branch of machine learning concerned with training a model, such as a neural net, to interact with an environment so as to maximize the scalar ``reward'' it receives from that environment over time.  \gls{rl} methods are useful in situations where we want to train a model to take actions that maximize (or equivalently, minimize) some quantity over time, but do not know in advance what actions can achieve this.  This is in contrast with supervised learning, where the desired outputs of the model are known exactly in advance, at least on some partial set of inputs.  In most \gls{co} problems, it is difficult to find globally optimal solutions but easier to gauge the quality of a given solution.  This makes \gls{rl} a natural fit to \gls{co} problems, and as documented by~\cite{bengio2021machine}, there is growing interest in the application of \gls{rl} with neural nets to \gls{co} problems such as the \gls{tsp}.  Most of the work cited in this section uses \gls{rl} methods to train neural net models.  

An important early work in this vein was that of \cite{vinyals2015pointer}, who proposed a deep neural net model called a Pointer Network, and trained it via supervised learning to solve instances of the \gls{tsp}.  Subsequent work, such as that of~\cite{dai2017learningCombinatorial}, \cite{Kool2019AttentionLT}, and \cite{sykora2020multi}, has built on this.  These works use similar neural net models with \gls{rl} to construct \gls{co} solutions, and attain impressive performance on the \gls{tsp}, the \gls{vrp}, and other \gls{co} problems.  More recently, \cite{fu2021generalize} train a model on small \gls{tsp} instances and present an algorithm that applies the model to much larger instances.  \cite{choo2022simulation} present a hybrid algorithm of Monte Carlo Tree Search and Beam Search that draws better sample solutions for the \gls{tsp} and \gls{cvrp} from a neural net policy like that of \cite{Kool2019AttentionLT}.

The above approaches all belong to the family of ``construction'' methods, which solve a \gls{co} problem by starting with an ``empty'' solution and adding to it until it is complete - for example, in the \gls{tsp}, this would mean constructing a path one node at a time, starting with an empty path and stopping once the path includes all nodes.  The solutions from these neural construction methods come close to the quality of those from specialized algorithms such as Concorde~\citep{concordeTspSolver}, while requiring much less run-time to compute~\citep{Kool2019AttentionLT}.

By contrast with construction methods, ``improvement'' methods start with a complete solution and repeatedly modify it, conducting a search through the solution space for improvements.  In the \gls{tsp}, this might involve starting with a complete path, and swapping pairs of nodes in the path at each step to see if the path is shortened.  Improvement methods tend to be more computationally costly than construction methods when generating a solution, but their more exhaustive search of the solution space can yield better results.

Like construction methods, much recent work has considered using deep learning and \gls{rl} for improvement methods.  Deep neural nets have been trained to choose the search moves to be made at each step of an improvement method~\citep{hottung2019neural, chen2019learning, d2020learning, wu2021learning, ma2021learning}.  \cite{kim2021learning} train one neural net to construct a set of initial solutions, and another to modify and improve them.  \cite{mundhenk2021symbolic} train a \gls{rnn} via \gls{rl} to construct a population of initial solutions for a genetic algorithm - itself an improvement method - and then use the outputs of the genetic algorithm as data to further train the \gls{rnn}.  And more recently, \cite{ye2024deepaco} train a neural net to provide a heuristic score for choices in \gls{co} problems in the context of an Ant Colony Optimization algorithm.  This work has shown impressive performance on the \gls{tsp}, \gls{vrp}, and similar \gls{co} problems.

We do the same here, applying a method similar to that of~\cite{Kool2019AttentionLT} to the \gls{ndp}.

\subsubsection{Graph Neural Nets}


Many classic \gls{co} problems like the \gls{tsp} and \gls{vrp} lend themselves to being described in terms of a graph.  For this reason, most of the above work uses a type of model called a \gls{gnn} that is designed to operate on graph-structured data \citep{bruna2013spectral,kipf2016semi,defferrard2016spectral,duvenaud2015convolutional}.  Inspired by the success of convolutional neural nets on computer vision tasks, \citet{bruna2013spectral} proposed a neural graph operator that performed convolution in the spectral domain of the graph.  This idea was expanded upon by~\citet{kipf2016semi} and \citet{defferrard2016spectral}.  Meanwhile, \citet{duvenaud2015convolutional} proposed a non-spectral graph convolution based only on the spatial neighbourhoods of nodes.  Since then, \glspl{gnn} have been applied in many domains, such as the predicting chemical properties of molecules~\cite{duvenaud2015convolutional, gilmer2017quantum}, analyzing large web graphs~\cite{ying2018webscale}, and in combination with \gls{rl}, designing printed circuit boards~\cite{mirhoseini2021graph}.  
An overview of \glspl{gnn} is provided by~\cite{battaglia2018relational}.

\begin{figure}
	\centering
	\includegraphics[scale=0.8]{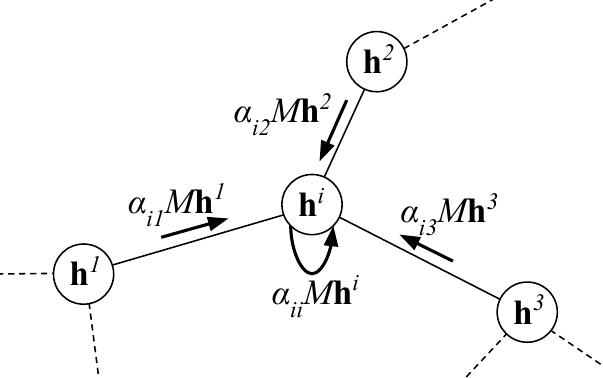}
	\caption{A schematic of the message-passing step of a \gls{gnn} layer for one node in a graph.  As described in \autoref{eqn:message_passing}, node $i$ receives messages $\mathbf{h}^j$ from each neighbouring node and from itself, which are the node's feature transformed by the learned matrix $M$ and scaled by the coefficients $\alpha_{ij}$.  The received messages are summed to form the layer's new embedding of node $i$.}
	\label{fig:message_passing}
\end{figure}

In general, a graph neural net operates on a graph described by an a collection of $d_v$-dimensional node features $\mathbf{x}_i$ for each of the $n$ nodes, the set of edges $\mathcal{E}$, and possibly a collection of $d_e$-dimensional features $E$ for each edge.  Like other neural net architectures, it is composed of a series of layers.  Each layer can be thought of as passing ``messages'' between nodes based on their relationship in the graph, and then having each node form a new embedding by transforming and combining the messages it receives.  Many graph neural net layer types can be described by the following equation:

\nomenclature{$N(i)$}{The set of nodes in a graph which are neighbours of node $i$.}

\begin{equation}\label{eqn:message_passing}
\mathbf{h}^i_{l+1} = \sigma \left ( \sum_{j \in N(i)} \alpha_{ij} M_l \mathbf{h}^j_l \right )
\end{equation}
Here, $\mathbf{h}^i_l$ is an $d_l$-dimensional embedding of node $i$'s descriptor (with $\mathbf{h}^i_0=\mathbf{x}_i$); $M_l$ is a learnable $d_l \times d_{l+1}$  weight matrix; $\alpha_{ij}$ is a weight that indicates how much node $j$ contributes to node $i$'s embedding; $N(i)$ is a set of nodes which are ``connected'' in some sense to $i$; and $\sigma$ is a non-linear function such as the sigmoid function or the ReLU function~\citep{fukushima1969visual}.  \autoref{fig:message_passing} provides a schematic illustration of this process.

The choices of $N(i)$ and $\alpha_{ij}$ differentiate \gls{gnn} architectures from one another.  A simple choice is to have $N(i) = \{ j | (i, j) \in \mathcal{E} \} \cup \{i\}$, and $\alpha_{ij} = |N(i)|^{-1}$: this means at each layer, each node receives a message from itself and its one-hop neighbours in the graph, and averages these before applying $\sigma$ to get its new feature.  If edge weights $e_{ij}$ are provided, another choice would be to let $\alpha_{ij} = e_{ij}$, so that a node's new feature is a weighted average of its neighbours.  $N(i)$ might be expanded to all nodes in the graph, with $\alpha_{ij}$ being a coefficient based on the distance in the graph between $i$ and $j$.
Another choice is that of the Graph Attention Network~\citep{velickovic2018graphattentionnetworks, gatv2conv}: here, $N(i) = \{ j | (i, j) \in \mathcal{E} \} \cup \{i\}$ as in the one-hop \gls{gnn}, but $\alpha_{ij}$ is replaced by $\alpha_{lij} = f(\Theta_l, \mathbf{h}^i_l, \mathbf{h}^j_l, \mathbf{e}_{ij})$, a learned function at each layer of node $i$ and $j$'s embeddings and of an edge feature $\mathbf{e}_{ij}$ if this is available.

\subsection{Optimization of Public Transit}

As stated above, the \gls{ndp} is an NP-hard problem, meaning that it is impractical to find optimal solutions in most cases.  While analytical optimization and mathematical programming methods have shown some success on small instances~\citep{vannes2003AnalyticRouteAndSchedule, guan2006AnalyticRoutePlanning}, they struggle to realistically represent the problem~\citep{guihaire2008transitReview, kepaptsoglou2009transitReview}.  Metaheuristic approaches, as defined by~\cite{sorensen2018history}, have thus been more widely used.  The most popular of these have been genetic algorithms, simulated annealing, and ant-colony optimization, along with hybrids of these methods~\citep{guihaire2008transitReview, kepaptsoglou2009transitReview, yang2020application, duran2022survey, husselmann2023improved}.  Recent work has shown success on the \gls{ndp} with other metaheuristics, such as a sequence-based selection hyper-heuristic~\citep{ahmed2019hyperheuristic}, beam search~\citep{islam2019heuristic}, and particle swarms~\citep{lin2022analysis}.  

Many different low-level heuristics have been applied within these metaheuristic algorithms, but most have in common that they select among possible neighbourhood moves uniformly at random.  One exception is \cite{husselmann2023improved}.  The authors design two heuristics based on a simple model of how each change the heuristic could make would affect the network's quality.  They use this model to adjust the probabilities of different changes being selected.  When they use these heuristics in a genetic algorithm along with a set of standard uniformly-random low-level heuristics, they obtain state-of-the-art results.  However, their simple model ignores passenger trips involving transfers, and how the user's preferences may alter the cost function.  

While neural nets have often been used for predictive problems in urban mobility~\citep{xiong1992transportation, rodrigue1997NNsForLandUseAndTransport, chien2002dynamic, jeong2004bus, akgungor2009NNsForAccidentPrediction, li2020graph} and for other transit optimization problems such as scheduling, passenger flow control, and traffic signal control~\citep{zou2006lightrail, jiang2018passengerInflow, ai2022deep, Yan2023DistributedMD, wang2024large}, they have not often been applied to the \gls{ndp}, and neither has \gls{rl}. Two recent exceptions are notable here: \cite{darwish2020optimising} and \cite{yoo2023reinforcement}.  Both use \gls{rl} to design routes and a schedule that obtain good results on the Mandl benchmark city~\citep{mandl1980evaluation}, a single small city with just 15 transit stops.  \cite{darwish2020optimising} use a \gls{gnn} approach inspired by~\cite{Kool2019AttentionLT}; in our own work we experimented with a nearly identical approach, but found it did not scale beyond very small instances like Mandl.  Meanwhile, \cite{yoo2023reinforcement} use tabular \gls{rl}, an approach which is practical only for small problem sizes.  Both of these approaches also require a new model to be trained on each problem instance.  

\section{The Transit Network Design Problem}\label{sec:tndp}

The planning of transit networks can be divided into three primary sub-problems:

\begin{itemize}
	\item \acrfull{ndp}: establishing the layout of the routes and their stops, given an existing network of roads and candidate stop locations
	\item \acrfull{fsp}: determining the departure frequency on each route
	\item \acrfull{tp}: determining routes and departure times for each vehicle in the system
\end{itemize}  

The three problems can be solved sequentially in isolation, and much work treats them as such \citep{guihaire2008transitReview, kepaptsoglou2009transitReview, abduljabbar2019applications}.  In this work, we consider only the \gls{ndp}.  But we note that their inter-relatedness mean that more optimal solutions may potentially be obtained by optimizing over multiple aspects of the problem at once.

\begin{figure}
    \centering
    \includegraphics[scale=0.5]{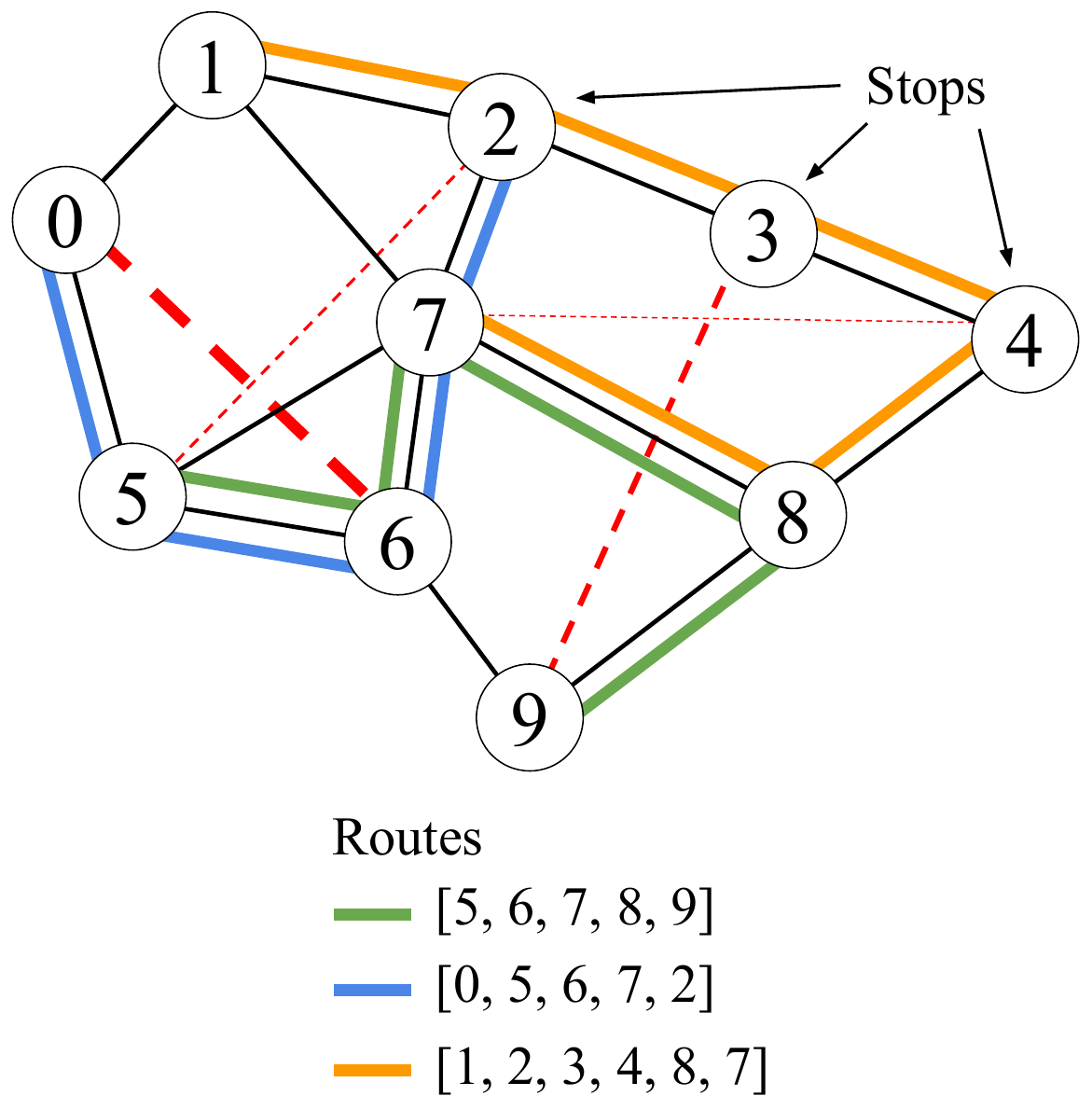}
    \caption{An example city graph with ten numbered nodes and three routes.  Street edges are black, routes are in colour, and two example demands are shown by dashed red lines.  The edges of the three routes form a sub-graph of the street graph $(\mathcal{N}, \mathcal{E}_s)$.  All nodes are connected by this sub graph, so the three routes form a valid transit network.  The demand between nodes 2 and 5, 0 and 6, and 7 and 4 can be satisfied directly by riding on the blue line, while the demand from 3 to 9 requires one transfer: passengers must ride the orange line from node 3 to 8, and then the green line from node 8 to 9.}
    \label{fig:tndp}
\end{figure}

We here use a graph formulation of the \gls{ndp} that is common in the literature~\citep{mumford2013new}.  In this formulation, we are given an augmented graph $\mathcal{G}$ that represents a city:
\nomenclature{$\mathcal{G}$}{A city graph, composed of a set of nodes $\mathcal{N}$, a set of street edges $\mathcal{E}_s$, and a demand matrix $D$.}
\nomenclature{$\mathcal{N}$}{The set of nodes in a city graph.}
\nomenclature{$n$}{The number of nodes in a city graph. $n = |\mathcal{N}|$.}
\nomenclature{$\mathcal{E}_s$}{The set of edges representing streets in a city graph $\mathcal{G}$, wieghted by driving time $\tau_{ij}$.}
\nomenclature{$D$}{An $n \times n$ demand matrix of the number of passengers who want to travel between each pair of nodes in a city graph.}
\nomenclature{$\tau_{ij}$}{The driving time of the edge between nodes $i$ and $j$ in a city graph.}
\nomenclature{$r$}{A transit route.}
\nomenclature{$\mathcal{R}$}{A transit network.}
\nomenclature{$T$}{An $n \times n$ matrix of the driving times between each pair of nodes in a city graph, along the shortest path between them over $\mathcal{E}_s$.}
\nomenclature{SP}{The set of shortest paths over the street edges $\mathcal{E}_s$ between each pair of nodes in a city graph.}

\begin{equation}
 \mathcal{G} = (\mathcal{N}, \mathcal{E}_s, D)
\end{equation}
This augmented graph is comprised of a set $\mathcal{N}$ of $n$ nodes, representing candidate stop locations; a set $\mathcal{E}_s$ of street edges $(i, j, \tau_{ij})$ connecting the nodes, with weights $\tau_{ij}$ indicating the time it takes a transit vehicle to drive between the nodes; and an $n \times n$ \gls{od} matrix $D$ giving the travel demand (in number of trips) $D_{ij}$ between every pair of nodes $i,j \in \mathcal{N} \times \mathcal{N}$.  Each node $i$ has an associated position in 2D space, $\mathbf{p}_i$.  A transit route $r$ is defined as a path through the street graph $\{\mathcal{N}, \mathcal{E}_s\}$: a sequence of nodes $[v_0, v_1,...,v_{k-1},v_k]$ where every pair of consecutive nodes $v_i, v_{j=i+1}$ has a matching edge ($(i, j, \tau_{ij}) \in \mathcal{E}_s$), and where no node is visited more than once in the sequence ($i \neq j \Rightarrow v_i \neq v_j$).  A transit network $\mathcal{R}$ is defined as a set of transit routes.  The goal in the \gls{ndp} is to find a transit network $\mathcal{R}$ that minimizes a cost function $C: \mathcal{G}, \mathcal{R} \rightarrow \mathbb{R}^+$.

We deal with the symmetric \gls{ndp}, meaning we assume that $D = D^\top$,
$(i, j, \tau_{ij}) \in \mathcal{E}_s$ iff. $(j, i, \tau_{ij}) \in \mathcal{E}_s$, and all routes are traversed both forwards and in reverse by vehicles on them, linking their nodes in both directions.  An example city graph with a transit network is shown in \autoref{fig:tndp}.


\nomenclature{$\mathcal{G_R}$}{The route graph induced over city graph $\mathcal{G}$ by transit network $\mathcal{R}$.}
\nomenclature{$\mathcal{E_R}$}{The edge set of the route graph $\mathcal{G_R}$.}

A transit network $\mathcal{R}$ induces a {\bf route graph}, which we denote $\mathcal{G_R}$.  The route graph is a graph indicating which trips are possible by transit between stops.  An edge between nodes $i$ and $j$ in the route graph means that a passenger can get from $i$ to $j$ by taking only a single transit route, with no transfers needed.  The route graph shares the same set of nodes $\mathcal{N}$ as $\mathcal{G}$, but the edges in its edge set $\mathcal{E_R}$ correspond to direct trips between nodes via transit routes:

\nomenclature{$\tau_{rij}$}{The time taken to get from node $i$ to $j$ by taking route $r$.}
\nomenclature{$\tau_{\mathcal{R}ij}$}{The shortest time to get from node $i$ to $j$ on any route in $\mathcal{R}$.}

\begin{equation}
\mathcal{E_R} = \{ (i,j, \tau_{rij}) | \; \exists \; r \in \mathcal{R}, i \in r \land j \in r \}
\end{equation}

For example, a route $r = [0, 1, 3, 4]$ in a city with five nodes induces edges in $\mathcal{G_R}$ between 0 and 1, 0 and 3, 0 and 4, 1 and 3, 1 and 4, and 3 and 4.  
 
Each edge in $\mathcal{E_R}$ also has a weight $\tau_{rij}$, the time of the trip between $i$ and $j$ provided by route $r$, where $r$ is the route that provides the shortest direct trip between $i,j$ of any route in $\mathcal{R}$.  It may be that $\tau_{rij} > T_{ij}$, as the route $r$ does not necessarily follow the shortest path between $i$ and $j$.

\nomenclature{SR}{The set of shortest paths between all node pairs over a route graph $\mathcal{G_R}$.}
\nomenclature{$sp_{ij}$}{The shortest path between all nodes $i$ and $j$ over route graph $\mathcal{G_R}$.}
\nomenclature{$n^T_{\mathcal{R}ij}$}{The number of transfers required to travel between nodes $i$ and $j$ over transit network $\mathcal{R}$.}

With $\mathcal{G_R}$ defined, we can then define $\textup{SR}$ as the collection of shortest paths $\textup{sr}_{ij}$ between all node pairs $i,j$ over the route graph $\mathcal{G_R}$.  $\mathcal{G_R}$ and $\textup{SR}$ are useful in several ways.  Firstly, they let us count $n^T_{\mathcal{R}ij}$, the number of transfers required between two nodes $i$ and $j$: it is simply two less than the number of nodes along $sp_{ij}$: $n^T_{\mathcal{R}ij} = |\textup{sr}_{ij}| - 2$, as long as the shortest path is chosen by breaking ties in favour of paths with fewer nodes (which we do).  Secondly, they let us find the time $\tau_{\mathcal{R}ij}$ for the shortest transit trip between $i$ and $j$, by summing the edge weights $\tau_{rij}$ along $\textup{sr}_{ij}$.

\nomenclature{$p_i$}{The 2D position of city graph node $i$.}

We note that in general neither $\mathcal{G}$ nor $\mathcal{G_R}$ are metric graphs: that is, it is not generally true that $\tau_{ij} = ||p_i - p_j||$ or that $\tau_{rij} = ||p_i - p_j||$.

\subsection{Constraints}\label{subsec:constraints}

In this version of the \gls{ndp}, a transit network must satisfy the following constraints to be considered valid:

\nomenclature{$S$}{The number of routes required by the user for a transit network.}
\nomenclature{$m_\text{min}$}{The minimum number of stops allowed on a route, set by the user.}
\nomenclature{$m_\text{max}$}{The maximum number of stops allowed on a route, set by the user.}

\begin{enumerate}
    \item Connectedness: $\mathcal{G_R}$ must be connected, providing some path over transit between every pair of nodes $(i,j) \in \mathcal{N}$ for which $D_{ij} > 0$.
    \item Number of routes: $\mathcal{R}$ must contain exactly $S$ routes ($|\mathcal{R}| = S$), where $S$ is a parameter set by the user.
    \item Number of stops: Every route $r \in \mathcal{R}$ must obey $m_\text{min} \leq |r| \leq m_\text{max}$, where $m_\text{min}$ and $m_\text{max}$ are parameters set by the user that limit the number of stops a route may visit.
    \item Simple paths: A route $r \in \mathcal{R}$ must be a simple path in $\mathcal{G}$: that is, it must not contain cycles, and every pair of consecutive nodes in $r$ must have a matching edge in $\mathcal{E}_s$.
\end{enumerate}

\subsection{Cost Function}\label{subsec:cost_function}

In this work, we assume that the cost function has the following structure.  The passenger cost, $C_p$, is the average passenger trip time over the network.  It is computed by assuming that every passenger chooses the shortest path available over the transit network from their origin to their destination, including by transferring between transit routes any number of times by getting off of route A at stop $i$ and then boarding route B which also stops at $i$.  Each transfer extends the time of a trip because the passenger must wait for the vehicle on route B to arrive; also, in reality transfers are viewed as inconvenient by passengers even if waiting times are short, so passengers tend to try to avoid them where possible~\citep{Chakrabarti2017getPeopleOutOfTheirCars}.  To reflect these factors, we impose a time penalty $p_T$ once for each transfer made on a trip, which will discourage the selection of trips containing transfers when using a shortest-path criterion, and set this time penalty to five minutes (300 seconds).  This is common practice for computing $C_p$ for the \gls{ndp} \citep{mumford2013new, john2014routing, kilic2014demand, husselmann2023improved}.  $C_p$ is thus computed as:

\nomenclature{$C$}{A cost function for a city graph $\mathcal{G}$ and a transit network $\mathcal{R}$.}
\nomenclature{$C_p$}{The passenger cost part of the cost function, defined as the average time of all passenger trips via transit.}
\nomenclature{$C_o$}{The operator cost part of the cost function, defined as the total driving time of all transit routes.}
\nomenclature{$\alpha$}{A weight in the range $[0,1]$ controlling the trade-off between passenger and operator costs in the cost function.}
\nomenclature{$\tau_r$}{The time taken to drive route $r$ in one direction.}
\nomenclature{$p_T$}{The time penalty incurred when a passenger makes a transfer between transit routes.  Set as 5 minutes everywhere here.}

\begin{equation}
    C_p(\mathcal{G}, \mathcal{R}) = \frac{\sum_{i,j} D_{ij}(\tau_{\mathcal{R}ij} + p_T n^T_{\mathcal{R}ij})}{\sum_{i,j} D_{ij}}
\end{equation}

The operating cost is the total driving time of the routes, or total route time:
\begin{equation}
    C_o(\mathcal{G}, \mathcal{R}) = \sum_{r \in \mathcal{R}} \tau_r
\end{equation}
Where $\tau_r$ is the time needed to completely traverse a route $r$ in one direction.

\nomenclature{$C_c$}{A component of the cost function used to penalize constraint violations.}
\nomenclature{$F_{un}$}{The fraction of node pairs in a city graph that are not connected by a transit network $\mathcal{R}$.}
\nomenclature{$F_s$}{The total number of stops by which all routes in a transit network go over the maximum or under the minimum allowed numbers.}
\nomenclature{$\delta_v$}{A delta function which is 1 if a transit network $\mathcal{R}$ violates any constraints, and 0 otherwise.}

To enforce constraints 1 and 3 on $\mathcal{R}$, we use a third term $C_c$, itself the sum of three terms:
\begin{align}
	C_c(\mathcal{G}, \mathcal{R}) =  F_{un} + F_s + 0.1 \delta_v
\end{align}

Here, $F_{un}$ is the fraction of node pairs $(i,j)$ with $D_{ij} > 0$ for which $\mathcal{R}$ provides no path: a measure of how many possible violations of constraint 1 really occur in $\mathcal{R}$. $F_s$ is a similar measure for constraint 3, but it is proportional to the actual number of stops more than $m_\text{max}$ or less than $m_\text{min}$ that each route has.  Specifically:
\begin{align}
	F_s = \frac{\sum_{r \in \mathcal{R}} \max(0, m_\text{min} - |r|, |r| - m_\text{max})}{S * m_\text{max}}
\end{align}

$\delta_v$ is a delta function that takes value 0 if $F_{un} = 0$ and $F_s = 0$, and takes value 1 otherwise.  

We use fractional measures $F_{un}$ and $F_s$ instead of absolute measures so that $C_c$ will tend to fall in the range $[0, 1]$ regardless of the size of the city graph.  This is desirable for reasons relating to the numerical stability of neural net training.  However, it has the drawback that if the city graph $C$ has very many nodes (large $n$), and if $\mathcal{R}$ violates only a few constraints, $F_{un}$ and $F_s$ may become vanishingly small compared to $C_o$ and $C_p$.  This could lead to algorithms ignoring small numbers of constraint violations.  To prevent this, we include the term $0.1 \delta_v$, which ensures that if any constraints are violated, $C_c$ cannot be less than 0.1, but it will be 0 if all constraints are respected.  This ensures that even one violation will have a significant impact on overall cost $C(\mathcal{G}, \mathcal{R})$, no matter the size of the city graph.

$C_c$ does not penalize violations of constraints 2 and 4, because as we will show in later chapters, the \gls{ndp} algorithms we consider in this thesis always respect those constraints by design.

The complete cost function is a weighted sum of the three parts:
\nomenclature{$\beta$}{The weight of the constraint violation term in the cost function.}
\begin{align}
    C(\mathcal{G}, \mathcal{R}) = \alpha w_p C_p(\mathcal{G}, \mathcal{R}) + (1 - \alpha) w_o C_o(\mathcal{G}, \mathcal{R}) + \beta C_c(\mathcal{G}, \mathcal{R})
\end{align}

The weight $\alpha \in [0, 1]$ controls the trade-off between passenger and operating costs, while $\beta$ is the penalty assigned for each constraint violation.  In practice, we hold $\beta$ constant in our experiments.  $w_p$ and $w_o$ are re-scaling constants chosen so that $w_p C_p$ and $w_o C_o$ both vary roughly over the range $[0, 1]$ for different $\mathcal{G}$ and $\mathcal{R}$; this is done so that $\alpha$ will properly balance the two, and to stabilize training of the \gls{gnn} policy.  The values used are:

\nomenclature{$w_p$}{The scaling parameter for the passenger cost.}
\nomenclature{$w_o$}{The scaling parameter for the operator cost.}

\begin{align}
    w_p &= (\max_{i,j}T_{ij})^{-1} \\
    w_o &= (S\max_{i,j}T_{ij})^{-1}
\end{align}

\section{Contribution to original knowledge}

In this work, we train neural net models using \gls{rl} to construct transit networks, and then use these models in combination with existing metaheuristic search algorithms for the \gls{ndp}.  We show that this technique is competitive with the state of the art on large benchmark cities, and sets a new state of the art on the largest benchmark city in terms of the operating cost of the transit network.  Furthermore, we show that our technique can be applied to a very large real-world problem instance and find transit networks that improve on the city's existing transit.  To our knowledge, the work presented in this thesis represents the first successful attempt to use deep reinforcement learning to address the \gls{ndp} for cities of realistic size.  

As noted by~\cite{fan2010metaheuristic}, metaheuristic search algorithms depend on three factors for their success:
\begin{enumerate}
	\item The representation of solutions,
	\item The algorithm for constructing the initial solution,
	\item The ``moves'' in solution space chosen at each step of search.
\end{enumerate}
Our neural policies share a standard solution representation (factor 1) with most metaheuristic approaches to this problem.  We use our neural policies to augment the second and third factors, showing in each case that this leads to significant improvements in performance over baseline algorithms.



\section{Dissertation Structure}

\begin{figure}
	\centering
	\includegraphics[width=\columnwidth]{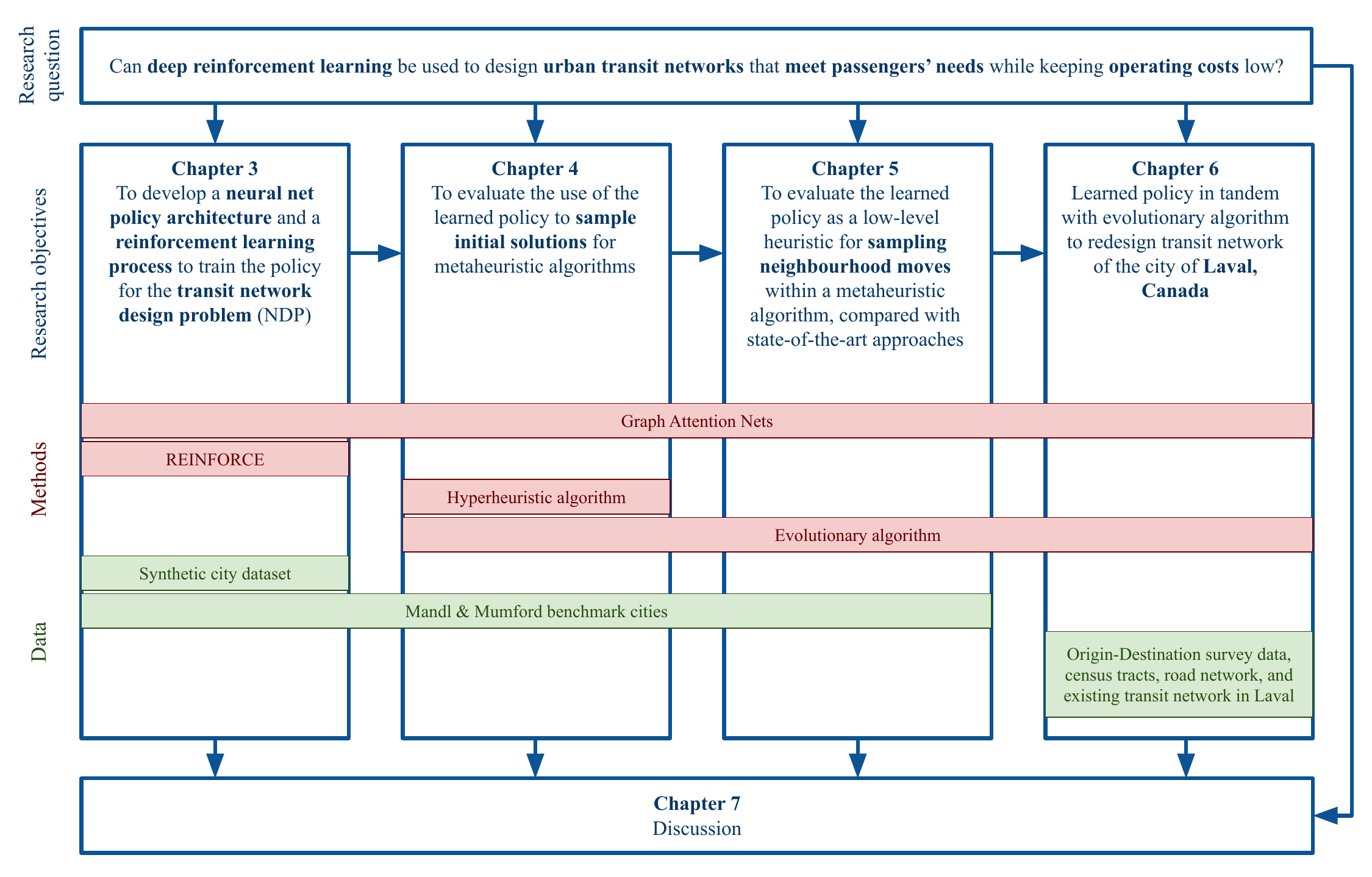}
	\caption{Overview of thesis research}
	\label{fig:thesis_flowchart}
\end{figure}

Each chapter of this dissertation concerns one of four research objectives, each of which is based on our central research question: Can deep reinforcement learning be used to design effective urban transit networks?  By ``effective'', we here mean that they meet passengers' needs for efficient trips with few transfers, while also meeting the transit agency's need to keep operating costs affordably low.  \autoref{fig:thesis_flowchart} outlines these research objectives, and the data and methods we use to accomplish them.  

In \autoref{chap:neural_policy}, we first formulate a construction approach to the \gls{ndp} as a \gls{mdp}.   Based on this formulation, we propose an architecture for a neural net model to address the \gls{ndp}, and an \gls{rl} algorithm and synthetic dataset to train this model to serve as an effective policy on the \gls{mdp}.  We present the results achieved by this policy on the widely-used Mandl~\citep{mandl1980evaluation} and Mumford~\citep{mumford2013new} benchmark cities.

In \autoref{chap:initialization}, we address the second of Fan and Mumford's three factors by using the neural policies trained in \autoref{chap:neural_policy} to generate the initial transit networks in two existing metaheuristic algorithms for the \gls{ndp}.  We compare the results with those achieved using the original initialization procedures of those algorithms, again using the Mandl and Mumford benchmark cities as our points of comparison.

In \autoref{chap:neighbourhood_moves}, we consider the use of the same neural policies to select neighbourhood moves within an evolutionary algorithm.  This addresses the third of Fan and Mumford's factors.  We compare the results with those of an evolutionary algorithm that does not use a learned policy to select neighbourhood moves, and perform a series of ablation studies to understand the impact of different components of our algorithm.  All experiments are again performed on the Mandl and Mumford benchmark cities.

In \autoref{chap:real_world}, we apply \autoref{chap:neighbourhood_moves}'s synthesis of a neural policy and an evolutionary algorithm to a simulated city based on real-world data from the city of Laval, Canada.  We perform experiments under three different assumptions about the planner's preferences, and compare the results to the city's current real transit network.  Finally, we present some concluding remarks in \autoref{chap:conclusion}.

\chapter{Designing Transit Networks with a Graph Attention Net}\label{chap:neural_policy}

In this chapter, we describe our approach to solving the \gls{ndp} using deep reinforcement learning alone.  We begin by presenting a formulation of the \gls{ndp} as a \acrlong{mdp}.  We then describe the neural net policy architecture that we designed to address this \acrlong{mdp}, and the reinforcement learning algorithm we used to train this policy.  We then evaluate learned policies on a widely-used benchmark, and consider the quality of these results in comparison with metaheuristic improvement methods.

\section{Markov Decision Process Formulation}\label{sec:mdp}

\nomenclature{$t$}{The timestep of an \gls{mdp}.}
\nomenclature{$s_t$}{The state of an \gls{mdp} at timestep $t$.}
\nomenclature{$\mathcal{S}$}{The set of possible states of an \gls{mdp}.}
\nomenclature{$\mathcal{A}_t$}{The set of possible actions at timestep $t$ of an \gls{mdp}.}
\nomenclature{$a_t$}{The action taken at timestep $t$ of an \gls{mdp}.}
\nomenclature{$a_t$}{The action taken at timestep $t$ of an \gls{mdp}.}
\nomenclature{$\pi$}{The policy that determines an agent's actions in an \gls{mdp}.}
\nomenclature{$\pi$}{A policy parameterized as a neural net with parameter values given by $\theta$.}
\nomenclature{$R_t$}{The reward received at timestep $t$ of an \gls{mdp}.}
\nomenclature{$G_t$}{The return (cumulative discounted reward) received by an agent in an \gls{mdp} from timestep $t$ onward.}
\nomenclature{$\gamma$}{The discount factor used to compute the return $G_t$ received by an agent in an \gls{mdp}.}
\nomenclature{$t_\text{end}$}{The final timestep of an \gls{mdp}.}
\nomenclature{$r_t$}{The route currently under construction at timestep $t$ in an \gls{mdp}.}
\nomenclature{$\mathcal{R}_t$}{The set of finished transit routes at timestep $t$ in an \gls{mdp}.}

A \acrfull{mdp} is a formalism originating with \cite{bellman1958dynamic}, and which is commonly used to define problems in \acrlong{rl}~\cite[Chapter~3]{sutton2018reinforcement}.  In an \gls{mdp}, an \textbf{agent} interacts with an {\bf environment} over a series of {\bf timesteps} $t$, starting at $t=1$.  At each timestep, the environment has \textbf{state} $s_t \in \mathcal{S}$, and the agent observes the state and takes some \textbf{action} $a_t \in \mathcal{A}_t$, where $\mathcal{A}_t$ is the set of available actions at $t$.  The environment then transitions to a new state $s_{t+1} \in \mathcal{S}$ according to the state transition distribution $P(s_{t+1} | s_t, a_t)$, and the agent receives a numerical \textbf{reward} $R_t \in \mathbb{R}$ according to the reward distribution $P(R_t | s_t, a_t, s_{t+1})$.  The agent chooses actions according to its \textbf{policy} $\pi(a_t|s_t)$, which is a probability distribution over $\mathcal{A}_t$ given the state $s_t$.  In \gls{rl}, the goal is to learn a policy $\pi$ that maximizes the return $G_t$, defined as a time-discounted sum of rewards: 
\begin{equation}\label{eqn:return}
G_t = \sum^{t_\text{end}}_{t'=t} \gamma^{t' - t} R_{t'}
\end{equation}

Where $\gamma \in [0,1]$ is a parameter that discounts rewards farther in the future, and $t_\text{end}$ is the final timestep of the \gls{mdp}.  The sequence of states visited, actions taken, and rewards received from $t=0$ to $t_\text{end}$ constitutes one \textbf{episode} of the \gls{mdp}.

We here describe the \gls{mdp} we use to represent a construction approach to the \gls{ndp}.  As shown in \autoref{eqn:state}, the state $s_t$ is composed of the set of routes $\mathcal{R}_t$ planned so far, and an incomplete route $r_t$ which is being planned.  
\begin{equation}\label{eqn:state}
    s_t = (\mathcal{R}_t, r_t)
\end{equation}
The starting state is $s_1 = (\mathcal{R}_1 = \{\}, r_1 = [])$.  At a high level, the \gls{mdp} alternates at every timestep $t$ between two modes: on odd-numbered $t$, the agent selects an extension to the route $r_t$ that it is currently planning; on even-numbered $t$, the agent chooses whether or not to stop extending $r_t$ and add it to the set of finished routes.

On odd-numbered timesteps, the available actions are drawn from $\textup{SP}$, the set of shortest paths between all pairs of nodes in $\mathcal{G}$.  If $r_t = []$, then:
\begin{equation}
 \mathcal{A}_t = \{a \; | \; a \in \textup{SP}, |a| \leq m_\text{max}\}   
\end{equation}
Otherwise, $\mathcal{A}_t = \text{EX}_{r_t}$, where $\text{EX}_{r_t}$ is the set of paths $a \in \textup{SP}$ that satisfy all of the following conditions:
\begin{itemize}
    \item $(i,j,\tau_{ij}) \in \mathcal{E}_s$ (the set of street edges), where $i$ is the first node of $a$ and $j$ is the last node of $r_t$, or vice-versa
    \item $|a| \leq m_\text{max} - |r_t|$, to respect transit network constraint 3
    \item $a$ and $r_t$ have no nodes in common, to respect transit network constraint 4
\end{itemize}

\begin{figure}
	\centering
	\includegraphics[scale=0.5]{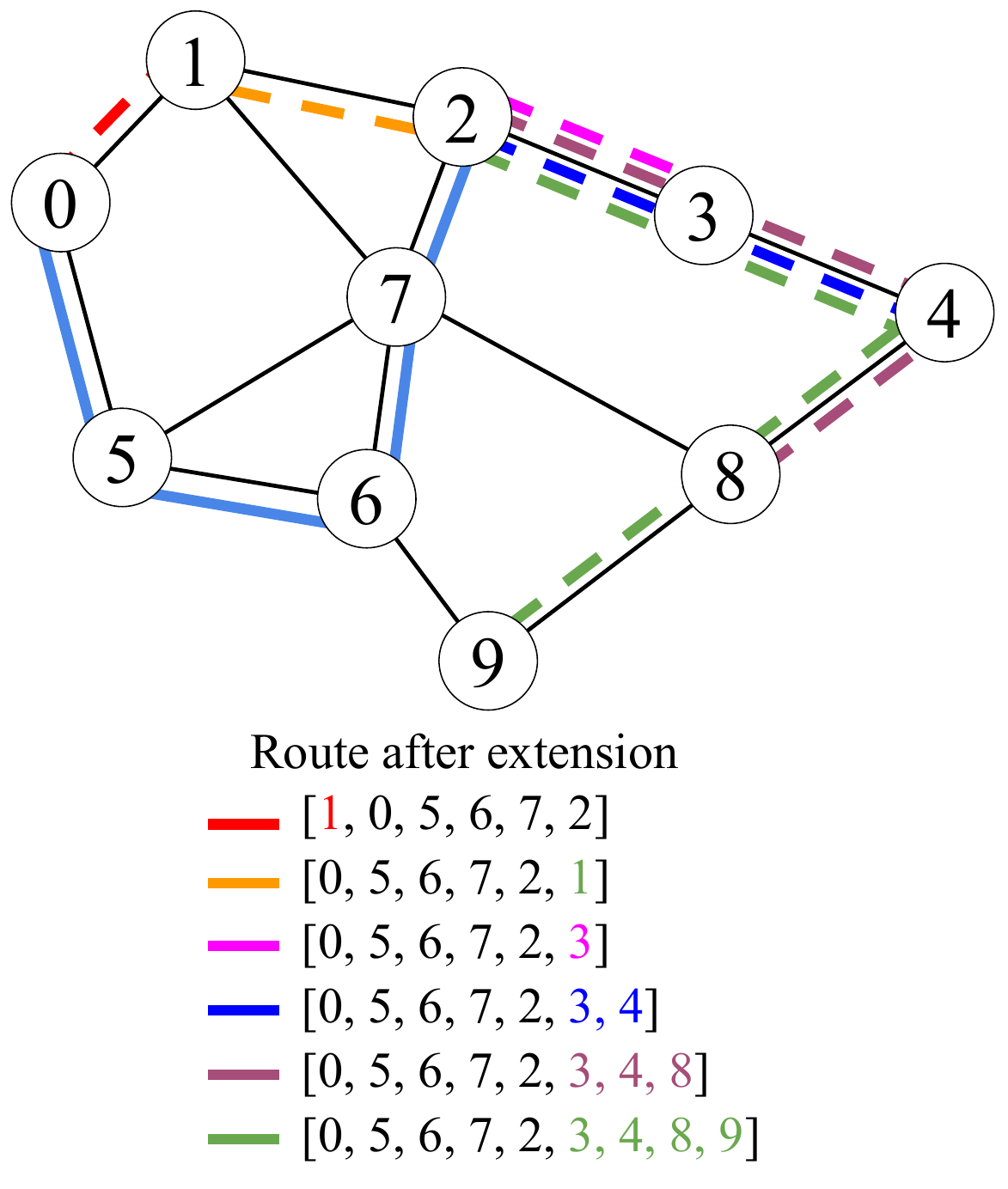}
	\caption{An in-progress route $r$ with the possible extensions available to it, which form $\mathcal{A}_t$.  The route so far is shown with solid blue lines, while the extensions are dashed lines, with different colours for each extension.  The corresponding routes after adding each extension are listed below the graph.  Only one extension (to node 1, shown in red) is possible from the route's starting terminal, since all nodes reachable from 1 are already on the route.  Five extensions are possible from the other end at this step.}
	\label{fig:extensions}
\end{figure}

\autoref{fig:extensions} shows an example of the extensions available to a partial route.  Once a path $a_t \in \mathcal{A}_t$ is chosen, $r_{t+1}$ is formed by appending $a_t$ to the beginning or end of $r_t$ as appropriate.


On even-numbered $t$, the action space depends on the number of stops in $r_t$:
\begin{align}\label{eqn:halt_actions}
    \mathcal{A}_t = \begin{cases}
        \{\textup{continue}\} & \text{if} |r_t| < m_\text{min} \text{ and } |\text{EX}_{r_t}| > 0 \\
        \{\textup{halt}\} & \text{if} |r_t| = m_\text{max} \text{ or } |\text{EX}_{r_t}| = 0 \\        
        \{\textup{continue}, \textup{halt}\} & \text{otherwise}
    \end{cases}
\end{align}
If $a_t = \textup{halt}$, $r_t$ is added to $\mathcal{R}_t$ to get $\mathcal{R}_{t+1}$, and $r_{t+1} = []$ is a new empty route.  If $a_t = \textup{continue}$, then $\mathcal{R}_{t+1} = \mathcal{R}_t$ and $r_{t+1}=r_t$.  Thus, the full state transition distribution is deterministic. 

The episode ends when the $S$-th route is added to $\mathcal{R}$: that is, if $|\mathcal{R}_{t+1}| = S$, the episode ends at timestep $t$.  This ensures network constraint 2 will be respected.  The output transit network is set to $\mathcal{R} = \mathcal{R}_{t+1}$, and the final reward is $R_{t_\text{end}} = -C(\mathcal{G}, \mathcal{R})$. At all prior steps, $R_t = 0$.

This \gls{mdp} formalization imposes some helpful biases on the kinds of transit networks we consider.  First, it requires any route connecting $i$ and $j$ to stop at all nodes along some path between $i$ and $j$, biasing planned routes towards covering more nodes.  Second, it biases routes toward directness by forcing them to be composed of shortest paths.  While an agent may construct arbitrarily indirect routes by choosing paths with length 2 at every step, this is unlikely because in a realistic street graph, the majority of paths in $\textup{SP}$ are longer than two nodes.  Third, the alternation between deciding to continue or halt a route and deciding how to extend the route means that the probability of halting does not depend on how many different extensions are available; so a policy learned in environments with fewer extensions should generalize more easily to environments with more, and vice versa.

\begin{figure}
    \centering
    \includegraphics[width=0.8\columnwidth]{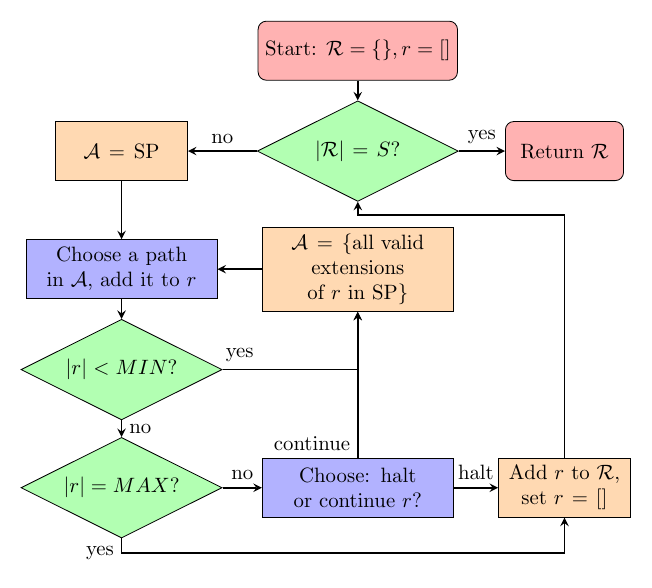}
    \caption{A flowchart of the transit network construction process defined by our \gls{mdp}.  Blue boxes indicate points where the timestep $t$ is incremented and the agent selects an action.  Red nodes are the beginning and ending of the process.  Green nodes are hard-coded decision points.  Orange nodes show updates to the state $\mathcal{S}$ and action space $\mathcal{A}$.}
    \label{fig:mdp_flowchart}
\end{figure}

\section{Neural Net Architecture and Training}\label{sec:system}

Reinforcement learning algorithms can be broadly categorized into value-learning methods and policy-gradient methods.  Value-learning algorithms attempt to estimate the ``value'', or time-discounted cumulative return, of states and actions.  If the value estimates are accurate, they can be used to define a policy $\pi$ that simply chooses the action with the highest estimated value at each step.  Policy gradient algorithms, on the other hand, view a policy as a probability distribution  $\pi(a|s)$ over actions given the current state, and they attempt to learn the policy's distribution directly.  They do this by updating an action's probability based on the observed results, increasing $\pi(a|s)$ when $a$ leads to higher-than-expected returns, and decreasing $\pi(a|s)$ in the opposite case.

Policy gradient methods produce non-deterministic policies that can be sampled stochastically to produce diverse solutions.  This makes them well-suited to \gls{co} problems: finding optimal solutions to these is often intractable, so it is useful to be able to sample multiple promising solutions and choose the best of them.  This property also makes these policies useful as a source of random ``neighbourhood moves'' in the space of solutions to a \gls{co}, the purpose to which we will put them later in \autoref{chap:neighbourhood_moves}.  For these reasons, we chose to use policy gradient methods of deep reinforcement learning to learn policies for the \gls{ndp}.

\nomenclature{$\theta$}{A setting of values for the parameters of a neural net.}

Our policy $\pi_\theta(a|s)$ is a neural net parameterized by $\theta$, which we train to maximize the cumulative return $G_t$ on the construction \gls{mdp} described in \autoref{sec:mdp}.  By then following this policy on the \gls{mdp} for some city $\mathcal{G}$, we can obtain a transit network $\mathcal{R}$ for that city.  We refer to this learned policy as the Learned Constructor (LC).  The policy can be used for planning in one of two modes: stochastic mode, where we randomly sample each action $a_t$ from the distribution $\pi_\theta(\cdot | s_t)$, and greedy mode, where we ``greedily'' choose the highest-scored action under the policy, $a_t = \max_{a \in \mathcal{A}} \pi(a|s_t)$.
Instead, the cost function's constraint-violation term $\beta C_c$ increases cost in proportion to the number of constraint violations, so that the algorithm will be more likely to keep solutions with fewer violations, eventually driving $C_c$ to zero.

\section{Policy Architecture}

\nomenclature{$\mathbf{y}_i$}{A node embedding vector computed by a graph neural net for node $i$.}

The policy $\pi_\theta$ is a neural net with three components: a \gls{gat} ``backbone'', a halting module $\textup{NN}_{halt}$, and an extension module $\textup{NN}_{ext}$.  The \gls{gat} outputs node embeddings $\mathbf{y}_i$, which are operated on by one of two policy ``heads'', depending on the timestep: $\textup{NN}_{ext}$ for choosing among extensions when $t$ is odd, and $\textup{NN}_{halt}$ for deciding whether to halt when $t$ is even.  

\begin{figure}
	\centering
	\includegraphics[width=\linewidth]{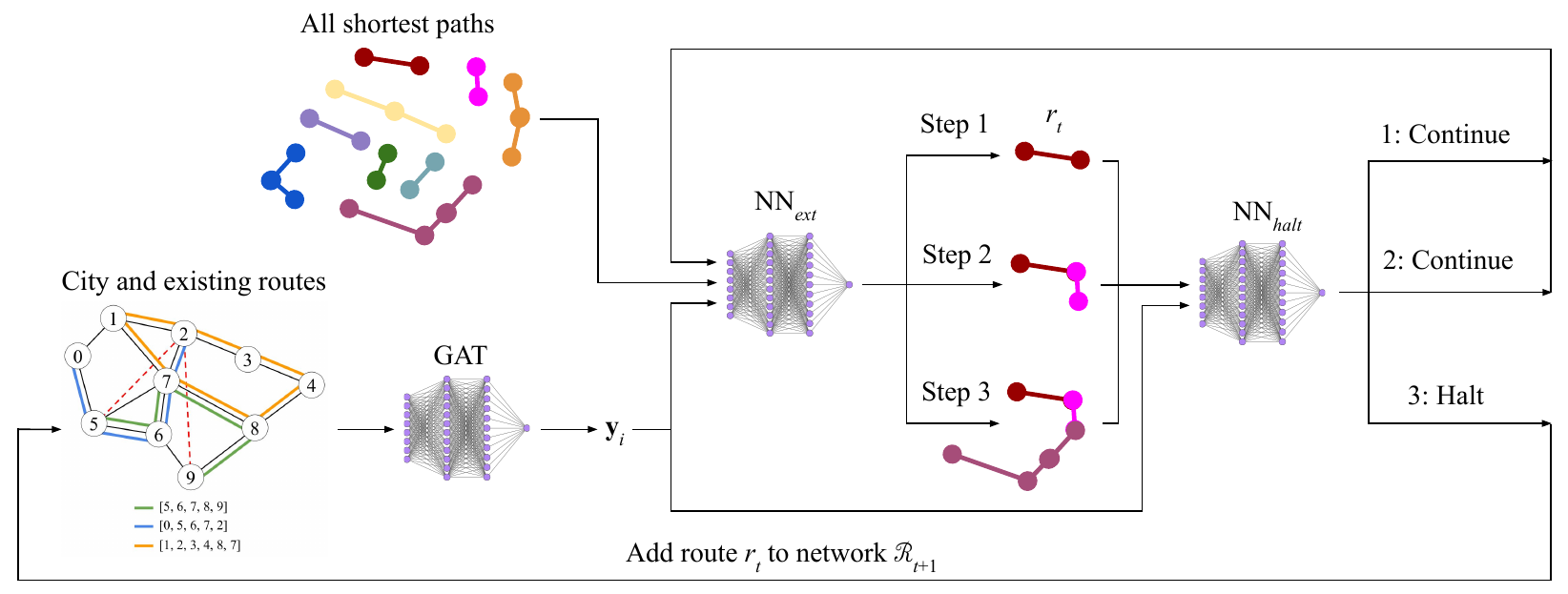}
	\caption{A schematic illustration of the role of each component of the policy net in the network construction process.  This loop continues until $|\mathcal{R}_t| = S$, that is, we have the desired number of routes.}
	\label{fig:nn_route_construction}
\end{figure}

\autoref{fig:nn_route_construction} illustrates the three components' role in the transit network construction process.

\subsection{Graph Attention Backbone}

As described in \autoref{sec:tndp}, we view the \gls{ndp} as a problem on graphs.  This makes \glspl{gnn} a natural design choice for a neural net architecture for this problem.  As described in \autoref{subsec:nn_review}, many \gls{gnn} architectures have been proposed.  Graph convolutional nets are among the earliest form of \gls{gnn} architectures, and have been used successfully in many tasks - as has a much-simplified model linear model, the Simplified Graph Convolution.

However, these models assume a predefined adjacency matrix for the graph.  This allows only the use of scalar edge features, and by design treats them as adjacencies that directly weight the contributions of a node's neighbours to its feature at the next \gls{gnn} layer.  In the \gls{ndp}, we must deal not only with the fixed street graph of a city $\mathcal{G}$, but also with the route graph $G_{\mathcal{R}}$, which changes at each step of the construction \gls{mdp}, and with its demand matrix, which can be viewed as a fully-connected graph with edge weights $D_{ij}$.  These three graphs each have distinct edge sets, with edge weights signifying different things, but share the same node set $\mathcal{N}$.  They can be viewed as forming a multi-layer graph.   Graph convolutions are not well-suited to this kind of structure.

One way of dealing with this multi-layer graph is to treat it as a fully-connected simple graph, with edge features that combine the edge information from $\mathcal{E}_s$, $\mathcal{E_R}$, and $D$.   Such an edge feature $\mathbf{e}_{ij}$ for the edge linking nodes $i,j$ would have two binary features indicating whether or not $i,j$ is in $\mathcal{E}_s$ and $\mathcal{E_R}$, and three scalar features, $\tau_{ij}, \tau_{rij}, D_{ij}$.  This edge information is important to designing a good transit network, so we need a neural architecture that will make use of these rich edge features.

For this, we turned to \acrlong{gat}s. A \gls{gat} is a graph neural net architecture that uses a multi-head attention mechanism~\citep{vaswani2017attention} to determine the contribution made to a node's feature by each of its neighbours.  \glspl{gat} can make use of multi-dimensional edge features, and the effect of edge features at each layer on the layer's output is a learned function, rather than a pre-defined one as in graph convolutions.  And they have been shown to perform well on diverse graph-related tasks.  We therefore choose the \gls{gat}v2 architecture proposed in~\cite{gatv2conv} for the layers of our policy net's first stage.  We note that a graph attention net operating on a fully-connected graph has close parallels to a Transformer model~\cite{vaswani2017attention}, but unlike Transformers this architecture enables the use of edge features that describe known relationships between elements.

\nomenclature{$X$}{Used in two ways: First, an $n \times d$ matrix of $d$-dimensional node features for a city graph's nodes.  Second, a set of points corresponding to solutions to a multi-objective problem, where each dimension of the point is the solution's value of one objective.}
\nomenclature{$E$}{a tensor of node-pair features for a graph (equivalently, edge features for a fully-connected graph).}
\nomenclature{$d_e$}{The dimension of the edge features of a graph.}
\nomenclature{$Y$}{A matrix of node embeddings for a graph produced by a \gls{gnn}.}
\nomenclature{$\mathbf{s}_t$}{A vector of global features of the state of an in-progress transit network and city, at timestep $t$ of the network construction \gls{mdp}.}

We refer to this first stage as the ``backbone'' \gls{gat} of the neural net.  It takes a matrix of $n$ node features $X$ and an $n \times n \times d_e$ tensor of edge features $E$ as input, and applies a series of \gls{gat} layers and non-linearities to these.  Each layer of the backbone applies the following operation: 
\begin{align}
H^l = \sigma(\textup{GAT}_{\theta^l}(H^{l-1}, E))
\end{align}
Where $\sigma$ is a non-linear activation function, $H^l$ is the matrix of node embeddings produced by the $l$-th layer, and $H^0 = X$.  The output node embedding matrix $Y$ is the output of the final $L$-th, layer, $H^L$.  Along with the edge features $E$ and a global state vector $\mathbf{s}$, these output node embeddings are given as input to the two policy heads.
 
\subsection{Halting Head}

The halting module $\textup{NN}_{halt}$ is a simple \gls{mlp}.  It takes as input a vector formed by concatenating the following elements:
\begin{itemize}
    \item the node embeddings of first and last nodes on the current route $r_t$; or if $|r_t| = 0$, a placeholder vector whose contents are learned during training
    \item the mean of the node embeddings over the graph $\frac{1}{n}\sum_i y_i$,
    \item a vector $\mathbf{s}_t$, detailed in \autoref{subsec:nn_feats}, that encodes some global information about the state $s_t$,
    \item the driving time of the route being planned, $\tau_{r_t}$.
\end{itemize}

The \gls{mlp} outputs a scalar $z$.  We treat this scalar as the log-probability of halting, and apply the sigmoid function to $h$ get the halting probability: 
\begin{equation}
\pi_\theta(\textup{halt}|s) = \sigma(z) = \frac{1}{1 + e^{-z}}, \; \pi(\textup{continue}) = 1 - \sigma(z)
\end{equation}

\subsection{Extension Head}\label{subsec:extension_head}

\nomenclature{$o_{ij}$}{A score for a pair of nodes $i,j$ in a city graph.}
\nomenclature{$o_a$}{A score for a candidate shortest-path $a$ at timestep $t$ of an \gls{mdp}.}
\nomenclature{$o_{(r_t|a)}$}{A score for a candidate shortest-path extension $a$ to the route $r_t$ being planned at timestep $t$ of an \gls{mdp}.}

The action space of all non-overlapping shortest paths starting from a given node $i$ can be very large.  To make learning a policy over this large action space tractable, we found it necessary to ``factorize'' the computation of probabilities for each shortest-path action.  The extension head of our policy net thus outputs a scalar score $o_{ij}$ for each node pair $(i,j)$ in the graph at each step.  From these, a score for each extension $o_a$ is computed by summing over $o_{ij}$ for all $(i,j)$ that would become directly connected by appending path $a$ to the current route $r_t$.



This should be a function of the time between nodes along the path, $\tau_{aij}$, as well as the demand between $i$ and $j$, and potentially of other features of $\mathcal{G}$ and $\mathcal{G_R}$.  But since the paths are all shortest paths, $\tau_{aij} = T_{ij}$ in every case, so the score $o_{aij}$ gained by linking nodes $i,j$ is independent of the chosen path $a$: $o_{aij} = o_{ij}$.  Let these be computed by some function $f(\cdot)$:
\begin{align}
    o_{ij} = 
\begin{cases}
f(T_{ij}, i, j, \mathcal{G}, \mathcal{G_R}), & \text{if} i \neq j \\
0, & \text{if} i = j
\end{cases}
\end{align}

Assuming these node-pair scores are an accurate indication of the benefit of linking those nodes, the score of the path as a whole should simply be the sum of scores of the node pairs it links:
\begin{align}\label{eqn:o_a}
o_a = \sum_{k \in a}\sum_{l \in a} o_{kl}
\end{align}

Then, suppose we have a partially-planned route $r$ and are considering paths $a \in \textup{SP}$ to extend $r$.  Each of these paths still links all the nodes along it, so can achieve score $o_a$.  But appending $a$ to $r$ will also link every node in $r$ to every node in $a$, so the benefit of extending $r$ by $a$ should depend on these as well.  However, the time taken to get from node $i \in r$ to node $j \in a$ may be greater than $T_{ij}$, since $r|a$ is not necessarily a shortest path.  So the shortest-path node-pair scores $o_{ij}$ here are not appropriate.  Instead, new node-pair scores $o_{(r|a)ij}$ should be computed based on the driving time $\tau_{(r|a)ij}$ that would result from joining $a$ to $r$.  Otherwise, the same function $f(\cdot)$ should be appropriate, with the new inter-node time:

\begin{align}
    o_{(r|a)ij} = 
\begin{cases}
f(\tau_{(r|a)ij}, i, j, \mathcal{G}, \mathcal{G_R}), & \text{if} \; i \neq j \\
0, & \text{if} \; i = j
\end{cases}
\end{align}

And the overall score should then be:
\begin{align}\label{eqn:o_ra}
    o_{(r|a)} = o_a + \sum_{i \in r} \sum_{j \in a} o_{(r|a)ij}
\end{align}

We design the extension policy head to compute the score function $f(\cdot)$.  In this policy head, the function $f(\cdot)$ that computes node-pair scores is an \gls{mlp}, called $\text{MLP}_{ext1}$, that takes as input $T_{ij}$ or $\tau_{(r|a)ij}$ (as appropriate), the backbone GAT's embeddings $y_i$ and $y_j$ of nodes $i$ and $j$, and the global state vector $\mathbf{s}$.  For each path $a \in \textup{SP}$, we sum these node-pair scores to get path scores $o_a$ as in \autoref{eqn:o_a}, and when extending a route $r$, we compute the extension scores $o_{(r|a)}$ as in \autoref{eqn:o_ra}.  We can efficiently compute the node-pair scores in a batch by pairing all node embeddings $y_i$ when $r_t = []$; and when $r_t \neq []$, we only compute scores $o_{(r|a)ij}$ for node pairs where exactly one of $i$ or $j$ is in $r_t$.  The resultant scalar scores, $o_a$ and $o_{(r|a)}$, are provided to the next stage of the extension policy head.

\nomenclature{$\hat{o}_a$}{The score computed by our neural net policy for action $a$.}

If $\alpha = 0.0$, then the benefit of choosing a path $a$ to start or extend a route depends on its driving time $\tau_a$, and generally on the state $s$.  And if $0 < \alpha < 1$, then the benefit of $a$ depends on these as well as on the edges it adds to $\mathcal{E}_\mathcal{R}$, and thus on $o_{(r|a)}$.  So our architecture ought to incorporate all of these factors.  We achieve this with a final \gls{mlp}, called $\text{MLP}_{ext2}$, that takes as input $\mathbf{s}_t, \tau_a,$ and the previous node-pair-based score ($o_a$ or $o_{(r|a)}$, depending on whether $r_t = []$), and outputs a final scalar score $\hat{o}_a$ for each candidate path.  The extension policy is then the softmax of these values:

\begin{equation}
\pi_\theta(a|s) = \frac{e^{\hat{o}_a}}{\sum_{a' \in \mathcal{A}} e^{\hat{o}_{a'}}}
\end{equation}

We thus treat the outputs of $\textup{NN}_{ext}$ as un-normalized log-probabilities of the possible actions during an extension step.

\nomenclature{$\textup{NN}_{halt}$}{The module of the neural net policy which makes halting decisions.}
\nomenclature{$\textup{NN}_{ext}$}{The module of the neural net policy which makes extension decisions.}

\subsection{Features and Net Architecture Details}\label{subsec:nn_feats}

The policy net operates on three inputs: an $n \times 4$ matrix of node feature vectors $X$, an $n \times n \times 13$ tensor of edge features $E$, and a global state vector $\mathbf{s}$.

A node feature $x_i$ is a vector with four elements: the $(x,y)$ spatial coordinates of the node, and its in-degree and out-degree in the street graph.

An edge feature $\mathbf{e}_{ij}$ is composed of the following elements:
\begin{itemize}
    \item $D_{ij}$, the demand between the nodes
    \item $s_{ij} = 1$ if $(i,j,\tau_{ij}) \in \mathcal{E}_s$, $0$ otherwise
    \item $\tau_{ij}$ if $s_{ij} = 1$, $0$ otherwise    
    \item $c_{ij} = 1$ if $\mathcal{R}$ links $i$ to $j$, $0$ otherwise
    \item $c_{0ij} = 1$ if $j$ can be reached from $i$ over $\mathcal{R}$ with no transfers, $0$ otherwise
    \item $c_{1ij} = 1$ if one transfer is needed to reach $j$ from $i$ over $\mathcal{R}$, $0$ otherwise
    \item $c_{2ij} = 1$ if two transfers are needed to reach $j$ from $i$ over $\mathcal{R}$, $0$ otherwise
    \item $\textup{self} = 1$ if $i = j$, $0$ otherwise
    \item $\tau_{\mathcal{R}ij}$ if $c_{ij} = 1$, $0$ otherwise
    \item $\tau_{rij}$ if $c_{0ij} = 1$ where $r$ the route that provides the shortest direct trip between $i$ and $j$, $0$ otherwise    
    \item $T_{ij}$, the shortest-path driving time between the nodes
    \item $\alpha$ and $1-\alpha$, the weights of the two components of the cost function $C$
\end{itemize}

\nomenclature{$\mathbf{e}_{ij}$}{A vector of features for the pair of nodes $i,j$.}

The global state vector $\mathbf{s}$ is composed of:
\begin{itemize}
    \item Average passenger trip time given current route set, $C_p(\mathcal{G}, \mathcal{R})$
    \item Total route time given current route set, $C_o(\mathcal{G}, \mathcal{R})$
    \item number of routes planned so far, $|\mathcal{R}|$
    \item number of routes left to plan, $S - |\mathcal{R}|$
    \item fraction of node pairs with demand $d_{ij} > 0$ that are not connected by $\mathcal{R}$
    \item $\alpha$ and $1-\alpha$, the weights of the two components of the cost function $C$
\end{itemize}

Note that $\alpha$ and $1-\alpha$ are included in both $\mathbf{s}$ and $\mathbf{e}_{ij}$.

\nomenclature{$\mu$}{The empirical mean of a set of numbers.}
\nomenclature{$\sigma$}{The empirical standard deviation of a set of numbers.}

Neural net training is sensitive to extreme values in the inputs, so to stabilize training, it is common practice to normalize the inputs.  We here do this by running the policy network over the validation set before training it.  We compute the means $\mu$ and standard deviations $\sigma$ of the numerical input feature values observed during this initial validation step, and store these as parameters of the policy network.  Then, during training and when later evaluating the policy network, we normalize all numerical input features by subtracting $\mu$ and dividing by $\sigma$, so the resulting normalized inputs are centred near 0 and have standard deviation near 1.  Boolean-valued features are not normalized, but provided as raw binary values.

We made use of the ReLU non-linearity~\citep{fukushima1969visual} as the activation function throughout our policy net, but we note that we experimented with other activation functions such as GELU~\citep{hendrycks2016gaussian}, and found that this made little difference to performance.  Unless otherwise specified, each neural net layer of each component outputs vectors of the common embedding dimension $d_{embed}$.  \autoref{tab:net_params} shows the hyperparameters we used for the policy net architecture.  These were arrived at by a manual search.

\nomenclature{$d_{embed}$}{The embedding dimension of the intermediate activations in our neural nets.}

\begin{table}
	\centering
	\begin{tabular}{lr}
		\toprule
		Hyper-Parameter & Value \\
		\midrule    
		\# backbone layers & 5 \\ %
		\# heads in \gls{gat} multi-head attention & 4 \\
		embedding dimension $d_{embed}$ & 64 \\
		\# layers of $\text{MLP}_{ext1}$  & 3 \\
		\# layers of $\text{MLP}_{ext2}$ & 3 \\
		hidden layer dimension of $\text{MLP}_{ext2}$ & 16 \\
		\# layers of $\text{NN}_{halt}$ & 3 \\
		\bottomrule
	\end{tabular}
	\caption{Policy net architectural hyper-parameters}
	\label{tab:net_params}
\end{table}

\section{Training The Policy Net}\label{sec:training}

In our prior work \citep{holliday2023augmenting, holliday2024autonomous}, we trained the policy net using the REINFORCE with Baseline method proposed by~\cite{williams1992reinforce}, using a reward function that was 0 at all steps except the final step, where it was the negative of the cost function.  This was inspired by the similar approach taken by~\cite{Kool2019AttentionLT}.  However, REINFORCE has been been built upon by a number of other policy gradient learning algorithms since its publication.  Also, sparse reward functions like this one, which is zero at most timesteps, convey little information about the effect of each action, and so can make learning difficult.

\nomenclature{$A_t$}{The advantage function computed at timestep $t$ of a training episode in \gls{ppo}.}

For these reasons, in this thesis we train the policy net using the \gls{ppo} algorithm and a more dense reward function, described in \autoref{subsec:reward}.  \gls{ppo} is a policy gradient method proposed by~\cite{schulman2017PPO}, because of its relative simplicity and because it remains a state-of-the-art method on many problems.  \gls{ppo} works by rolling out a number of steps $h$ of a ``batch'' of episodes in parallel, and then computing an ``advantage'' for each timestep, $A_t = \sum^{h-1}_{t'=0} \gamma^{t'} R_{t+t'} + \gamma^h V(s_{t+h}) - V(s_t)$.  This advantage weights the updates to the policy, such that actions with a positive advantage are made more likely, and those with a negative advantage are made less so. Updates to the policy are also weighted by the ratio of the current policy's probability for the chosen action and that of the policy that actually chose the action, to avoid the divergence that can plague off-policy learning.  This is accomplished by attempting to maximize the CLIP objective defined by \cite{schulman2017PPO}:
\begin{equation}
L^{CLIP}(\theta) = \mathbb{E}_t \left [ \min \left (
	\frac{\pi_\theta(a_t|s_t)}{\pi_{\theta_{\textup{old}}}(a_t|s_t)}A_t,
	\textup{clip} \left (
		\frac{\pi_\theta(a_t|s_t)}{\pi_{\theta_{\textup{old}}}(a_t|s_t)}, 
		1-\epsilon, 
		1+\epsilon \right )
	A_t \right ) \right ]
\end{equation}

Where $\theta_{\textup{old}}$ are the policy parameters prior to the updates made in this training iteration.  The ``clip($x,y,z$)'' function returns $y$ if $x<y$, $z$ if $x>z$, and $x$ otherwise.  $\epsilon$ is a parameter usually set to a value in the range $[0, 0.5]$.  Unlike \cite{schulman2017PPO}, we do not include any entropy-maximization term in our training objective, as we found that doing so worsened the performance of the resulting policies.

\nomenclature{$V(s)$}{A value function that estimates the expected return achieved by some policy $\pi$ from state $s$ in an \gls{mdp}.}
\nomenclature{$G^h_t$}{The discounted sum of rewards from timestep $t$ to timestep $t+h$ in an \gls{mdp}.}
\nomenclature{$\mathbf{x}_i$}{A vector of features of street graph node $i$.}

The value function $V(s_t)$ is itself a small \gls{mlp} neural net that is trained in parallel with the policy to predict the discounted return $G^h_t = \sum^{h-1}_{t'=1} \gamma^{t'} R_{t+t'} + \gamma^h V(s_{t+h})$.  The input to $V(s_t)$ is a vector composed of $\alpha$ and several statistics of $\mathcal{G}$: the average of the node features $\frac{\sum_{i}\mathbf{x}_i}{n}$, the total demand $\sum_{i,j} D_{ij}$, the means and standard deviations of the elements of $D$ and $T$, and the cost component weights $\alpha$ and $1-\alpha$.  Along with these, it takes the state vector $\mathbf{s}_t$.

After each batch of training episodes, the mean squared error of the value net's estimates of the advantage $V(s_t)$ against the actual advantages $A_t$:

\nomenclature{$Q$}{The size of the mini-batch used in \gls{ppo}.}
\begin{align}
\text{Loss}_\text{value} = \frac{1}{Q} \sum_{i \in Q} [A_t - V(s_t)]^2
\end{align}
where $Q$ is the minibatch size.

We experimented with using an \gls{mlp} that would compute the value estimate $V(s_t)$ based on the same node embeddings $Y$ as $\textup{NN}_{ext}$ and $\textup{NN}_{halt}$, but found that this performed slightly worse than using the separate \gls{mlp}.

\subsection{Reward Function}\label{subsec:reward}

We use a reward function that is based on how the cost of the partial network $\mathcal{R}_t\cup \{r_t\}$ changes with each timestep.  Specifically, we define the reward as:
\begin{equation}
R_t = C'(\mathcal{G}, \mathcal{R}_t \cup \{r_t\}) - C'(\mathcal{G}, \mathcal{R}_{t+1} \cup \{r_{t+1}\})
\end{equation}

Note that this means that on even timesteps (when halt-or-continue actions are taken), $R_t = 0$ because $\mathcal{R}_t$ and $r_t$ do not change as a result of a halt-or-continue action.

\nomenclature{$\delta_{\mathcal{R}ij}$}{A delta function that takes value 1 if $\mathcal{R}$ provides a path from $i$ to $j$, and 0 if not.}
\nomenclature{C'}{A modified cost function used to compute the reward when training neural net policies.}

In the above equation, $C'$ is a modified cost function that we define to serve as a good basis for this reward, specifically:
\begin{align}
C'_p(\mathcal{G}, \mathcal{R}) &= \frac{\sum_{i,j} D_{ij}(\delta_{\mathcal{R}ij}\tau_{\mathcal{R}ij} + (1-\delta_{\mathcal{R}ij})2\max_{kl} T_{kl})}{\sum_{i,j} D_{ij}} \\
C'_c(\mathcal{G}, \mathcal{R}) &= F_{un} + 0.1 \delta'_v \\
C'(\mathcal{G}, \mathcal{R}) &= \alpha w_p C'_p(\mathcal{G}, \mathcal{R}) + (1 - \alpha) w_o C_o(\mathcal{G}, \mathcal{R}) - \beta C'_c
\end{align}

$\delta_{\mathcal{R}ij}$ is a delta function that takes value 1 if $\mathcal{R}$ provides a path from $i$ to $j$, and 0 if not.  $\delta'_v$ takes value 1 if $F_{un} > 0$, and 0 otherwise.  The second term in the numerator of $C'_p$ is meant to represent the high cost of failing to satisfy some demand, in a way that is proportional to the amount of demand that is unsatisfied, unlike $F_{un}$ which reflects only how many node pairs are unconnected.  Each desired trip that cannot be made over network $\mathcal{R}$ is treated as taking time equal to twice the diameter of the street graph.  

If $\mathcal{R}$ satisfies the connectedness constraint of \autoref{subsec:constraints}, it connects all pairs of nodes, so $C'_p$ simplifies to $C_p$.  But during the \gls{mdp} rollout, it is necessarily the case that the connectedness constraint is violated at some timesteps, before the network is complete.  In these cases, $C_p$ may increase from one timestep to the next as far-apart node pairs that were previously unconnected become connected by a new action.  This would tend to penalize connecting those routes by increasing overall cost, when what we want is to reward the connection of new node pairs.  We use $C'_p$ instead because it decreases when a new node pair becomes connected, unless the connecting route is improbably circuitous.

It could be argued that simply increasing $\beta$ would accomplish the same, but in practice we found that this was not the case: training with $C'_p$ instead of $C_p$ yielded better-performing models regardless of $\beta$'s value.

Similarly, we use $C'_c$ instead of $C_c$ because by the structure of the \gls{mdp}, the number-of-stops constraint of \autoref{subsec:constraints} is guaranteed to be satisfied when the \gls{mdp} terminates, but before it terminates this will often not be the case.  Penalizing the policy for these temporary violations of the number-of-stops constraint serves no purpose, and may swamp out useful learning signals.  So we construct $C'_c$ to be independent of the number-of-stops constraint 3, unlike $C_c$.

\subsection{Training Data}\label{subsec:training_data}

To train the policy, we generate our own dataset of synthetic city graphs.  The generation process begins by generating the nodes and street network using one of these processes chosen at random:
\begin{itemize}
	\item 4-grid: Place $n$ nodes in a rectangular grid as close to square as possible.  Add edges from each node to its horizontal and vertical neighbours.
	\item 8-grid: The same as the above, but also add edges between diagonal neighbours.
	\item $4$-nn: Sample $n$ random 2D points uniformly in a square to give $\mathcal{N}$.  Add street edges to each node $i$ from its four nearest neighbours.
	\item Minimum spanning tree: Sample $n$ points as above, treat them as a fully-connected graph $J$ with edge weights as euclidean distances between points, and find the minimum spanning tree (MST) $G$ of $J$.  Then add more edges from $J$ to $G$, in order of increasing weight, until a desired number of edges $e$ is reached, where $e$ is a linear function of $n$.
	\item Voronoi: Sample $m$ random 2D points, and compute their Voronoi diagram~\cite{fortune1995voronoi}.  Take the shared vertices and edges of the resulting Voronoi cells as $\mathcal{N}$ and $\mathcal{E}_s$.  $m$ is chosen so $|\mathcal{N}| = n$.    
\end{itemize}

We note that the Minimum-spanning-tree process is the same process used to generate the Mumford benchmark cities \citep{mumford2013dataset}, discussed further in \autoref{subsec:nn_eval}.

\nomenclature{$\rho$}{Probability of deleting a street edge when generating city graphs for the training set.}
\nomenclature{$x_i$}{The x-coordinate in the 2D plane of street graph node $i$}
\nomenclature{$y_i$}{The y-coordinate in the 2D plane of street graph node $i$}

For each process except Voronoi, each edge in $\mathcal{E}_s$ is then deleted with user-defined probability $\rho$.  If the resulting street graph is not strongly connected, it is discarded and the process is repeated.  Nodes are sampled in a $30 \textup{km} \times 30 \textup{km}$ square, and a fixed vehicle speed of $v = 15 \textup{m/s}$ is assumed to compute street edge weights $\tau_{ij} = ||(x_i, y_i) - (x_j, y_j)||_2 / v$.  Finally, we generate the \gls{od} matrix $D$ by setting diagonal demands $D_{ii} = 0$ and uniformly sampling off-diagonal elements $D_{ij}, i \neq j$ in the range $[60, 800]$.

The five different graph types were chosen to give a variety of spatial graphs that would reflect the variety in types of spatial layouts among real cities, the layouts of which are sometimes grid-like, other times more haphazard and random-seeming, but always with nodes connecting mostly to other nearby nodes.  The demand-sampling process was chosen to match that used to generate the Mumford benchmark cities, as these were our main evaluation target; this process also has the virtue of being very simple, although it loses the spatial structure of demand that can be present in real cities.

Our dataset consisted of $2^{15} = 32,768$ of these synthetic cities, each with $n=20$ nodes.  We make a 90:10 split of this dataset into training and validation sets.

\nomenclature{$C_v$}{The average cost function value for a set of transit networks on the city graphs in the validation set computed during training.}

In each iteration of training, we run learned construction with the policy net on one full \gls{mdp} episode over each city in a ``batch'' that is sampled from the training set, and update the policy and value nets according to the \gls{ppo} learning algorithm.  After every ten iterations, the policy is run on the entire validation set, and the average cost $C_v$ of the generated transit networks on the validation set is recorded.  At the end of training, the parameters $\theta$ from with the lowest $C_v$ are returned, giving the final policy $\pi_\theta$.

In each episode of training, we sample a different cost weight $\alpha$ in the range $[0, 1]$ for each city, so that the policy net can learn to adapt its policy to the user's preference for optimizing the passenger versus operator costs.
$S, m_\text{min}, m_\text{max}$, and the constraint weight $\beta$ are held constant across training.  The values used for these parameters, and the sampling process for $\alpha$, are presented in \autoref{tab:learning_params}.

All neural net inputs are normalized so as to have unit variance and zero mean across the entire dataset during training.  The scaling and shifting normalization parameters are saved as part of the policy net and are applied to new data presented to $\pi_\theta$ after training.

\subsection{Training Parameters}

Training was performed using the well-known Adam optimizer~\citep{kingma2015adam}.  The value function estimator used during training was an \gls{mlp} with 3 layers and hidden-layer dimension 36, which is 2 times its input dimension.

In each epoch of training, we augmented the training dataset by applying a set of random transformations.  These included rescaling the node positions by a uniformly-sampled random factor uniformly sampled in the range $[1-a, 1+a]$, rescaling the demand magnitudes by a random factor uniformly sampled in the range $[1-b, 1+b]$, mirroring node positions about the $y$ axis with probability $0.5$, and rotating the node positions by a random angle in $[0, 2\pi)$ about their geometric centre.

We also randomly sampled $\alpha$ separately for each training city in each batch.  Each time $\alpha$ is sampled, it had equal probability of being set to $0$, set to $1$, and sampled uniformly in the range $[0,1]$, so as to encourage the policy to learn to handle both extreme and intermediate $\alpha$ values.

This data augmentation lends greater variety to the finite dataset, increasing the performance of the resulting policy on new environments not seen during training.

\autoref{tab:learning_params} gives the hyper-parameters used during training.  To find the values for decay and learning rate for the policy net, and node position and demand scaling ranges $a$ and $b$, we ran an automatic hyperparameter search based on tree-structured Parzen estimation~\citep{bergstra2013making} using the Optuna library~\citep{optuna_2019}.  Other parameters were set through a manual search or based on recommended values from the literature.

\nomenclature{$\epsilon$}{Used in two ways: First, to denote the threshold used in \gls{ppo}'s CLIP function.  Second, to denote a small constant value used in computing the hypervolume of a set of points.}

\begin{table}[]
    \centering
    \begin{tabular}{lr}
        \toprule
		Hyper-Parameter & Value \\
		\midrule
		Value function learning rate & $5 \times 10^{-4}$ \\
		Value function weight decay & 0.01 \\
		Policy learning rate & 0.0016 \\
		Policy weight decay & $8.4 \times 10^{-4}$ \\
		Number of training iterations & 200 \\
		Discount rate $\gamma$ & 0.95 \\
		PPO CLIP threshold $\epsilon$ & 0.2 \\
		Batch size & 256 \\
		Horizon & 120 \\
		Num. epochs & 1 \\
		Constraint weight $\beta$ & 5.0 \\
		$S$ & 10 \\
		$m_\text{min}$ & 2 \\
		$m_\text{max}$ & 12 \\
		Adam $\alpha$ & 0.001 \\
		Adam $\beta_1$ & 0.9 \\
		Adam $\beta_2$ & 0.999 \\
		\bottomrule
    \end{tabular}
    \caption{Training Hyper-Parameters}
    \label{tab:learning_params}
\end{table}

Training proceeds for 200 iterations in batches of 256 cities, as we found that the policy stopped improving after this.  A total of 51,200 episodes are rolled out during training.  \gls{ppo} has a parameter for the number of ``epochs'' per batch, that is, the number of training passes it makes over a batch of state transitions.  Repeatedly iterating over the same batch can make training more efficient when computing the rewards requires costly simulation.  But in our case, computing the rewards is much less costly than the forward and backward passes of the policy net itself, so we set the number of \gls{ppo} epochs per batch to 1 so that more diverse episodes will be seen across training.

\nomenclature{$h$}{The episode rollout horizon used in \gls{ppo}.}
The episode horizon for rollouts $h$ is set to 120 because we find that with our settings of $n, S, m_\text{min}, m_\text{max}$, this meant that virtually every training episode would terminate within $h$ steps.  When an episode terminated in $t' < h$ steps, that episode was restarted from scratch with the same city $\mathcal{G}$ and $\alpha$ and rolled out again, so that $h$ distinct steps were recorded for each $\mathcal{G}, \alpha$ pair in the batch.

The search we performed for the value of $n$ deserves some discussion.  We experimented with training datasets of cities with $n=20$, $50$, and $100$, with $S, m_\text{min}, m_\text{max}$ being increased in proportion to $n$.  The time and memory requirements of training for a fixed number of epochs grew considerably with $n$.  At $n=100, S=50$, this pushed the limits of what our commercial desktop computer could handle.  To evaluate the different policies, we ran them on two of the benchmark cities described in \autoref{subsec:nn_eval}: the smallest, with $n=15$, and the largest, with $n=127$.  

We were surprised to find that the policy trained on $n=20$ performed best on both the $n=15$ and $n=127$ benchmark graphs, while increasing the training set $n$ to $50$ and then to $100$ made the policy successively worse on both benchmark graphs.  This goes against our expectations that increasing $n$ in the training set ought to produce a policy that performs better at larger $n$.  Based on these results, we chose keep $n=20$ in our training process, and did not investigate this mystery further, but it does warrant further inquiry.  In particular, it would be interesting to train a model with $n$ starting at 20 and gradually increasing.

\section{Experiments}\label{sec:nn_results}

\subsection{Training}\label{subsec:nn_training_results}

\begin{figure}[]
  \centering
  \includegraphics[width=0.8\linewidth]{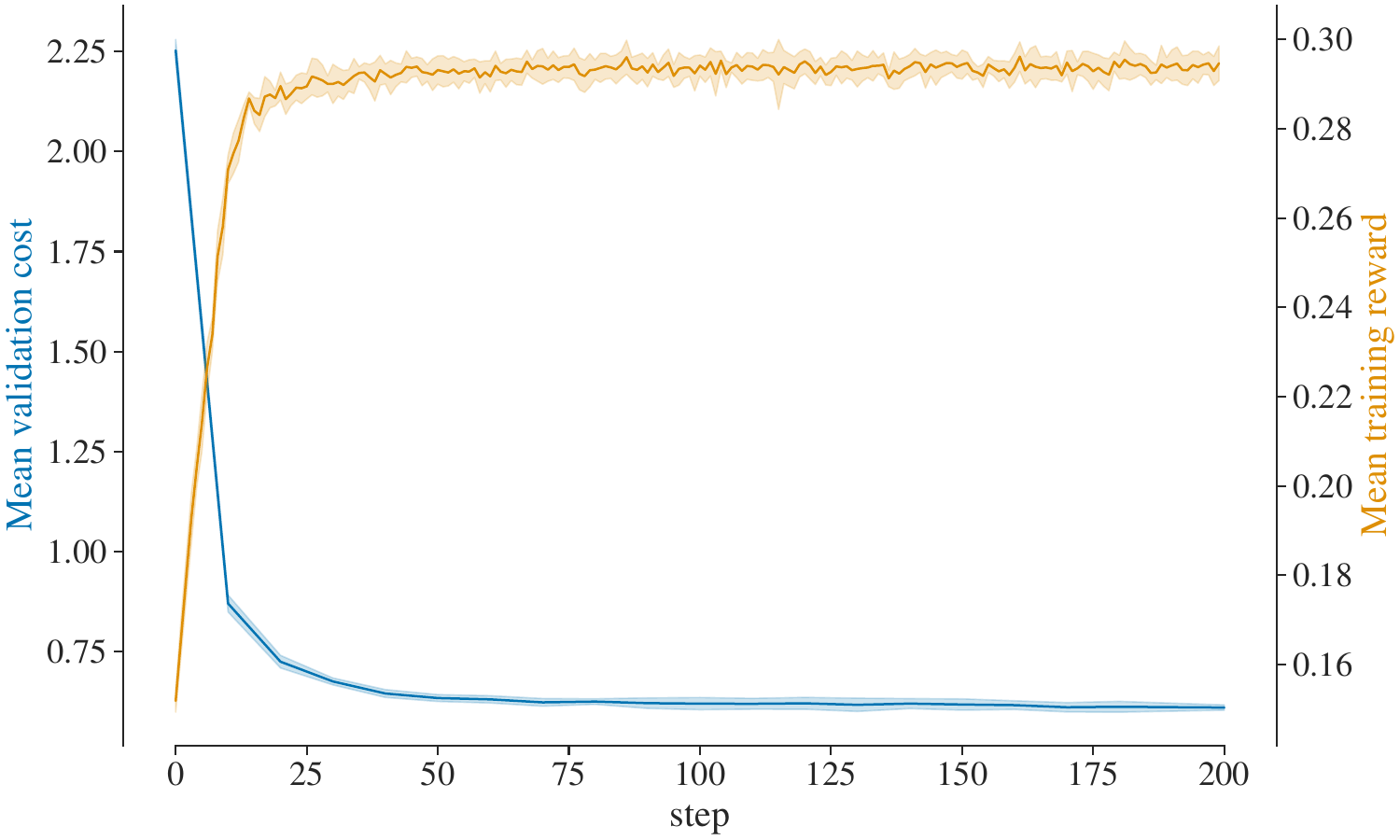}  
  \caption{Mean cost on the validation set, and mean reward during training, achieved by our policy net over five epochs of training.  The curves show the mean over ten training runs with different random seeds, and the shaded areas shows one standard deviation over the ten seeds.}
  \label{fig:training_curves}
\end{figure}

In these experiments, we trained ten different policy net models $\pi_{\theta_k}$ with ten different random seeds $k$ using the procedure described in \autoref{sec:training}.  \autoref{fig:training_curves} shows the cost achieved on the validation set and the reward over each training episode averaged over these ten runs.  The costs achieved by the policy decreases rapidly over the first 50 training batches.  Improvement then slows considerably, and has levelled off by about 200 batches, at which point we halt training.  Over the course of training, the cost on the validation set drops by about 70\%.

We see also that the validation cost and the training rewards follow the mirrored trends over training.  This is plausible, given that the two sets are sampled from the same distribution of synthetic cities.  It shows that the policy net is not simply memorizing the best actions to perform on the specific examples in the training set, but is learning some policy that is effective across city graphs from this distribution.

\subsection{Evaluation}\label{subsec:nn_eval}

We evaluate the trained policies on five benchmark cities from two different benchmarks: Mandl~\cite{mandl1980evaluation}, and Mumford~\cite{mumford2013new}.  Both benchmarks are widely used to evaluate \gls{ndp} algorithms~\cite{john2014routing, kilic2014demand, husselmann2023improved}.  The Mandl dataset is one small synthetic city, while the Mumford dataset consists of four synthetic cities, labelled Mumford0 through Mumford3, and it gives values of $S$, $m_\text{min}$, and $m_\text{max}$ to use when benchmarking on each city.  The values $n$, $S$, $m_\text{min}$, and $m_\text{max}$ for Mumford1, Mumford2, and Mumford3 are taken from three different real-world cities and their existing transit networks, giving the dataset a degree of realism.  Details of these benchmarks, and the parameters we use when evaluating on them, are given in \autoref{tab:dataset}.

\begin{table*}
	\caption{Statistics of the five benchmark problems used in our experiments.  The area is estimated assuming all vehicles travel at 15 meters per second.} 
	\label{tab:dataset}
	\centering
	\begin{tabular}{lcccccc}
		\toprule
		City & \# nodes $n$ & \# street edges $|\mathcal{E}_s|$ & \# routes $S$ & $m_\text{min}$ & $m_\text{max}$ & Area (km$^2$) \\
		\midrule
		Mandl   & 15  & 20  & 6  & 2  & 8 & 352.7 \\
		Mumford0 & 30  & 90  & 12 & 2  & 15 & 354.2 \\
		Mumford1 & 70  & 210 & 15 & 10 & 30 & 858.5 \\
		Mumford2 & 110 & 385 & 56 & 10 & 22 & 1394.3 \\
		Mumford3 & 127 & 425 & 60 & 12 & 25 & 1703.2 \\
		\bottomrule
	\end{tabular}
\end{table*}

As noted in \autoref{subsec:cost_function}, in all of our experiments, we set the transfer penalty used to compute average trip time $C_p$ to $p_T = 300$s (five minutes), the typical value used these benchmarks~\citep{mumford2013new}.

\begin{figure}
	\centering
	\begin{subfigure}[t]{0.4\linewidth}
		\includegraphics[width=\linewidth]{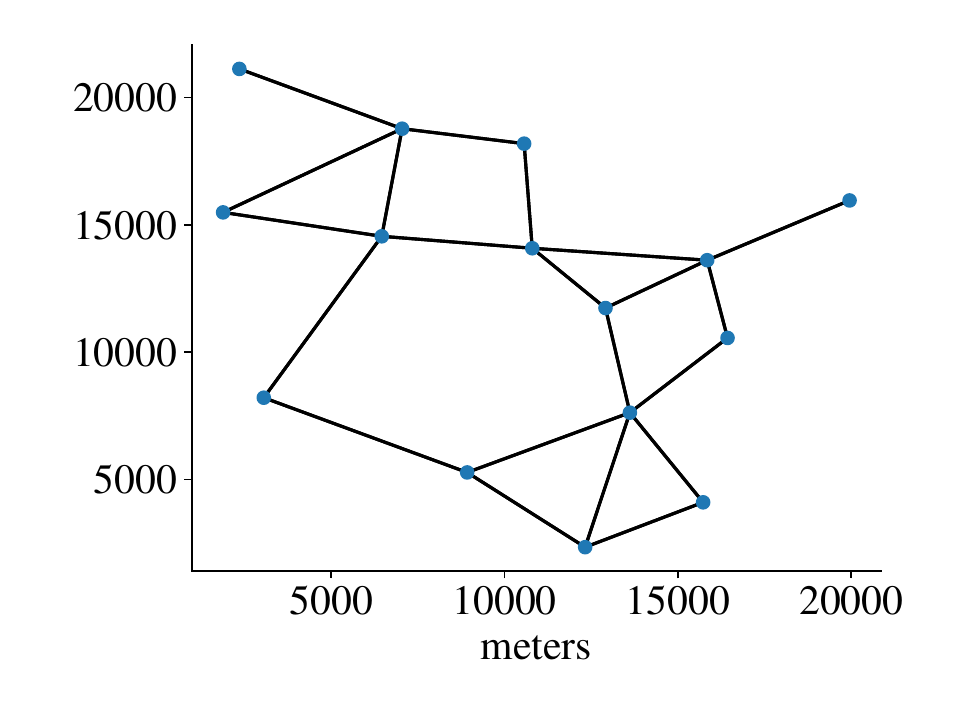}
		\caption{Mandl}
		\label{subfig:mandl}
	\end{subfigure} \\
	\begin{subfigure}[t]{0.4\linewidth}
		\includegraphics[width=\linewidth]{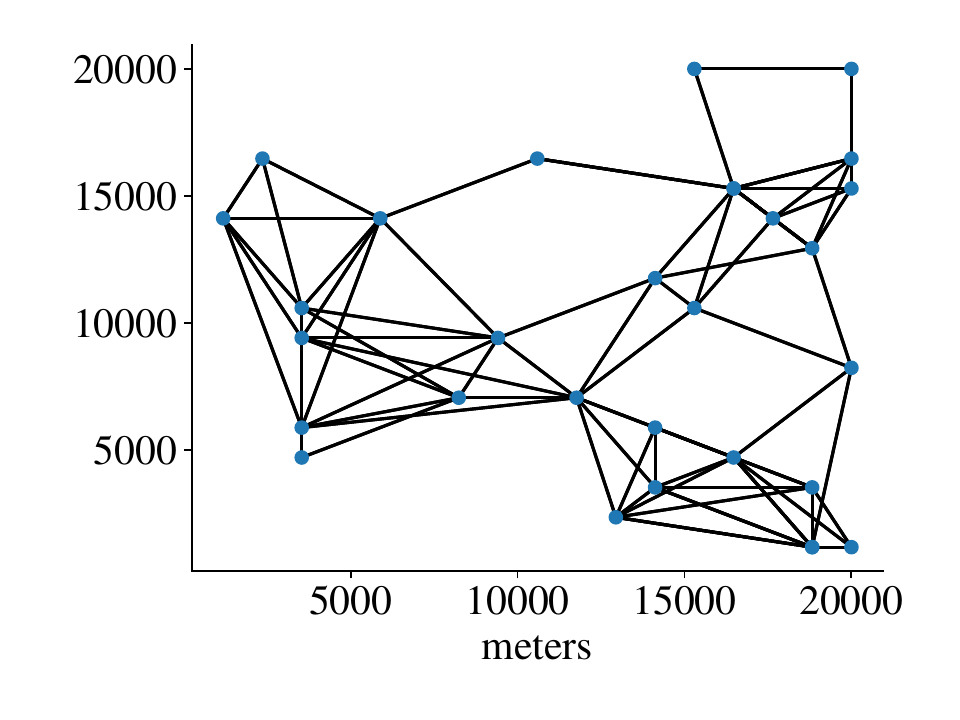}
		\caption{Mumford0}
		\label{subfig:mumford0}
	\end{subfigure}
	\begin{subfigure}[t]{0.4\linewidth}
		\includegraphics[width=\linewidth]{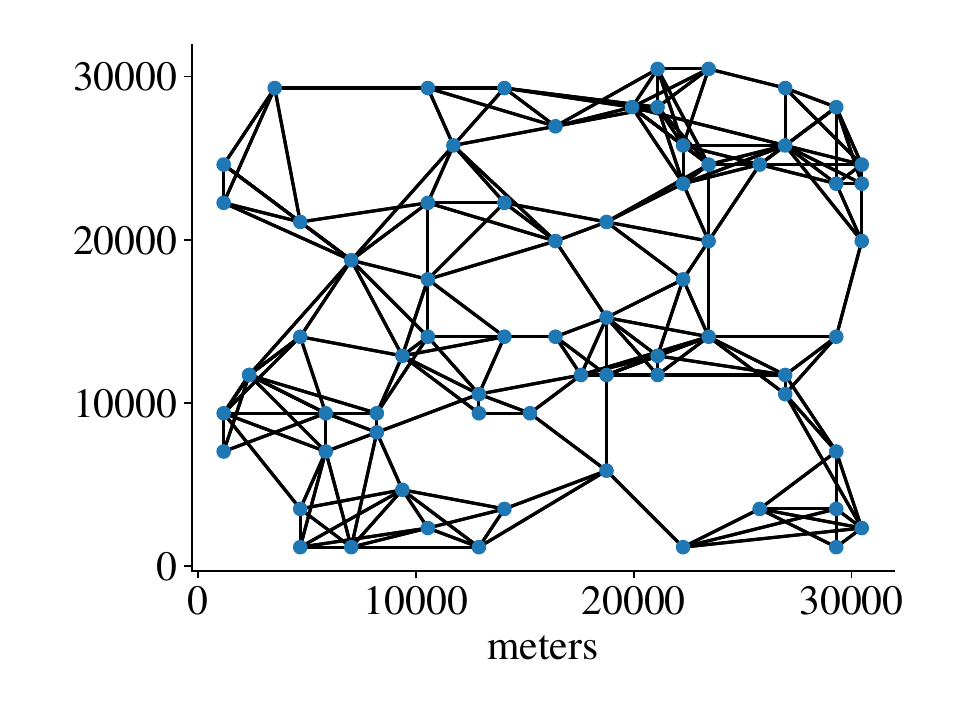}
		\caption{Mumford1}
		\label{subfig:mumford1}
	\end{subfigure} \\ 
	\begin{subfigure}[t]{0.4\linewidth}
		\includegraphics[width=\linewidth]{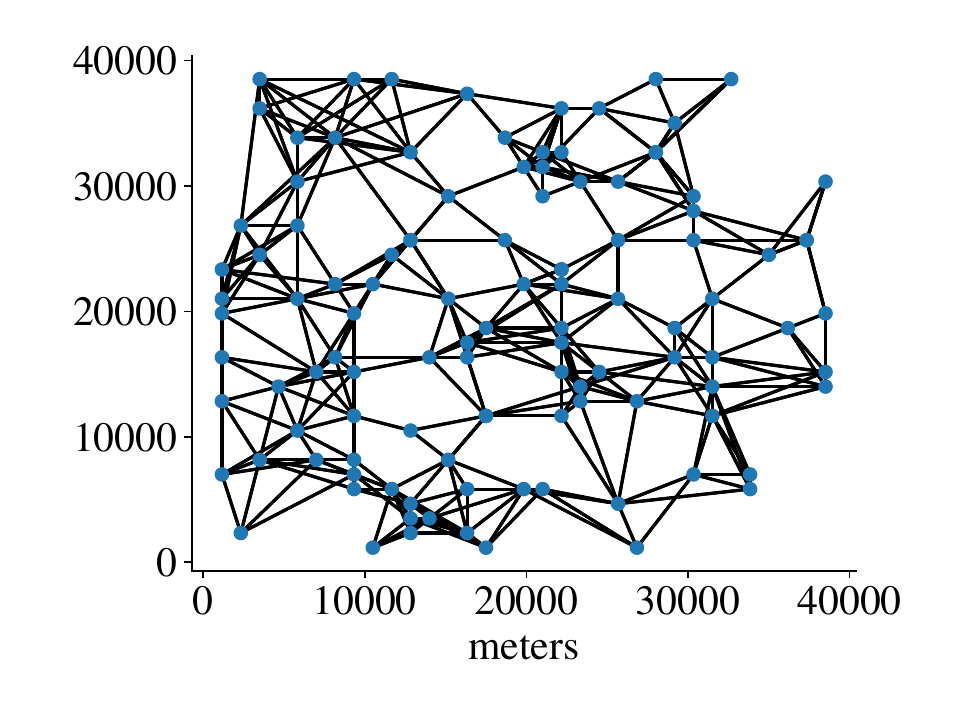}
		\caption{Mumford2}
		\label{subfig:mumford2}
	\end{subfigure}
	\begin{subfigure}[t]{0.4\linewidth}
		\includegraphics[width=\linewidth]{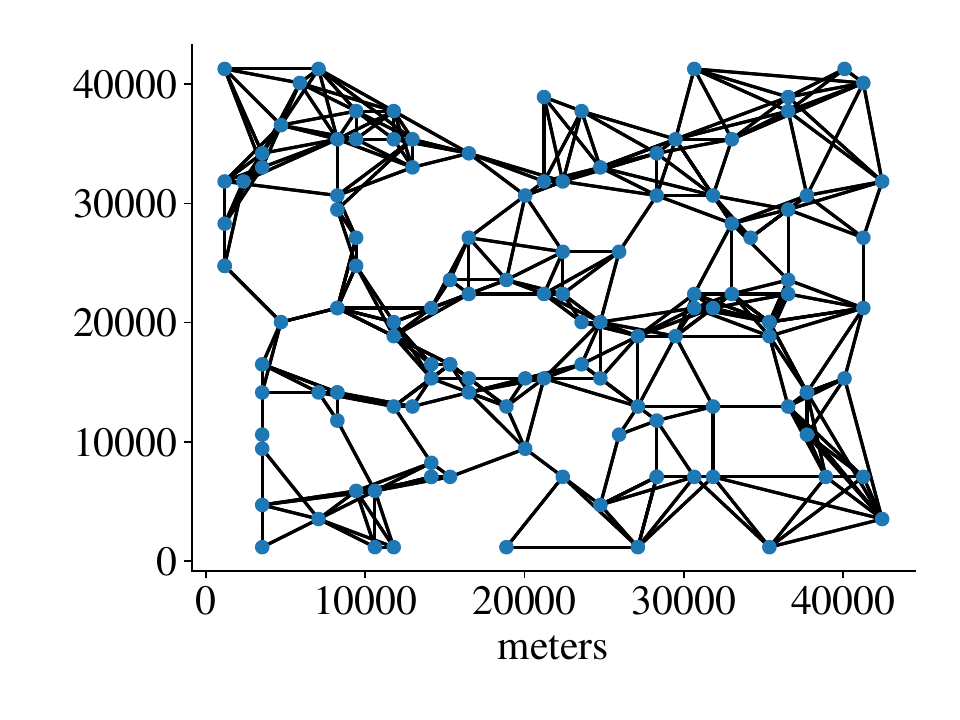}
		\caption{Mumford3}
		\label{subfig:mumford3}
	\end{subfigure}
	\caption{The Mandl benchmark city and the four Mumford benchmark cities.}
	\label{fig:mumford_cities}
\end{figure}

The data files that define both benchmarks are obtained from~\cite{mumford2013dataset}.  These files provide, for each city, the node positions $\mathbf{p}_i$ on an arbitrary scale, as well as the driving time $\tau_{ij}$ in seconds of all street edges and the demand matrix $D$.  To convert the node positions from their arbitrary scale to a scale of meters, we assumed a fixed driving speed for vehicles of 15 meters per second, which equates to 54 kilometers per hour, a reasonable average speed for a city bus.  We used this to compute a scaling factor for the node positions as follows:
\begin{equation}
f = \frac{1}{|\mathcal{E}_s|} \sum_{(i,j,\tau_{ij}) \in \mathcal{E}_s} \frac{\tau_{ij} * 15 \text{m/s}}{||\mathbf{p}_i - \mathbf{p}_j||_2}
\end{equation}
and then multiplied the positions by this factor to get approximate positions in meters $\mathbf{p}'_i = f \mathbf{p}_i$.  \autoref{fig:mumford_cities} shows the Mandl and Mumford benchmark cities based on these scaled node positions $\mathbf{p}'_i$.

\nomenclature{$\mathbf{p}_i$}{The position in the 2D plane of node $i$.}

To evaluate a trained policy on one of these cities $\mathcal{G}$, we sampled 100 networks from the policy by running it for 100 separate episodes on $\mathcal{G}$.  We then chose the network $\mathcal{R}$ among these for which $C(\mathcal{G}, \mathcal{R})$ was lowest.  We refer to this as the LC-100 algorithm, ``LC'' standing for ``learned construction''.  We also separately evaluate each policy ``greedily'' by running one episode on $\mathcal{G}$ where instead of sampling $a_t$ according to $\pi(\cdot | s_t)$, we deterministically take actions $a_t = \text{argmax}_a \pi_\theta(a|s_t)$.  We can also view this as using the policy $\pi_{\theta G}(a|s_t) = 1 \iff a = \text{argmax}_{a'} \pi_\theta(a' | s_t), 0 \text{otherwise}$.  We refer to this as the ``LC-Greedy'' algorithm.

To provide a point of comparison, we also ran experiments on this benchmark with a purely random policy $\pi_\text{random}$, which gives equal probability to all actions: 
\begin{equation}
\pi_\text{random}(a|s_t) = \frac{1}{|\mathcal{A}_t|} \forall a \in |\mathcal{A}_t|
\end{equation}
We followed a similar procedure to ``LC-100'' using this policy, which we refer to as ``RC-100'' for ``random construction''.

The trade-off parameter $\alpha$ in the cost function $C(\mathcal{G}, \mathcal{R})$ reflects the range of possible preferences a user may have over minimizing passenger cost $C_p$ versus operator cost $C_o$.  An \gls{ndp} algorithm should be sensitive to this parameter, producing networks that reflect the user's preference.  To assess this, we ran each algorithm on each benchmark city with eleven values of $\alpha$, ranging from $\alpha = 0.0$ to $\alpha = 1.0$ in increments of 0.1.  Additionally, as LC-100 and RC-100 are both stochastic algorithms, and as the training process is itself stochastic, we repeated each run 10 times with 10 different random seeds $k$ and, if a learned policy was needed, we used the parameters $\theta_k$ trained with random seed $k$.

\begin{figure*}
	\centering
	\includegraphics[width=0.85\textwidth]{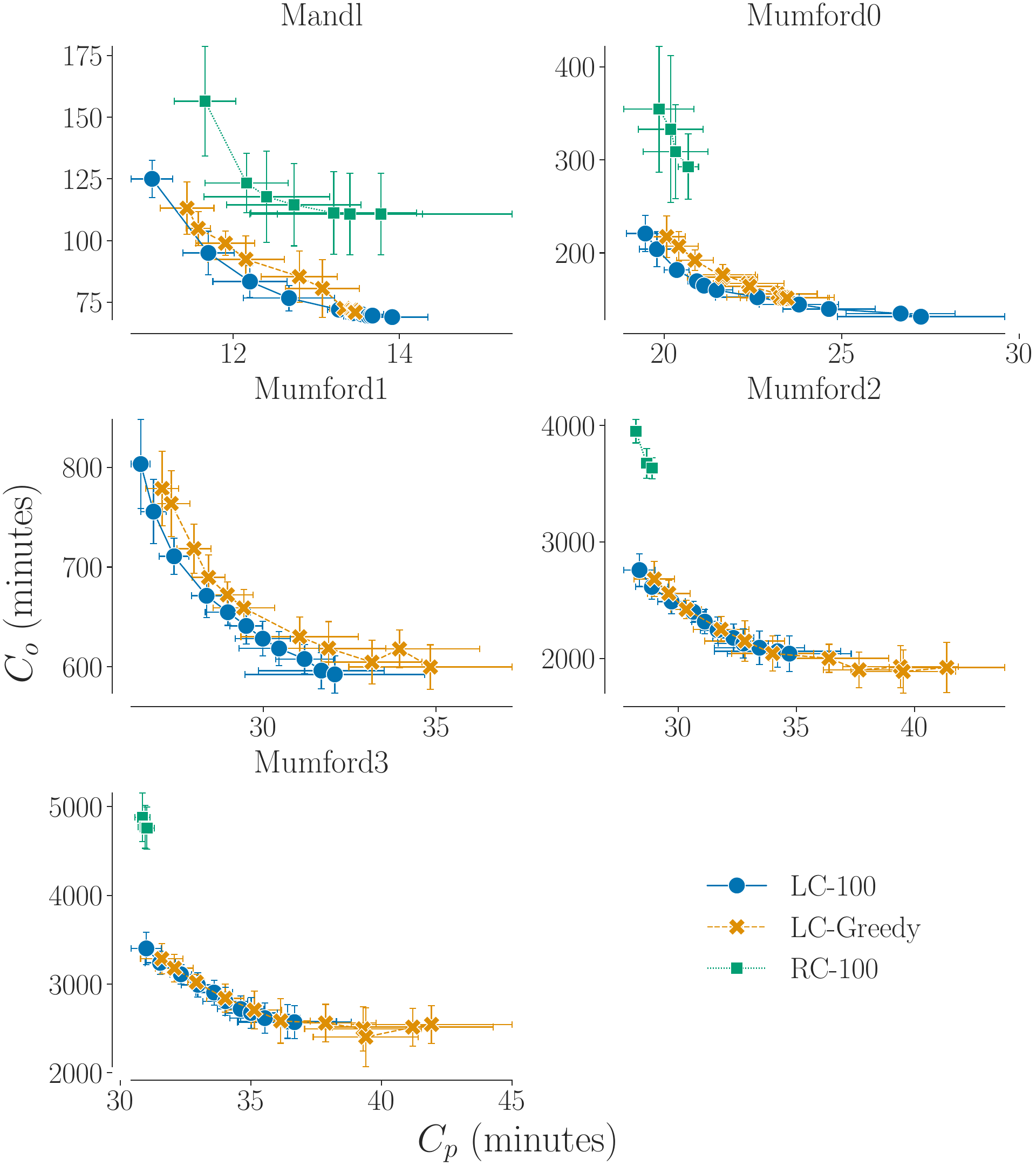}
	\caption{Values of average trip time $C_p$ (on the x-axis) and total route time $C_o$ (on the y-axis), across values of $\alpha$ from $0$ at lower-right to $1$ at upper-left, in increments of $0.1$.  Each point is the mean over 10 random seeds for one value of $\alpha$ (excluding those that generated invalid networks), and bars around each point indicate one standard deviation.  Lines link pairs of points with consecutive $\alpha$ values.  Note that RC-100 is not shown for Mumford1 because none of its generated networks were valid for that city.}
\label{fig:nn_pareto}	
\end{figure*}

\autoref{fig:nn_pareto} shows the results of this evaluation.  Each sub-figure shows, for one benchmark city, $C_p$ and $C_o$ for the transit networks produced by each algorithm over the values of $\alpha$.  As there is necessarily a trade-off between $C_p$ and $C_o$, these $(C_p, C_o)$ points form a curve for each algorithm that indicates how good of a trade-off it can achieve between $C_p$ and $C_o$ across values of $\alpha$.  As we seek to minimize both quantities, the down-and-leftward direction represents strict improvement.  

On Mandl and Mumford0, LC-100 dominates the random policy RC-100, producing networks for some values of $\alpha$ that dominate all of the networks produced by RC-100.  On Mumford1, RC-100 fails to generate any valid networks at all.  On Mumford2 and Mumford3, the two largest cities, the situation is a little more complex.  There, RC-100's networks for the different $\alpha$ values are tightly clustered in an area with high $C_o$ and low $C_p$.  While these networks have much higher $C_o$ than any of LC-100's or LC-Greedy's networks, making them worse for most values of $\alpha$, at the extreme of $\alpha = 1.0$ (where $C_o$ is irrelevant) RC-100's mean $C_p$ is slightly lower than that of LC-100 and LC-Greedy, though within one standard deviation of LC-100.  

This is surprising.  Na\"ively, we would expect that the policy net would learn to match the performance of $\pi_\text{random}$ by simply matching its probabilities: $\pi_\theta(\cdot | s_t, \alpha) = \pi_\text{random}$.  Yet it fails to do so.  To test whether this failure was due to the policy's having been trained over the full range of $\alpha$ values, we trained ten additional policy nets with ten random seeds in the same manner as described in \autoref{sec:training}, except that we held $\alpha=1.0$ throughout training.  The first set of policies we trained in this way did not perform any better at $\alpha=1.0$ than our existing policies.  We then tried adjusting the batch size and minibatch size, learning rate, and number of training iterations.  Finally we found that setting the batch size and minibatch size to 16, while leaving the other values unchanged from \autoref{tab:learning_params}, allowed us to train policies at $\alpha=1.0$ that achieved considerably lower $C_p$.  We called these the $\pi_{\theta_{\alpha=1}}$ policies.  

We then tried training policies over the full range of $\alpha$ with this batch size and minibatch size, but found that the these policies performed considerably worse overall than the ones trained with batch size 256.  So the smaller batch size is beneficial at $\alpha=1.0$ but not over the whole range of $\alpha$.

\begin{figure}
	\centering
	\includegraphics[width=\linewidth]{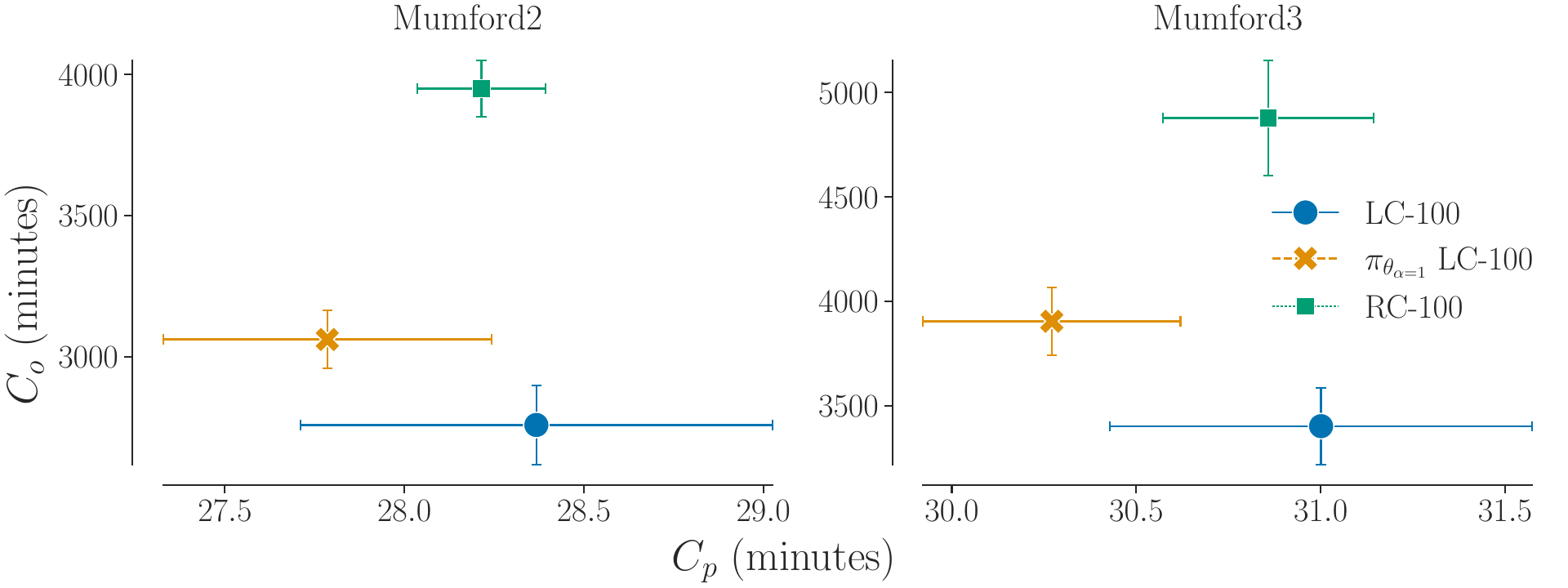}	
	\caption{Values of average trip time $C_p$ (on the x-axis) and total route time $C_o$ (on the y-axis) of solutions from LC-100 with $\pi_\theta$, LC-100 with $\pi_{\theta_{\alpha=1}}$, and RC-100, all with $\alpha=1.0$.}
	\label{fig:nn_extremes}
\end{figure}

We evaluated the $\pi_{\theta_{\alpha=1}}$ policies with the LC-100 algorithm over the five benchmark cities.  \autoref{fig:nn_extremes} shows the results at $\alpha=1.0$, in comparison with those from $\pi_\theta$ and $\pi_\text{random}$.  We see that, indeed, the $\pi_{\theta_{\alpha=1}}$ policies outperform both RC-100 and our variable-$\alpha$ policies on all five benchmark cities.  This shows that while training our policies over the full range of $\alpha$ gives good performance over the range, it does mean that the policies don't perform as well at the extremes of $\alpha$ as is possible.  While they are capable of learning a better-than-random policy at $\alpha=1.0$, they fail to fully learn the best policies at both $\alpha=1.0$ and $\alpha<1.0$, as this is a more challenging learning goal than $\alpha=1.0$ alone.  

Comparing LC-100 and LC-Greedy in \autoref{fig:nn_pareto}, we observe that on Mandl, Mumford0, and Mumford1, LC-100 dominates LC-Greedy, its curve being lower and to the left of it.  On Mumford2 and Mumford3, again the situation is more complex: LC-100 performs a bit better for $\alpha=1.0$, but LC-Greedy achieves lower $C_o$ at $\alpha \leq 0.4$, and in between their performance is comparable.  Overall, LC-100 is biased slightly towards favouring $C_p$ over $C_o$, in comparison with LC-Greedy.

The reason may be related to the high performance of RC-100 at $\alpha=1.0$.   RC-100 is a uniformly random policy, while LC-Greedy is a purely deterministic policy; LC-100, as a non-uniform stochastic policy, can be seen as being ``in between'' RC-100 and LC-Greedy.  Formally, we can think of this in terms of the temperature parameter, $temp$, of the softmax function.  $\pi_\text{random}$ is what we would get by taking $\pi_\theta$ (for which $temp = 1$ in our experiments) and setting $temp = \infty$; whereas $\pi_{\theta G}$ is the limit of $\pi_\theta$ as $temp \rightarrow 0$.  So it may be the case that our policies are somehow favouring $C_o$ over $C_p$, and the shift towards more stochasticity in the policy as we go from $\pi_{\theta G}$ to $\pi_{\theta}$ to $\pi_\text{random}$ gradually relaxes this preference for actions that favour $C_o$, which automatically improves $C_p$ due to the trade-off between the two.

\begin{figure}
	\includegraphics[width=0.6\linewidth]{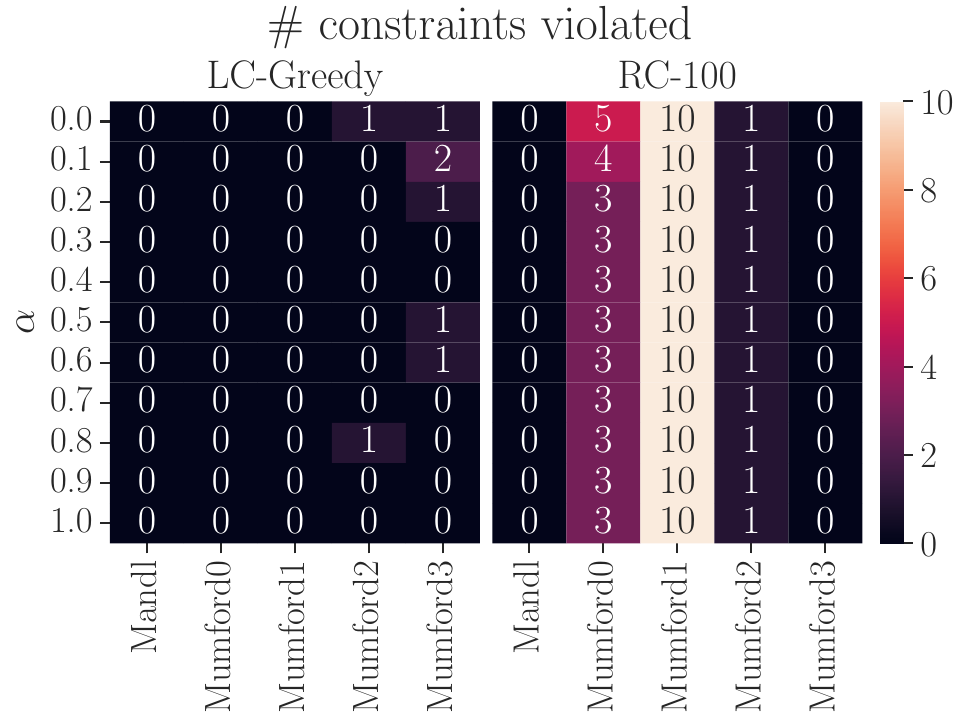}
	\caption{Of the ten runs with different random seeds for each benchmark city and $\alpha$, these heat maps show the number of solutions produced by each algorithm that violated one or more network constraints.  LC-100 is not shown, as it produced no constraint violations in any case.}
	\label{fig:constraints_violated}
\end{figure}

We must also consider how often each algorithm produce invalid transit networks that violate one or more of the constraints listed in \autoref{subsec:constraints}.  LC-100 produced no invalid networks with any of the learned policies.  Across the 110 trials (10 random seeds and 11 $\alpha$ values) for each city, however, LC-Greedy produced invalid networks in two trials on Mumford2, and in six trials on Mumford3.  Meanwhile, RC-100 produced invalid networks in 36 of 110 trials on Mumford1, 11 of 110 on Mumford2, and in all 110 trials on Mumford1.  \autoref{fig:constraints_violated} shows how many of the ten generated networks for each $\alpha$ and benchmark city violated a constraint.  

Using LC-Greedy, one of the ten trained policies generated an invalid network for 4 of the 11 values of $\alpha$ on Mumford3, and at $\alpha=0.1$ there were two invalid networks.  On Mumford2, just 2 of the 11 $\alpha$ values got an invalid network from one model in LC-Greedy mode.  On the other benchmark cities, all LC-Greedy networks were valid.  This is a low failure rate, but as most of the $\alpha$ values where constraint violations occur are $\leq 0.5$, it casts doubt on the apparent superiority of LC-Greedy over LC-100 at low $\alpha$ that we concluded from \autoref{fig:nn_pareto}.  LC-100 is more reliable than LC-Greedy, and so one would be justified in using it regardless of the value of $\alpha$.  

RC-100, meanwhile, produced many invalid networks, but mostly on Mumford0 and Mumford1 (in fact, it failed to produce any valid network on Mumford1).  On the larger Mumford2 and Mumford3 cities, it produced valid networks except with one random seed on Mumford2.  Perhaps this is because of the much larger values of $S$ specified on those cities made it easier to obtain full network coverage, so as not to violate the connectedness constraint.

Comparing our learned policy with a random policy shows that, while the \gls{ndp} is very challenging, there are regularities in its structure across different instances, enough so that heuristics can be learned from training on small instances that generalize well to much larger instances.

\subsection{Comparison with State of the Art Methods}

While the comparison of our learned policies with a random policy shows that they learn something useful, existing methods for the \gls{ndp} provide a more rigorous benchmark of performance.  \autoref{tab:nn_benchmark} compares LC-100's performance with that of a selection of other methods that have been benchmarked on the Mumford cities.  Apart from ours, all of the methods presented in this table are metaheuristic improvement methods that perform some search of the space of complete transit networks.  \cite{mumford2013new} use the multi-objective genetic algorithm SEAMO2~\citep{valenzuela2002simple, mumford2004simple} to achieve their results, running it on the Mumford cities for 200 generations with a population size (the number of networks considered in parallel) of 200.  Thus, a total of 40,000 transit networks are proposed, and their costs evaluated, over a single run.  By comparison, LC-100 considers 100 networks, and LC-Greedy considers just 1.  Despite this considerable gap in the scope of search, on the largest two cities (Mumford2 and Mumford3) LC-100 outperforms \cite{mumford2013new} at $\alpha=1$, and LC-Greedy outperforms \cite{mumford2013new} at $\alpha=0$.  

\begin{table*}
	\centering
	\begin{tabular}{llcc}
		\toprule
		City & Method & $C_p$ at $\alpha=1$ & $C_o$ at $\alpha=0$ \\
		\midrule
		Mandl & LC-100 & 11.00 & 69 \\
		& LC-Greedy & 11.44 & 71 \\
		& \cite{mumford2013new} & 10.27 & \textbf{63} \\
		& \cite{john2014routing} & 10.25 & \textbf{63} \\
		& \cite{ahmed2019hyperheuristic} & \textbf{10.18} & \textbf{63} \\
		& \cite{husselmann2023improved} & 10.19 & \textbf{63} \\
		\midrule
		Mumford0 & LC-100 & 19.47 & 131 \\
		& LC-Greedy & 20.07 & 152 \\
		& \cite{mumford2013new} & 16.05 & 111 \\
		& \cite{john2014routing} & 15.40 & 95 \\
		& \cite{ahmed2019hyperheuristic} & \textbf{14.09} & \textbf{94} \\
		& \cite{husselmann2023improved} & 14.34 & \textbf{94} \\
		\midrule
		Mumford1 & LC-100 & 26.46 & 592 \\
		& LC-Greedy & 27.07 & 600 \\	
		& \cite{mumford2013new} & 24.79 & 568 \\
		& \cite{john2014routing} & 23.91 & 462 \\
		& \cite{ahmed2019hyperheuristic} & \textbf{21.69} & \textbf{408} \\
		& \cite{husselmann2023improved} & 21.94 & 465 \\
		\midrule
		Mumford2 & LC-100 & 28.37 & 2040 \\
		& LC-Greedy & 29.00 & 1924 \\
		& \cite{mumford2013new} & 28.65 & 2244 \\
		& \cite{john2014routing} & 27.02 & 1875 \\
		& \cite{ahmed2019hyperheuristic} & \textbf{25.19} & \textbf{1330} \\
		& \cite{husselmann2023improved} & 25.31 & 1545 \\
		\midrule
		Mumford3 & LC-100 & 31.00 & 2571 \\
		& LC-Greedy & 31.59 & 2486 \\		
		& \cite{mumford2013new} & 31.44 & 2830 \\
		& \cite{john2014routing} & 29.50 & 2301 \\
		& \cite{ahmed2019hyperheuristic} & 28.05 & \textbf{1746} \\
		& \cite{husselmann2023improved} & \textbf{28.03} & 2043 \\
		\bottomrule
	\end{tabular}
	\caption{A comparison of our algorithms with a selection of methods from recent literature, in terms of $C_p$ achieved when $\alpha=1$ and $C_o$ achieved when $\alpha=0$.  Values for LC-100 and LC-Greedy are means over 10 random seeds.  Bold values are the best on the environment.}
\label{tab:nn_benchmark}
\end{table*}

The improvement in performance of our method relative to that of \cite{mumford2013dataset} as $n$ increases may be explainable by the constant number of networks that their algorithm considers in their evaluation.  As $n$ and $S$ grow, so does the space of possible transit networks.  A fixed number of networks (40,000 in Mumford's case) will then cover a smaller fraction of this search space, so we would expect the quality of the best of these to decrease relative to the quality of the best possible network for that city.  Meanwhile, if there are heuristics for this problem that generalize well across values of $n$ and $S$, then the relative quality of networks found via this heuristic may remain constant, or decline less quickly, as $n$ and $S$ grow.  These results are more evidence that the heuristics learned by our policies generalize well to large problem instances, and make clear that our method holds real promise for the \gls{ndp}.

\section{Summary}

While our learned policies outperform the first method to show results on the Mumford benchmark, they are themselves outperformed by the more recent work of \cite{john2014routing}, \cite{ahmed2019hyperheuristic}, and \cite{husselmann2023improved}.  All three of these use metaheuristic improvement methods, and so are more exhaustive and computationally costly than LC-100 and LC-Greedy: \cite{john2014routing} and \cite{husselmann2023improved}'s methods both take about two days on the largest city, Mumford3, using desktop hardware, while \cite{ahmed2019hyperheuristic}'s method takes ten hours.  By contrast, LC-100 and LC-Greedy take less than five minutes to run on Mumford3 on commercial hardware.  But as \autoref{tab:nn_benchmark} shows, the greater cost pays off in the quality of solutions found.  Observing this, we next sought ways of combining our learned policies with improvement metaheuristics, in the hope that this combination could improve on the performance of both.  

\chapter{Neural Nets as Initialization Procedures}\label{chap:initialization}

The quality of solutions found by a metaheuristic improvement algorithm is usually highly dependent on its initial solution, the place in solution space where it begins the search.  A poor starting point in a wide ``valley'' of equally low-quality solutions, or at the peak of a sub-optimal maximum, can cause the algorithm to search fruitlessly, never finding a hill to climb.  A good starting point, on or near the steep slope of a high-quality region, can allow it to find excellent solutions very quickly.

Existing metaheuristic algorithms for the \gls{ndp}	generally use simple construction heuristics to produce an initial transit network~\citep{nikolic2013transit, mumford2013dataset, john2014routing, ahmed2019hyperheuristic, husselmann2023improved}.  Most such heuristics greedily maximize or minimize some quantity, such as demand directly served or constraints violated.  They may involve many random restarts to find an initial network with no constraint violations.  These algorithms are engineered to be simple and fast at the expense of output quality, since the metaheuristic search that follows bears the burden of achieving high quality.

The neural net policy architecture we developed in \autoref{chap:neural_policy} is comparable in speed to these initialization algorithms: the LC-100 algorithm described in \autoref{subsec:nn_eval} runs in under 60 seconds on a desktop computer with a commercial GPU, whereas most initialization algorithms take seconds to minutes to run, and in at least one case, take hours~\citep{kilic2014demand}.  But its outputs appear to be much higher-quality, as it exceeds the performance of at least \cite{mumford2013dataset}'s metaheuristic algorithm on the Mandl and Mumford benchmarks.  This raises the question: do transit networks produced by the learned policy serve as good initial networks for improvement metaheuristics?

To answer this question, we conducted experiments with two metaheuristic algorithms for the \gls{ndp}, using different initialization procedures and comparing the performance of the algorithms' output transit networks.  The two metaheuristic algorithms we considered were a simple evolutionary algorithm (modified slightly from the algorithm of \cite{nikolic2013transit}), and the hyper-heuristic algorithm of \cite{ahmed2019hyperheuristic}.  We chose the former for its ease of implementation and fast running-time, and the latter because it was a recent algorithm whose authors reported state-of-the-art results on the Mumford benchmark, and because, not being an evolutionary algorithm, it would make for an interesting comparison with the first algorithm.

For each of the two algorithms, we performed experiments with three different initialization procedures:
\begin{itemize}
	\item The initialization procedure used in the original work (which is different for the two algorithms),
	\item The initialization procedure used by \cite{john2014routing},
	\item LC-100 with the learned policies $\pi_\theta$ from \autoref{chap:neural_policy}.
\end{itemize}

This allows us to compare, for each algorithm, our learned initialization procedure LC-100 with two human-designed procedures: one chosen by the authors of the algorithm, and one from a different algorithm (that of \cite{john2014routing}).  This will give some indication of the extent to which the original initialization procedures for each algorithm are especially suited to that algorithm.  In the following sections, we describe the two algorithms and their initialization procedures, and then present the experimental details and our results.

\section{Evolutionary Algorithm}\label{sec:evo_alg}


The metaheuristic improvement algorithm we used in these experiments is an evolutionary algorithm.  Evolutionary algorithms are a broad class of improvement algorithms in which a population of solutions undergo repeated stages of random modification and selection.  The modification step explores the solution space, while the selection step filters out low-quality solutions from the population.  Over many iterations, the quality of the population's solutions rises, and finally one can select the best solution from among them.

Our evolutionary algorithm is based on that of~\cite{nikolic2013transit}, with several modifications to the algorithm that we found improved its performance on the Mumford benchmark cities.  We chose this algorithm because its ease of implementation and short running times (with appropriate parameters) aided prototyping and experimentation.  

\nomenclature{$B$}{The population of solutions used in an evolutionary algorithm.}
\nomenclature{$b$}{The index of a solution in the population $B$ used in an evolutionary algorithm.}
\nomenclature{$F$}{The number of mutations performed per iteration in our evolutionary algorithm.}
\nomenclature{$I$}{The total number of iterations performed in our evolutionary algorithm.}

The algorithm operates on a population of transit networks $B = \{\mathcal{R}_b | 1 \leq b \leq |B|\}$, and performs alternating stages of mutation and selection.  In the mutation stage, the algorithm applies two ``mutators'', type 1 and type 2, to equally-sized subsets of the population chosen at random; if the mutated network $\mathcal{R}'_b$ has lower cost than its ``parent'' $\mathcal{R}_b$, it replaces its parent in $B$: $\mathcal{R}_b \leftarrow \mathcal{R}'_b$.  This is repeated $F$ times in the stage.  Then, in the selection stage, networks  either ``die'' or ``reproduce'', with probabilities inversely related to their cost $C(\mathcal{G}, \mathcal{R}_b)$.  After $I$ repetitions of mutation and selection, the algorithm returns the best network found over all iterations.

\nomenclature{$p_d$}{The probability of a node being deleted from a selected route during the second mutation operation of our evolutionary algorithm.}

Both mutators begins by selecting, uniformly at random, a route $r$ in $\mathcal{R}_b$ and a terminal node $i$ on that route.  The type-1 mutator then selects a random node $j \neq i$ in $\mathcal{N}$, and replaces $r$ with the shortest path between $i$ and $j$, $\text{SP}_{ij}$.  The probability of choosing each node $j$ is proportional to the amount of demand directly satisfied by $\text{SP}_{ij}$.  The type-2 mutator chooses with probability $p_d$ to delete $i$ from $r$; otherwise, it adds a random node $j$ in $i$'s street-graph neighbourhood to $r$ (before $i$ if $i$ is the first node in $r$, and after $i$ if $i$ is the last node in $r$), making $j$ the new terminal.  Following~\cite{nikolic2013transit}, we set the deletion probability $p_d = 0.2$ in our experiments.  

\begin{figure}
	\includegraphics[width=0.4\textwidth]{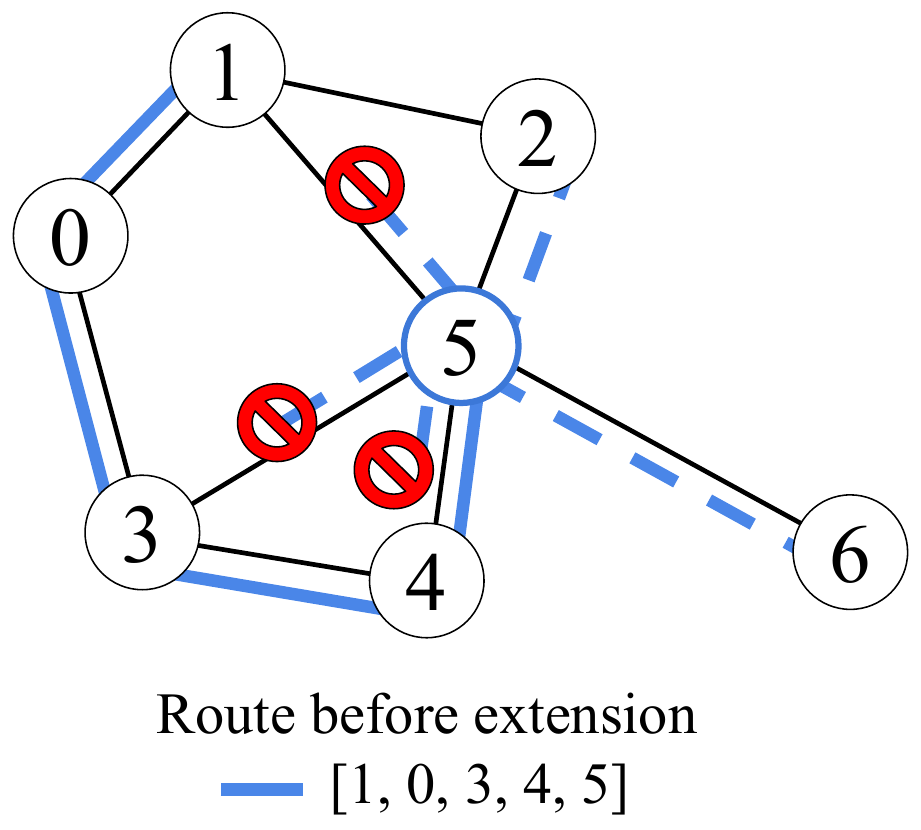}
	\caption{A schematic of which extensions of a route are allowed by the type-2 mutator.  Here, the mutator will extend a route from its terminal at node 5 to a neighbouring node.  Nodes 1, 2, 3, 4, and 6 are neighbours of node 5, but nodes 1, 3, and 4 are already on the route, so may not be chosen.  This leaves nodes 2 and 6 as valid choices, so one of these two will be chosen randomly.}
	\label{fig:type2mut}
\end{figure}

As shown in \autoref{fig:type2mut}, we disallow the type-2 mutator from choosing extending nodes that are already on the route, to avoid violating the simple-path constraint described in \autoref{subsec:constraints}.  The number-of-routes constraint is respected by construction, since the type-1 mutator replaces each route it deletes with another route.  Violations of the connectedness and number-of-stops constraints may occur, but the cost function's constraint-violation term $\beta C_c$ means that the algorithm will be more likely to keep solutions with fewer violations, and so ought eventually to drive $C_c$ to zero.

\nomenclature{$p_b$}{Probability of solution $b$ in our evolutionary algorithm being selected to reproduce.}

The parameters of this algorithm are the population size $|B|$, the number of mutations per mutation stage $F$, and the number of iterations of mutation and selection stages $I$.  In addition to these, the algorithm takes as input the city $\mathcal{G}$ being planned over, the set of shortest paths through the city's street graph $\textup{SP}$, and the cost function $C$.  The full procedure is given in \autoref{alg:ea}.
\begin{algorithm}
	\caption{Evolutionary Algorithm}\label{alg:ea}
	\begin{algorithmic}[1]
		\State {\bfseries Input:} city graph $\mathcal{G} = (\mathcal{N}, \mathcal{E}_s, D)$; set of shortest paths in city graph $\textup{SP}$; cost function $C$; population size $|B|$; number of iterations $I$; number of mutations per iteration $F$
		\State Construct initial network: $\mathcal{R}_0 \leftarrow \text{initialize}(\mathcal{G})$
		\State $\mathcal{R}_b \leftarrow \mathcal{R}_0 \; \forall \; b \in \text{integers } 1 \text{ through } |B|$
		\State $\mathcal{R}_\textup{best} \leftarrow \mathcal{R}_0$
		\For {$i=1$ to $I$}
		\State // Mutation stage
		\For{$j=1$ to $F$}
		\For{$b=1$ to $|B|$}
		\If {$b \leq |B| / 2$}
		\State $\mathcal{R}'_b \leftarrow$ type\_1\_mut($\mathcal{R}_b$)
		\Else
		\State $\mathcal{R}'_b \leftarrow$ type\_2\_mut($\mathcal{R}_b$)
		\EndIf
		\If {$C(\mathcal{G}, \mathcal{R}'_b) < C(\mathcal{G}, \mathcal{R}_b)$}
		\State $\mathcal{R}_b \leftarrow \mathcal{R}'_b$
		\EndIf
		\EndFor
		\State Randomly shuffle network indices $b$
		\EndFor
		\State // Selection stage
		\State $C_{max} \leftarrow \max_b C(\mathcal{G}, \mathcal{R}_b)$
		\State $C_{min} \leftarrow \min_b C(\mathcal{G}, \mathcal{R}_b)$
		\For{$b=1$ to $|B|$}
		\If {$C(\mathcal{G}, \mathcal{R}_b) < C(\mathcal{G}, \mathcal{R}_\textup{best})$}
		\State $\mathcal{R}_\textup{best} \leftarrow \mathcal{R}_b$
		\EndIf
		\State // Select surviving networks
		\State $O_b \leftarrow \frac{C_{max} - C(\mathcal{G}, \mathcal{R}_b)}{C_{max} - C_{min}}$
		\State $s_b \sim \textup{Bernoulli}(1 - e^{-O_b})$
		\State // Set survivor reproduction probabilities
		\State $p_b \leftarrow \frac{O_bs_b}{\sum_{b'} O_{b'}s_{b'}}$        
		\EndFor
		\If {$\exists \; b \in [1, |B|]$ s.t. $s_b = 1$}
		\State // Replace non-survivors with survivors' offspring 
		\For{$b=1$ to $|B|$}
		\If {$s_b = 0$}
		\State $k \sim P(k)$, where $P(k=b') = p_{b'}$
		\State $\mathcal{R}_b = \mathcal{R}_k$
		\EndIf
		\EndFor        
		\EndIf
		\EndFor
		\State return $\mathcal{R}_\textup{best}$
	\end{algorithmic}
\end{algorithm}

We note that \cite{nikolic2013transit} describe their algorithm as a ``bee colony optimization'' algorithm.  As noted in \cite{sorensen2015metaheuristics}, ``bee colony optimization'' is merely a relabelling of the components of one kind of evolutionary algorithm.  It is mathematically identical to this older, well-established metaheuristic.  To avoid confusion and the spread of unnecessary terminology, we here describe the algorithm as an evolutionary algorithm.

\section{Hyper-Heuristic with Great Deluge}

The Sequence-Based Selection Hyper-heuristic algorithm for the \gls{ndp} proposed by \cite{ahmed2019hyperheuristic} is an improvement metaheuristic that operates on a single network, unlike evolutionary algorithms which operate on a population of multiple networks.  The term ``hyper-heuristic'' in the literature denotes an adaptive mechanism within a metaheuristic algorithm that keeps track of how successful each low-level heuristic has been at improving a solution, and uses this record to preferentially select more successful low-level heuristics as the algorithm proceeds.

\nomenclature{$H$}{The set of heuristics used by the Hyperheuristic algorithm.}
\nomenclature{$h_i$}{The $i$-th heuristic used by the Hyperheuristic algorithm.}
\nomenclature{$Limit$}{The limit on the running time of the Hyperheuristic algorithm.}
\nomenclature{$Tran$}{The transition matrix for the heuristics used in the Hyperheuristic algorithm.}
\nomenclature{$Seq$}{The sequence-construction matrix used in the Hyperheuristic algorithm.}

At a high level, \cite{ahmed2019hyperheuristic}'s algorithm (henceforth denoted HH) has these components: a set $H$ of seven low-level heuristics, a cost function $C(\mathcal{G}, \mathcal{R})$, a $7 \times 7$ transition matrix $Tran$, a $7 \times 2$ ``sequence-construction matrix'' $Seq$, and an acceptance rule.  

Each of the seven low-level heuristics applies a different class of random modification to the existing network by applying some operation to one or two random routes involving adding, deleting, or rearranging random nodes on the routes.  For example, one heuristic removes a node on a route; another selects a route and inserts a random node in $\mathcal{N}$ at a random position; another exchanges two nodes between two routes.  All seven heuristics have in common that they modify at most two routes by altering either one or two total nodes, and that their selections of which routes and nodes to alter are made uniformly at random.  Also, each heuristic can modify routes in ways that violate the constraints described in \autoref{subsec:constraints}.  

The cost function is much like the one we describe in \autoref{subsec:cost_function}, but the scaling parameters $w_o$ and $w_p$ are not used, and separate weights are used for $C_p$ and $C_o$:
\begin{align}
	C_{HH}(\mathcal{G}, \mathcal{R}, \alpha, \beta, \gamma) = \alpha C_p(\mathcal{G}, \mathcal{R}) + \gamma C_o(\mathcal{G}, \mathcal{R}) + \beta C_c(\mathcal{G}, \mathcal{R})
\end{align}
The authors set the constraint-violation penalty $\beta = \infty$.  This way, if a modification violates any constraints, the algorithm will reject it.  Also, as their low-level heuristics may cause a route to violate the simple-path constraint, their $C_c$ term differs from ours in that $C_c > 0$ if that constraint is violated, so these routes will be rejected as well.

Starting from some initial network $\mathcal{R}_0$, HH works by repeatedly assembling sequences of the heuristics in $H$ and applying them to its current network $\mathcal{R}$ to get a modified network $\mathcal{R}'$.  To assemble a sequence of heuristics, it selects the first, $h_0$, uniformly randomly, and selects subsequent heuristics stochastically according to $Tran$, such that $P(h_i = k) = \frac{Tran_{h_{i-1}, k}}{\sum_l Tran_{h_{i-1}, l}}$.  After selecting each heuristic, it decides stochastically according to $Seq$ whether to end the sequence and apply it to $\mathcal{R}$: $P(end) = \frac{Seq_{h_i,0}}{Seq_{h_i,0} + Seq_{h_i, 1}}$.  This process repeats until $end$ is chosen; then the sequence $[h_0,...,h_{last}]$ is applied to $\mathcal{R}$, yielding $\mathcal{R}'$.

The algorithm then applies an acceptance rule to determine whether to accept the $\mathcal{R}'$ as its new current solution, replacing $\mathcal{R}$.  If $\mathcal{R}'$ is accepted, $Tran$ and $Seq$ are updated to increase the likelihood of each of the choices that formed the sequence.  The sequence is then cleared, and the algorithm repeats this process, continuing until a time limit $Limit$ is reached, at which point the lowest-cost network $\mathcal{R}_{best}$ found over the course of the run is returned.  In essence, the algorithm continually learns over its run which heuristics perform well when following which others, and which heuristics do or do not provide a good end to a sequence.  For the full details of the algorithm, we refer the reader to \cite{ahmed2019hyperheuristic}.

\nomenclature{$f_0$}{A parameter of the acceptance threshold used in the Hyperheuristic algorithm.}
\nomenclature{$\Delta f$}{A parameter of the acceptance threshold used in the Hyperheuristic algorithm.}
\nomenclature{$L_\tau$}{The acceptance threshold used in the Hyperheuristic algorithm after $\tau$ seconds have elapsed.}

\cite{ahmed2019hyperheuristic} experiment with multiple acceptance rules, and find that the Great Deluge rule provides the best performance.  This rule accepts a network $\mathcal{R}'$ if it strictly improves on the current network $\mathcal{R}$; otherwise, it accepts $\mathcal{R}'$ if its cost $C(\mathcal{G}, \mathcal{R}')$ is below a threshold $L_\tau = f_0 + \Delta f (1 - \frac{\tau}{Limit})$, where $\tau$ is the running time so far, and $f_0$ and $\Delta f$ are user-defined parameters.  This amounts to a linearly-decreasing acceptance threshold on cost, which starts at $f_0 + \Delta f$ and ends at $f_0$.

As \cite{ahmed2019hyperheuristic} did not provide a reference implementation, we implemented HH with the Great Deluge acceptance rule ourselves, based on their publication.  The authors report that speed is one of the advantages of their algorithm over others, claiming that they are able to run HH for 365,000 iterations on the Mumford3 city with $Limit = 63.5$ minutes, and for 524,000 iterations on Mumford2 with $Limit = 55$ minutes, on commercial desktop hardware - where each iteration corresponds to adding one heuristic to the current sequence.  We found that our implementation was much slower, requiring upwards of 10 hours to run for 365,000 iterations on Mumford3 on our high-end desktop hardware.  This is most likely due to our implementation being written in Python, an interpreted language that is slow for some purposes, unlike the authors' implementation which was written in C++ [C. Mumford, personal communication, December 2024].

Regarding the parameters of the acceptance rule, the authors describe $f_0$ as ``the final expected objective value'' and $\Delta f$ as ``the maximum change in the objective value''.  They state that the initial threshold is the cost of the initial network, implying $L_0 = f_0 + \Delta f = C(\mathcal{G}, \mathcal{R}_0)$, but provide no other information about the values of $f_0$ and $\Delta f$ used in their experiments, or how to choose such values.  We attempted to inquire with the authors about this question, but the only author we could reach was unable to provide an answer.  We then performed some preliminary experiments with different values of $f_0$ and $\Delta f$, but because of the slowness of our implementation, we were not able to search very extensively.  Although we were unable to find values of $f_0$ and $\Delta f$ that replicated the state-of-the-art results reported by \cite{ahmed2019hyperheuristic}, we found that setting $f_0=0$ and $\Delta f = C(\mathcal{G}, \mathcal{R}_0)$ gave results that were at least better than those of our evolutionary algorithm described in \autoref{alg:ea}.

\section{Initialization procedures}

\subsection{Demand-Maximizing Shortest Paths}\label{subsec:nikolic}

In their original work, \cite{nikolic2013transit} construct $\mathcal{R}_0$ by greedily selecting routes from $\text{SP}$, the set of all shortest paths in the street graph, that maximize the amount of directly-satisfied demand - that is, ignoring demand satisfied by transfers to or from other routes.  After each route is selected, the demands it satisfies directly are zeroed out, and not considered in the selection of subsequent routes.  No attempt is made to avoid constraint violations: \cite{nikolic2013transit} ignore the constraints we outline in \autoref{subsec:constraints}, and we rely on the $\beta C_c$ term in the cost function to encourage the evolutionary algorithm itself to remove them.  We refer to this initialization procedure as Nikolić (2013).  It generates initial networks that have relatively low $C_p$ (and $C_o$, since the routes are all shortest paths), but which may not be valid.

\subsection{Validity-Maximizing Shortest Paths followed by Repair}\label{subsec:ahmed}

\nomenclature{$\mathbb{P}$}{The set of candidate routes used to construct the starting transit network in the Hyperheuristic algorithm.}

To initialize HH, \cite{ahmed2019hyperheuristic} use a two-stage procedure that aims only to find an initial network $\mathcal{R}_0$ with no constraint violations.  In the first stage, a transit network is constructed by first finding the set $\text{SP}$ of all shortest paths between all node pairs in $\mathcal{N} \times \mathcal{N}$.  This is filtered to remove paths $r$ for which $|r| < m_\text{min}$ or $|r| > m_\text{max}$, leaving a set of candidate routes we will call $\mathbb{P}$.  Then, starting with an empty network $\mathcal{R}_0 = \{\}$, routes are added one at a time from $\mathbb{P}$ according to which route minimizes the number of constraint violations: 
\begin{equation}
	\mathcal{R} \leftarrow \mathcal{R} \cup \{\arg\min_{r \in \mathbb{P}} C_c(\mathcal{R} \cup \{r\})\}
\end{equation}

This repeats until the desired number of routes are obtained ($|\mathcal{R}| = S$).

If $C_c(\mathcal{R}) = 0$ at the end of the first stage, the second stage, called the ``repair'' stage, begins.  It enters a loop where it selects a random heuristic from the seven heuristics in $H$, and applies it to $\mathcal{R}$ to get $\mathcal{R}'$; $\mathcal{R}'$ replaces $\mathcal{R}$ if $C_c(\mathcal{R}') \leq C_c(\mathcal{R})$, and is discarded otherwise.  This loop repeats until $C_c(\mathcal{R}) = 0$.  The resulting network becomes $\mathcal{R}_0$, the initial network for HH.  We refer to this initialization procedure as Ahmed (2019).

Ahmed (2019) ignores the quality of $\mathcal{R}_0$ in terms of $C_p$ and $C_o$, in favour of ensuring that it is a valid network.  We note that each low-level heuristic in $H$ is more likely to create new constraint violations than to repair existing ones, and as a result, we found that the repair procedure is very time-consuming for large graphs: On Mumford2 and Mumford3, it took between 10 and 20 hours, as long or longer than the HH algorithm itself.

\subsection{Growing Routes based on Diverse Edge Weights}\label{subsec:john}

The initialization procedure used by \cite{john2014routing} is intended for multi-objective optimization algorithms that require a diverse set of initial solutions.  It therefore attempts to produce not a single network, but a diverse set of networks that both achieve low $C_p$ and $C_o$ while also satisfying all constraints.  To generate a set of $M$ networks, it first generates a set of $M$ different weighted street graphs $\{\mathcal{N}, \mathcal{E}_s, W^m\}$, where $W^m$ is a set of weights of edges in $\mathcal{E}_s$.    Then, each of these weighted street graphs is used to construct a network in two stages.  The first stage assembles routes and adds them to $\mathcal{R}$ until all nodes in $\mathcal{N}$ are covered by $\mathcal{R}$.  If $|\mathcal{R}| < S$ at the end of the first stage, the second stage is carried out to add more routes to connect node pairs that are not yet directly connected, until $|\mathcal{R}| = S$.  We refer to this initialization procedure as John (2014).

\subsubsection{Pre-calculation stage: Weighted Street Graphs}

\nomenclature{$\lambda_1$}{The weight of the normalized drive time used to compute edge weights in the initialization algorithm of John et al. (2014).}
\nomenclature{$\lambda_2$}{The weight of inter-node demand used to compute edge weights in the initialization algorithm of John et al. (2014).}

A diverse set of $M$ weighted street graphs are first computed using scalar parameters $\lambda_1$ and $\lambda_2$, each of which has $M$  values $\{(\lambda^0_1, \lambda^0_2), (\lambda^1_1, \lambda^1_2), ..., (\lambda^M_1, \lambda^M_2)\}$ different specified by the user.  The weight $w^m_{ij}$ of edge $(i,j)$ in the $m$th weight set $W^m$ is computed as a weighted sum of the normalized edge drive time and the normalized demand:

\nomenclature{$W^m$}{The set of edge weights used to assemble routes in the initialization algorithm of John et al. (2014).}
\nomenclature{$w^m_{ij}$}{The weight of the edge between nodes $i$ and $j$ used to assemble routes in the initialization algorithm of John et al. (2014).}

\begin{equation}
w^m_{ij} = \lambda_{1,m} \frac{\tau_{ij}}{\max_{k,l\in \mathcal{E}_s} \tau_{kl}} + \lambda_{2,m} \left (1 - \frac{D_{ij}}{\max D} \right )
\end{equation}
Each weighting $W^m$ is used in one run of the first and second stages in order to select edges to include in routes, with lower weights being preferred.

\subsubsection{First stage: Score-based route assembly}

\nomenclature{$\mathcal{N_R}$}{The subset of nodes in a street graph that are visited by a transit network $\mathcal{R}$.}
\nomenclature{$\mathcal{E}_c$}{A set of candidate edges in a street graph, a subset of $\mathcal{E}_s$, that are used to assemble routes in the initialization algorithm of John et al. (2014).}

We define $\mathcal{N_R} \subset \mathcal{N}$ as the set of nodes visited by at least one route $r \in \mathcal{R}$; if $i \notin \mathcal{N_R}$, then $i$ is not served by the transit network.

Using weight set $W^m$, this stage constructs one route at a time and adds it to $\mathcal{R}$.  Each route is constructed in steps, starting from $r = []$; at each step one edge is selected from $\mathcal{E}_s$ and added to $r$.  The step first assembles a set $\mathcal{E}_c$ of candidate edges, where $\mathcal{E}_c \subset \mathcal{E}_s$.  If $|\mathcal{E}_c| > 0$, the candidate edge with the lowest score (breaking ties randomly) is selected: $(i,j) = \arg\min_{(i,j) \in \mathcal{E}_c} w^m_{ij}$, and it is appended to $r$ at the appropriate place.  If $\mathcal{E}_c = \{\}$ (there are no valid extensions for $r$) and $|r| < m_\text{min}$, $r$ is reset to $[]$ and construction starts again; if $\mathcal{E}_c = \{\}$ and $|r| \geq m_\text{min}$, or if $|r| = m_\text{max}$, $r$ is considered finished and is added to $\mathcal{R}$.  Then, if $\mathcal{N_R} \neq \mathcal{N}$, construction of a new route begins; if $\mathcal{N_R} = \mathcal{N}$, all nodes are covered by transit, and the first stage ends.

At the start of the first stage, when $r = []$ and $\mathcal{R} = \{\}$, all edges are candidates: $\mathcal{E}_c = \mathcal{E}_s$.  When starting subsequent routes (that is, when $r=[]$ and $|\mathcal{R}| > 0$), candidate edges are those which connect a served node to an unserved node:
\begin{equation}
	\mathcal{E}_c = \{i,j | (i,j,\tau_{ij}) \in \mathcal{E}_s, i \in \mathcal{N_R}, j \notin \mathcal{N_R}\}
\end{equation}

In steps after a route's first edge is selected, when $r = [s, ..., e]$, $\mathcal{E}_c$ is formed from the set of edges from either end of $r$ that do not connect to nodes already on $r$:
\begin{equation}
	\mathcal{E}_c = \{i,j | (i,j,\tau_{ij}) \in \mathcal{E}_s, j \in \{s, e\}, i \notin r \}
\end{equation}
Furthermore, if any extensions of $r$ would extend service to an unserved node, one of those extensions must be chosen.  That is, if $\exists \; k \in \{j | (i,j) \in \mathcal{E}_c\} \; \text{s.t.} \; k \notin \mathcal{N_R}$, we update the candidate set:
\begin{equation}
	\mathcal{E}_c \leftarrow \{i,j | (i,j) \in \mathcal{E}_c, j \notin \mathcal{N_R}\}
\end{equation}

\subsubsection{Second stage: Growing network to required size}

\nomenclature{$\mathcal{P_R}$}{The set of node pairs in a street graph that are not directly linked by any route in transit network $\mathcal{R}$.}

If at this point $|\mathcal{R}| < S$, the second stage begins.  This stage begins by building a collection $\mathcal{P_R}$ of node pairs that are not directly linked by any route in $\mathcal{R}$:
\begin{equation}
\mathcal{P_R} = \{i,j | i \in \mathcal{N}, j \in \mathcal{N}, \nexists \; r \in \mathcal{R} \; \text{s.t.} \; i \in r \; \text{and} \; j \in r\}
\end{equation}
It iterates over the pairs $(i,j) \in \mathcal{P_R}$ in descending order of $D_{ij}$. For each pair, it uses Yen's $k$-shortest-path algorithm~\citep{yen1970algorithm} with $k=10$ and $W^m$ as the edge weights to find the ten lowest-weight paths between $i$ and $j$ through $\mathcal{E}_s$.  It iterates over these ten paths until it finds one, $r$, for which $m_\text{min} \leq |r| \leq m_\text{max}$ and $\tau_{rij} < \tau_{\mathcal{R}ij}$; if it finds such an $r$, it adds $r$ to $\mathcal{R}$.  After finding or failing to find $r$ for $(i,j)$, it repeats this process with the next node pair, until $|\mathcal{R}| = S$.

Finally, the procedure returns $\mathcal{R}$ for use as $\mathcal{R}_0$ by the outer algorithm.

\subsubsection{Adaptation to single-objective setting}

In his doctoral thesis~\citep{john2016thesis}, John describes setting $\lambda_1$ and $\lambda_2$ to values ranging from $0$ to $1$ in increments of $0.05$, giving 21 unique values for each, and pairing all combinations of these to give 441 $(\lambda_{1,m}, \lambda_{2,m})$ pairs, so that his initialization algorithm produces 441 transit networks.  We are concerned in this chapter with initializing single-objective metaheuristic algorithms, each of which requires only a single initial solution.  We therefore adapt John's algorithm to this setting by specifying only a single weight pair $(\lambda_1, \lambda_2)$ to compute a single set of street edge weights $W$, and generating a single network from this.  We discuss the specific choice of values of $\lambda_1$ and $\lambda_2$ in \autoref{subsec:init_results}.

\section{Experiments}\label{sec:init_experiments}

As in \autoref{chap:neural_policy}, both the initialization methods and the metaheuristic improvement methods that use them are evaluated on the Mandl~\citep{mandl1980evaluation} and Mumford~\citep{mumford2013new} benchmark cities, over the same range of $\alpha$ values - from $\alpha=0$ to $\alpha=1$ in steps of $0.1$ - and the same ten random seeds.

\subsection{Initialization Results}\label{subsec:init_results}

\begin{figure*}
	\centering
	\includegraphics[width=0.85\textwidth]{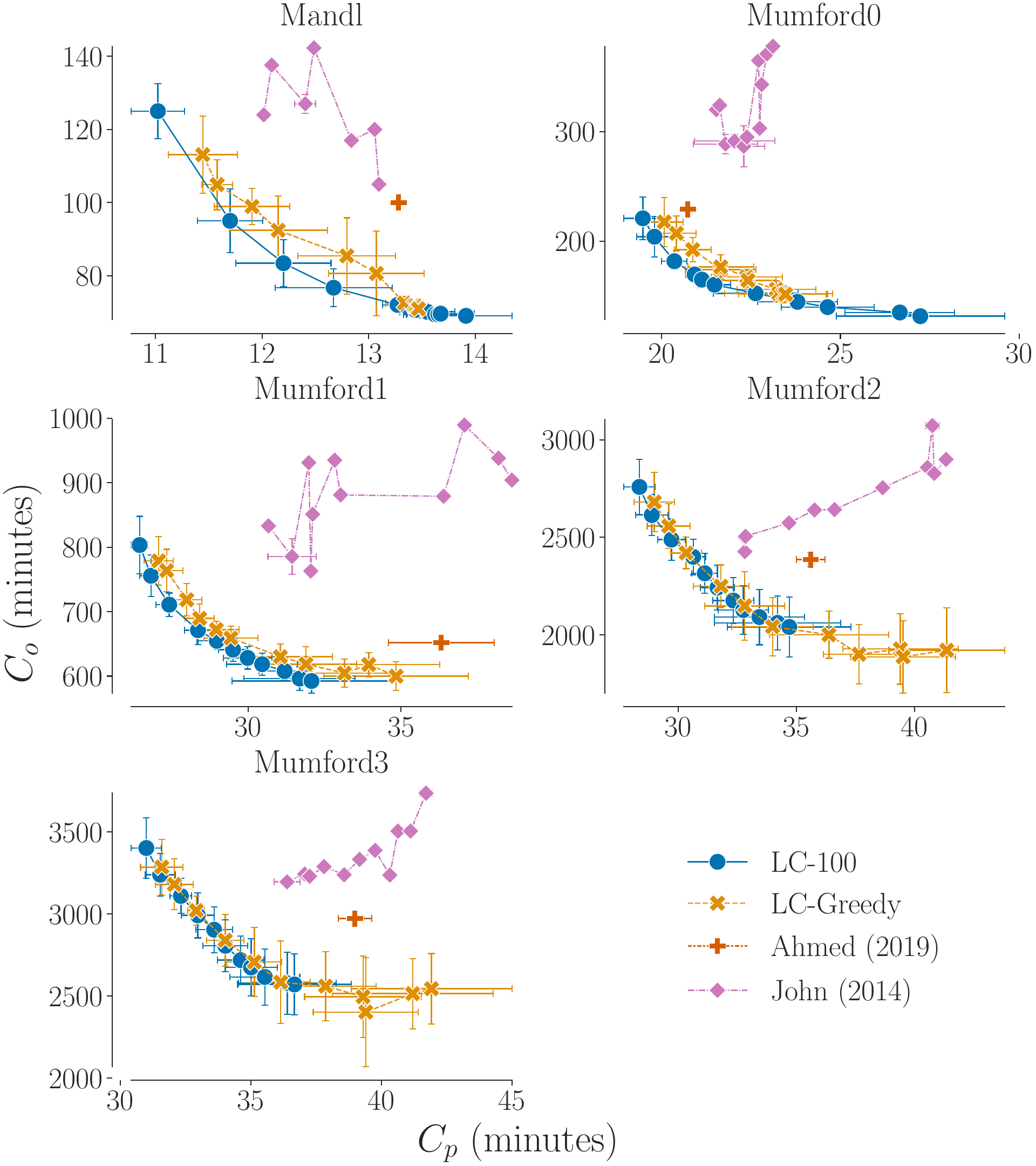}
	\caption{Results for the initialization methods.  Nikolic (2013) is excluded as all of its networks are invalid.  Curves for LC-100 and John (2014) show values of average trip time $C_p$ (on the x-axis) and total route time $C_o$ (on the y-axis) across values of $\alpha$ from $0$ to $1$ in increments of $0.1$.  Ahmed (2019) does not depend on $\alpha$, so only one point is shown for it.  Each point is the mean over 10 random seeds for one value of $\alpha$ and one method, and bars around each point indicate one standard deviation.}
	\label{fig:init_only}
\end{figure*}

We began by running each of the three pre-existing initialization methods - Nikolić (2013), John (2014), and Ahmed (2019) - over the range of random seeds and $\alpha$ values and comparing the results with those from LC-100 and LC-Greedy, as presented in \autoref{sec:nn_results}.  \autoref{fig:init_only} shows the results of each.

To choose values of $\lambda_1$ and $\lambda_2$ in John (2014), we observed that, like $1 - \alpha$ and $\alpha$, $\lambda_1$ and $\lambda_2$ are weights that control the linear trade-off between a term related to route travel time ($\lambda_1, 1-\alpha$) and a term related to demand ($\lambda_2, \alpha$).  While the terms are not identical, a relatively high $\lambda_1$ will (like $1 - \alpha$) prioritize shorter routes, while a high $\lambda_2$ will (like $\alpha$) prioritize satisfying demand directly, without transfers.  This analogy suggests setting $\lambda_1 = 1 - \alpha$ and $\lambda_2 = \alpha$, and so that was how we set these parameters to obtain the results for John (2014) presented in \autoref{fig:init_only}.

We note that at most $\alpha$ values, John (2014) creates the same network regardless of random seed; this is because randomness only enters into the algorithm if there is a tie in the scores of candidate edges, which happens rarely.  As a result, most of the points shown for John (2014) in \autoref{fig:init_only} have no error bars.  Also, on the Mandl city, John created the same network for several adjacent values of $\alpha$ - the changes in $\lambda_1,\lambda_2$ were in those cases too small to induce any change in outcome for this very small city, so these points are ``stacked'' in the plot, giving the appearance of fewer than 10 points.

We observe from this figure, however, that setting $\lambda_1 = 1 - \alpha$ and $\lambda_2 = \alpha$ does not lead John (2014) to make an appropriate trade-off between $C_p$ and $C_o$ as we would hope.  Surprisingly, for the three largest cities, Mumford1, 2, and 3, the values imposed by $\alpha=0.0$ - that is, $\lambda_1 = 1.0, \lambda_2 = 0.0$ - appear to perform best or near-best in terms of both $C_p$ and $C_o$, with both cost metrics tending to grow as $\alpha$ grows.  For this reason, in the subsequent experiments in which we use John (2014) to initialize different metaheuristic improvement methods, we always set $\lambda_1 = 1.0, \lambda_2 = 0.0$, regardless of $\alpha$.

Broadening our attention to the full set of methods in \autoref{fig:init_only}, we observe that both John (2014) and Ahmed (2019) produce network or sets of networks that are dominated by LC-100 and LC-Greedy.  Ahmed (2019)'s networks are biased towards low $C_o$ and high $C_p$, relative to those from LC-100 and LC-Greedy, whereas John (2014)'s networks strike a more even balance.  

From these results we might expect that one of LC-100 or LC-Greedy would provide better performance than John (2013) and Ahmed (2019) when used to initialize a metaheuristic improvement process.  These results provide key context for interpreting the results of the next section.

\subsection{Metaheuristic Improvement Results}\label{subsec:post_init_results}

\subsubsection{Evolutionary Algorithm (EA)}

\begin{figure*}
	\centering
	\includegraphics[width=0.85\textwidth]{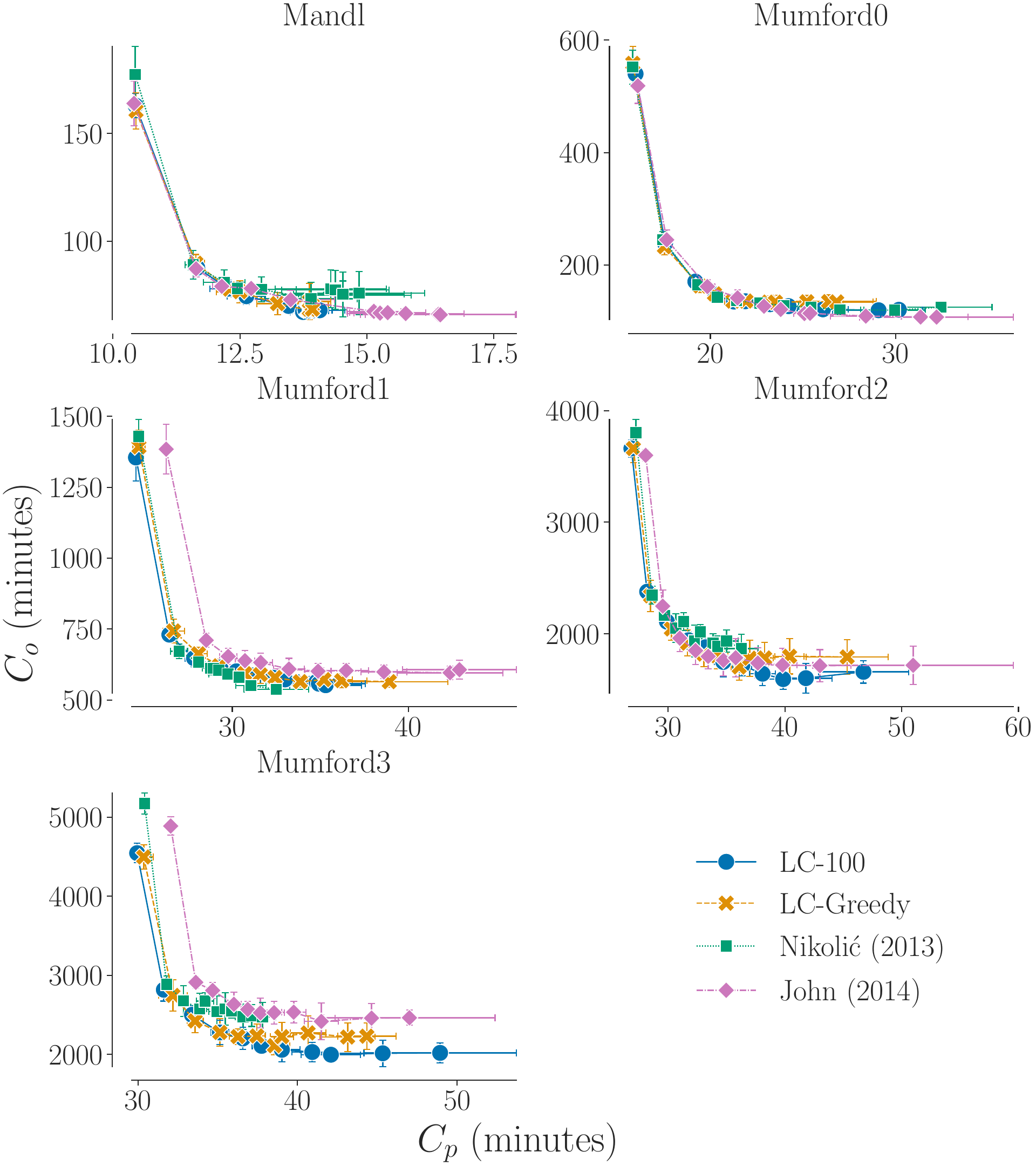}
	\caption{Results for EA with four different initialization methods.  Each curve shows values of average trip time $C_p$ (on the x-axis) and total route time $C_o$ (on the y-axis) across values of $\alpha$ from $0$ at lower-right to $1$ at upper-left, in increments of $0.1$.  Each point is the mean over 10 random seeds for one value of $\alpha$ and one method, and bars around each point indicate one standard deviation.  Lines link pairs of points for the same method and consecutive $\alpha$ values.}
	\label{fig:ea_init}
\end{figure*}

\autoref{fig:ea_init} shows the results obtained by running EA after initializing it by four different methods: LC-100, LC-Greedy, Nikolić (2013) (its original initialization procedure), and John (2014).  On Mandl and Mumford0, the performance of all methods is very similar.  But Mumford1, John (2014) is dominated by the rest, and on Mumford2 and Mumford3, LC-100 shows better performance especially at low values of $\alpha$, though it also performs best by a small margin at $\alpha=1.0$.   

\begin{figure*}
	\centering
	\includegraphics[width=0.4\textwidth]{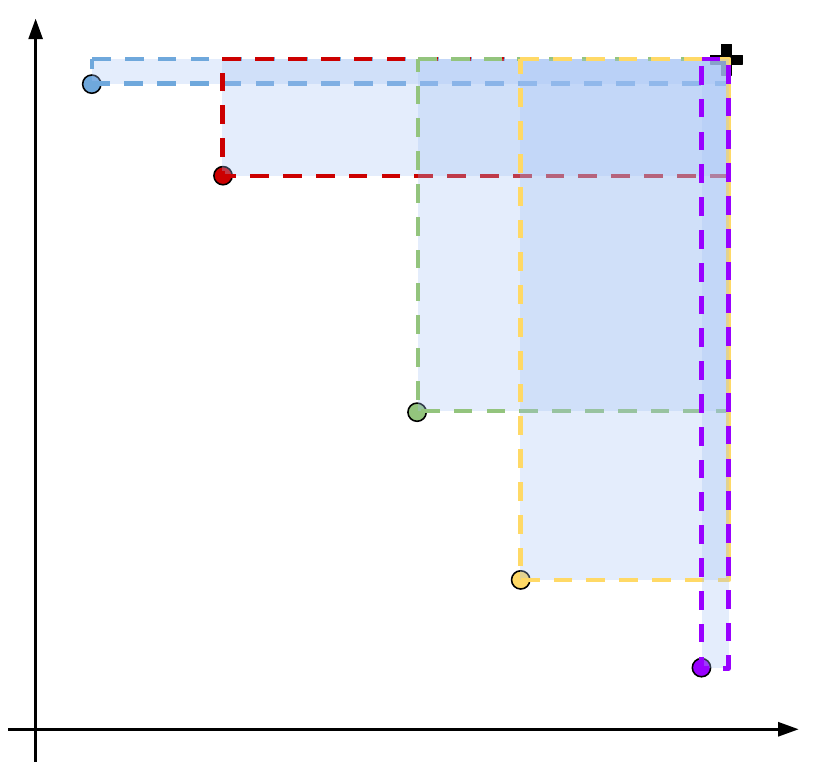}
	\caption{A schematic of a hypervolume computed based on five points in two dimensions.  The point $p$, represented by + sign, is chosen to be $\epsilon$ greater than the largest magnitude of any point on each dimension.  Then, we can define an axis-aligned rectangle for each point $x$ with $x$ at one corner and $p$ at the other.  They hypervolume of the set of five points is then the area of the union of these five rectangles, shown as the total shaded area.}
	\label{fig:hypervolume}
\end{figure*}

To allow a clearer comparison between the different initializations, we used the means of $C_p$ and $C_o$ computed for each $\alpha$ value of each method, shown in \autoref{fig:ea_init}, to compute a hypervolume (also known as an S-metric) for each method.  The hypervolume is a metric commonly used to evaluate a set of solutions in a multi-objective context.  Each solution in a set can be viewed as a point $x$ in a $d$-dimensional space, where $d$ is the number of different objectives we care about - in this context, $d=2$, the two objectives being $C_p$ and $C_o$.  Given some reference point $p$ in this space, we can define a hyper-rectangle for each solution point $x$ with opposite corners at $p$ and $x$, which has a ``hyper-volume'' $v = \prod_{i \leq d} |p_i - x_i|$.  The hypervolume $v(X)$ of the set $X$ is simply the hypervolume of the shape formed by taking the union of these hyper-rectangles for each point $x \in X$.  Assuming that we want to minimize each objective, we can choose $p_i$ to be a point greater on each dimension than any point in the Pareto set $X'$ of $X$: the set $X' \subset X$ such that no point $x \in X'$ is dominated by any other point $x' \in X'$.  Then, the hypervolume $v(X)$ grows as the set of solution points in $X$ improves.  \autoref{fig:hypervolume} illustrates this concept.

\begin{figure*}
	\centering
	\includegraphics[width=\textwidth]{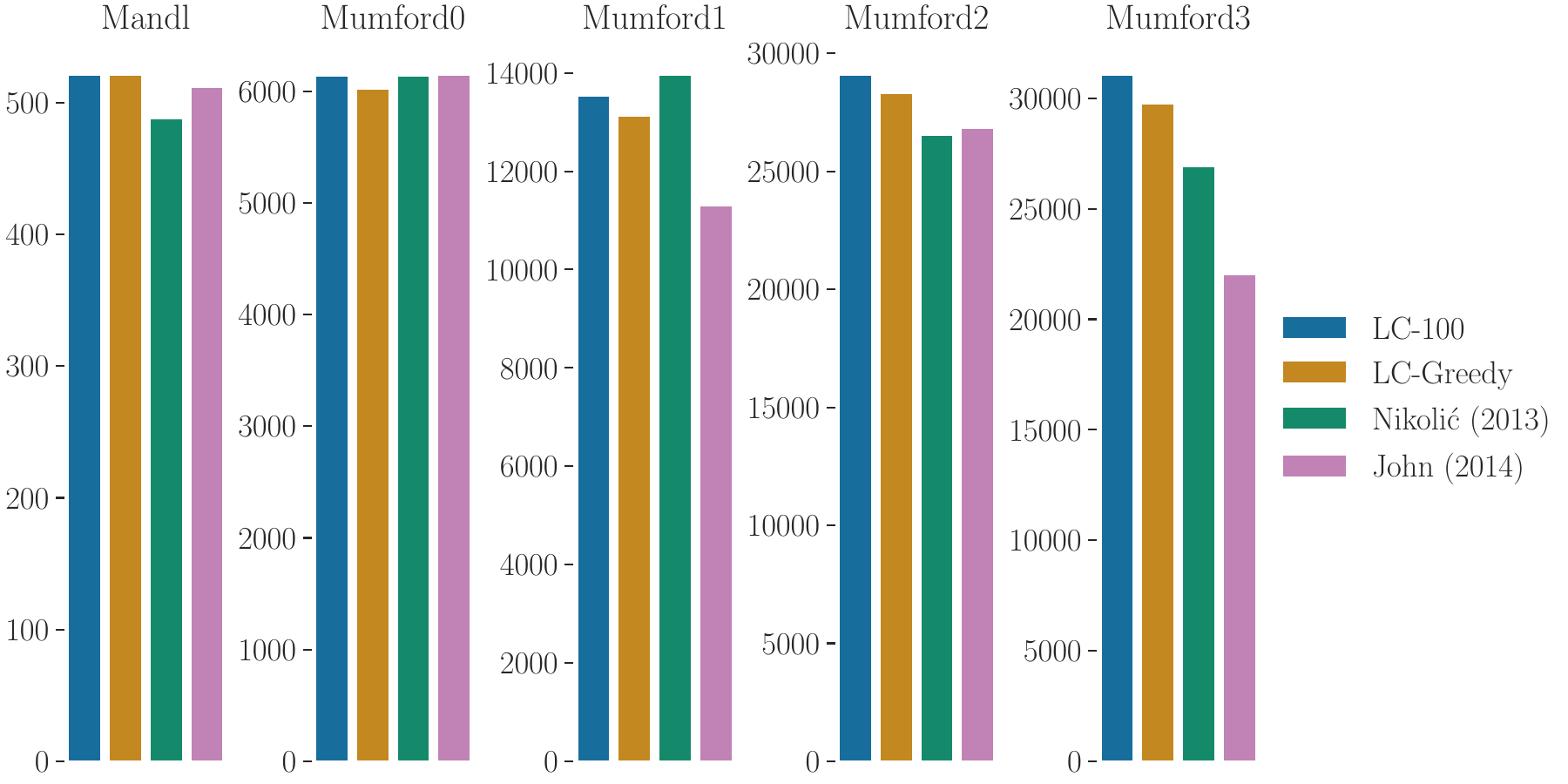}
	\caption{Hypervolumes achieved by EA with different initializations.}
	\label{fig:ea_hv}
\end{figure*}

\autoref{fig:ea_hv} compares the hypervolumes achieved by EA with the four initialization methods.  Each method's hypervolume is computed with respect to the same reference point $p = (C^{max}_p + \epsilon, C^{max}_o + \epsilon)$, where $C^{max}_p$ and $C^{max}_o$ are the largest values of the mean $C_p$ and $C_o$ observed across all methods and values of $\alpha$, and $\epsilon$ is a very small constant, which we set to $10^{-5}$ minutes.  This metric makes it clearer that on Mumford1, 2, and 3, LC-100 and LC-Greedy outperform John (2014), and they outperform Nikolić (2013) on the largest two cities, Mumford2 and 3.  LC-100 performs slightly better than LC-Greedy across the Mumford cities.  


\subsubsection{Hyper-Heuristic with Great Deluge (HH)}

In their experiments, \cite{ahmed2019hyperheuristic} set a time limit, $Limit$, for the HH algorithm to run based on the number of nodes $n$ in the target city.  To constrain the time taken to run these experiments, we instead set a limit of $500,000$ iterations for all experiments, which is in between the numbers of iterations that occurred in their experiments on Mumford2 and on Mumford3.  The form of the Great Deluge acceptance rule is changed accordingly: the threshold equation $L_\tau = f_0 + \Delta f (1 - \frac{\tau}{Limit})$ becomes $L_\tau = f_0 + \Delta f (1 - \frac{i}{500,000})$, where $i$ is the number of iterations so far.  Assuming constant time per iteration, this keeps the behaviour identical to the original time-limited HH.

As shown in \autoref{fig:constraints_violated}, some of LC-Greedy's networks for Mumford3 have constraint violations.  In these cases, we apply the repair stage of Ahmed (2019) to it to obtain a valid network, before running HH.

\begin{figure*}
	\centering
	\includegraphics[width=0.85\textwidth]{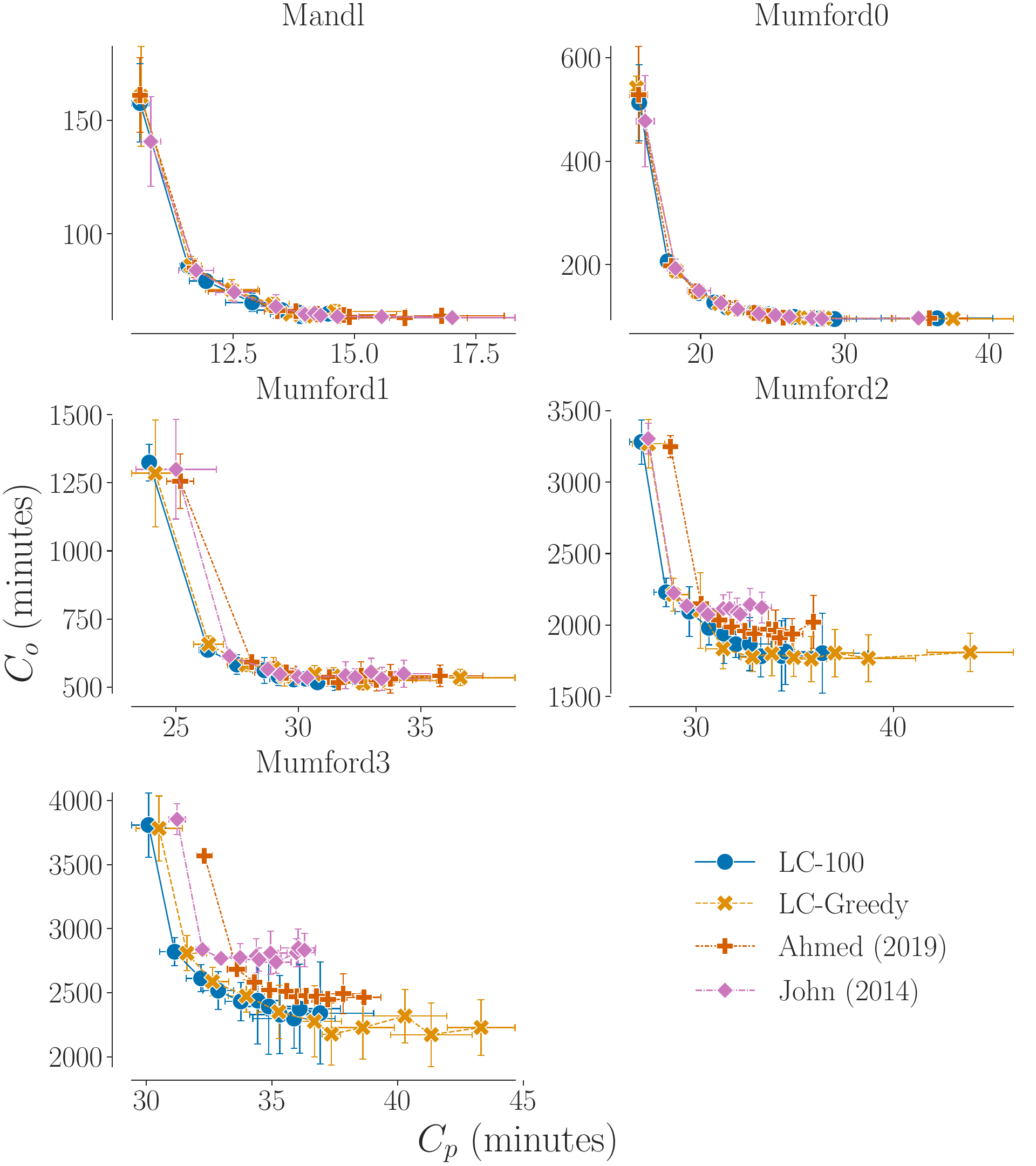}
	\caption{Results for HH with three different initialization methods.  Each curve shows values of average trip time $C_p$ (on the x-axis) and total route time $C_o$ (on the y-axis) across values of $\alpha$ from $0$ at lower-right to $1$ at upper-left, in increments of $0.1$.  Each point is the mean over 10 random seeds for one value of $\alpha$ and one method, and bars around each point indicate one standard deviation.  Lines link pairs of points for the same method and consecutive $\alpha$ values.}
\label{fig:hh_init}
\end{figure*}

\begin{figure*}
	\centering
	\includegraphics[width=\textwidth]{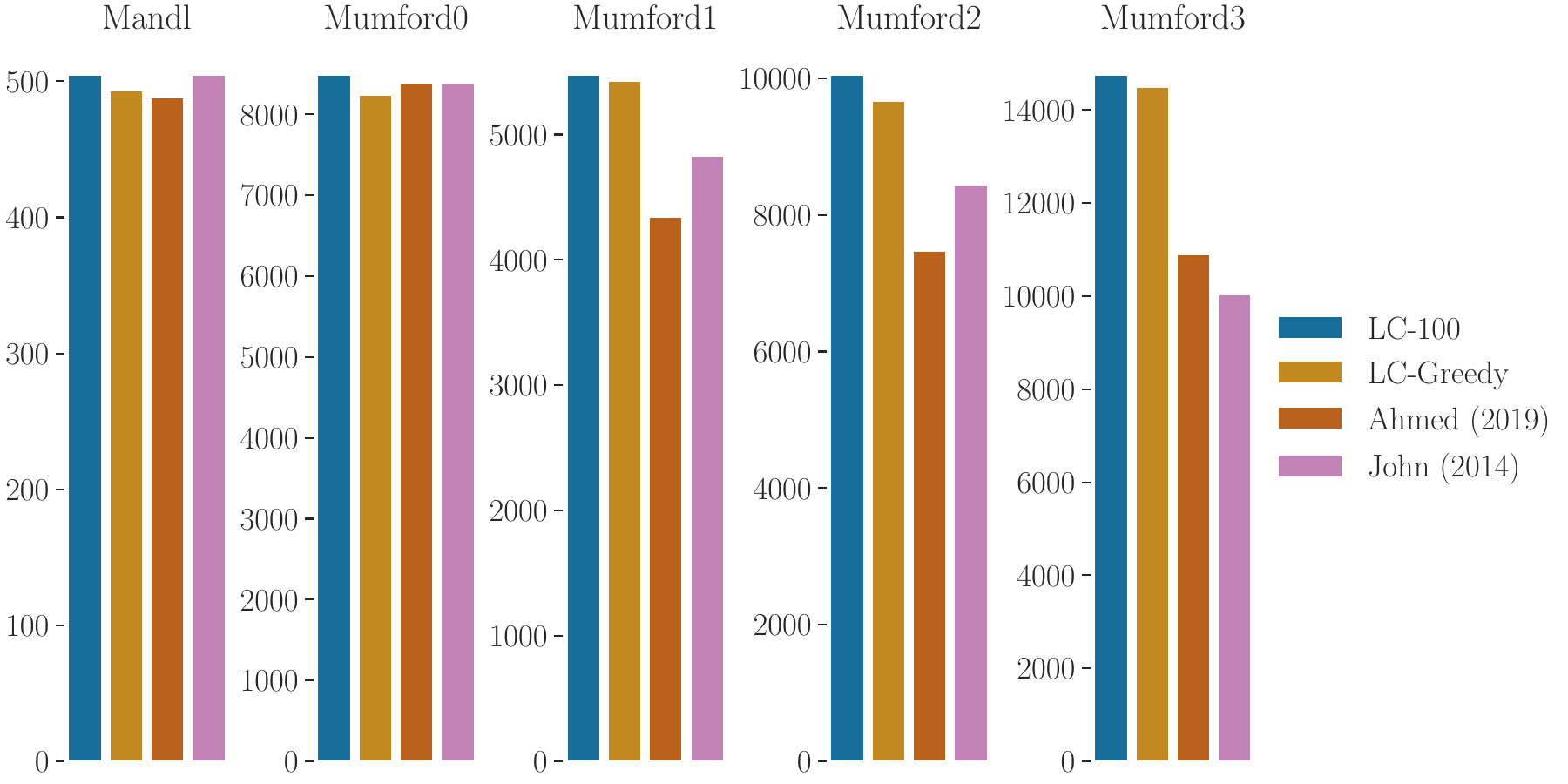}
	\caption{Hypervolumes achieved by HH with different initializations.}
	\label{fig:hh_hv}
\end{figure*}

\autoref{fig:hh_init} shows the results obtained by running HH over the range of $\alpha$ values with four different initialization methods.  As in the case of EA, we see here that LC-100 and LC-Greedy dominate the non-learned methods on Mumford2 and Mumford3, while all methods perform very similarly on Mandl, Mumford0, and Mumford1, which are smaller.  We also observe that, unlike EA, HH initialized with LC-Greedy dominates LC-100 at lower values of $\alpha$ on Mumford2 and Mumford3, though LC-100 still dominates at higher values of $\alpha$.  So unlike EA, HH appears to preserve the relative advantages of LC-100 and LC-Greedy.

\autoref{fig:hh_hv} shows the hypervolumes attained by HH with these four initialization methods.  Confirming what we observe in the \autoref{fig:hh_init}, LC-Greedy and LC-100 attain the largest hypervolumes on Mumford2 and 3, but this figure reveals that they perform best on Mumford1 as well.  Interestingly, despite LC-Greedy's better performance at low $\alpha$, LC-100 obtains a slightly larger hypervolume than LC-Greedy on each city.

We also note that with Ahmed (2019), we have the HH algorithm as originally described by \cite{ahmed2019hyperheuristic}.  But the performance we obtain with this on both $C_p$ and $C_o$ is significantly worse than the results reported in that work on Mumford1, 2, and 3.  On Mumford3, the mean result we obtained at $\alpha=1.0$ is about 16\% greater in $C_p$, and at $\alpha=0.0$, it is about 37\% greater in $C_o$.

\section{Summary}

Across both sets of experiments, we see that using our learned policies to construct initial transit networks for metaheuristic improvement methods improve the results of those methods when optimizing transit networks, compared with the use of various human-engineered construction algorithms.  This is particularly true on large city graphs with 100 nodes or more, that is, graphs more like those of real-world cities: on the largest benchmark city, our initialization increased the hypervolume of solutions across $\alpha$ by 15\% versus the best other initialization method when using an evolutionary algorithm, and by 36\% versus the best other method when using the hyperheuristic algorithm.  

Having found that our learned policies can help with the second of Fan and Mumford's three factors in metaheuristic optimization, we turned next to the third factor: the choice of moves in search space.  We will explore whether the neural policies we have trained to construct transit networks can be repurposed to modify existing networks in a way that is useful to a metaheuristic algorithm.

\chapter{Neural Nets as Low-Level Heuristics}\label{chap:neighbourhood_moves}

The third factor that Fan and Mumford list as essential to the performance of a metaheuristic improvement algorithm is the set of neighbourhood moves that are available during search.  This move set is dictated by the operations used to modify solutions during the algorithm's run.  These operators are heuristics about what kinds of neighbourhood moves should be considered.  They are commonly referred to as the low-level heuristics, because they operate at the ``low'' level of modifying networks directly, by comparison with the metaheuristic (such as evolution or annealing) which operates at the ``high'' level of deciding which modified networks to keep and which to discard.  In the context of evolutionary algorithms specifically, it these heuristics are also often called ``mutators'', by analogy with mutation's role in biological evolution.  In this chapter, we consider whether the neural net policies described in \autoref{chap:neural_policy} can enact good low-level heuristics within a metaheuristic algorithm.

We designed the following neural-policy-based heuristic.  To modify a network $\mathcal{R}$, it selects a random route $r \in \mathcal{R}$ and deletes it from $\mathcal{R}$ to get $\mathcal{R}' = \mathcal{R} \setminus \{r\}$, and then runs learned construction with policy $\pi_\theta$ starting from $\mathcal{R}'$.  Since $|\mathcal{R}'| = S - 1$, this constructs just one new route $r'$ and adds it to $\mathcal{R}'$ to get the modified network: $\mathcal{R} \leftarrow \mathcal{R}' \cup \{r\}$.

To evaluate this learned low-level heuristic, we use it in a modified version of the evolutionary algorithm described in \autoref{chap:initialization}, where it replaces the type-1 mutator.  We replace the type-1 mutator because its space of changes (replacing one route by a shortest path) is similar to that of the learned heuristic (replacing one route by a new route composed of shortest paths), while the type-2 mutator's space of changes is quite different (lengthening or shortening a route by one node).

We refer to this modified algorithm as NEA, for ``neural evolutionary algorithm'', and the unmodified algorithm with the type-1 mutator as EA.

In \autoref{chap:initialization}, we found that using LC-100 to generate $\mathcal{R}_0$ in the evolutionary algorithm improved its performance, so this was how we initialized both EA and NEA, except where otherwise mentioned.  For NEA, we use the same policy $\pi_\theta$ in the LC-100 initialization as in the learned heuristic.

\section{Comparison with Baseline Evolutionary Algorithm}\label{sec:baseline}

We run each algorithm under consideration on all five of these synthetic cities over eleven different values of $\alpha$, ranging from $0.0$ to $1.0$ in increments of $0.1$.  This lets us observe how well the different methods perform under a range of possible preferences, from the extremes of the operator perspective ($\alpha = 0.0$, caring only about $C_o$) and passenger perspective ($\alpha = 1.0$, caring only about $C_p$) to a range of intermediate perspectives.  The constraint weight is set as $\beta=5.0$ in all experiments, the same value used in training the policies.  We found that this was sufficient to prevent any of the constraints in \autoref{subsec:constraints} from being violated by any transit network produced in our experiments.

Because these algorithms are stochastic, for each city we perform ten runs of each algorithm with ten different random seeds.  For algorithms that make use of a learned policy, ten separate policies were trained with the same set of random seeds (but using the same training dataset), and each was used when running algorithms with the corresponding random seed.  The values reported are statistics computed over the ten runs.

We first compared our neural evolutionary algorithm (NEA) with our baseline evolutionary algorithm (EA).  We also compare both with the initial networks from LC-100, to see how much improvement each algorithm makes over the initial networks.  In the EA and NEA runs, the same parameter settings of $B=10, I=400, F=10$ were used, following the values used in~\cite{nikolic2013transit}.  

\autoref{tab:baseline} displays the mean and standard deviation of the cost $C(\mathcal{G}, \mathcal{R})$ achieved by each algorithm on the Mandl and Mumford benchmarks for $\alpha=0.0$ (commonly referred to as the ``operator perspective'' in the literature), $\alpha=1.0$ (commonly referred to as the ``passenger perspective''), and $\alpha=0.5$ (which we refer to as the ``balanced perspective'').  We see that on Mandl, the smallest of the five cities, EA and NEA perform virtually identically for the operator and passenger perspectives, while NEA offers a slight improvement for the balanced perspective.  But for the larger cities of the Mumford dataset, NEA performs considerably better than EA for both the operator and balanced perspectives.  On Mumford3, the largest of the five cities, NEA solutions have 13\% lower average cost for the operator perspective and 5\% lower for the balanced perspective than EA. 

For the passenger perspective NEA's advantage over EA is smaller, but it still outperforms EA on Mumford1 and Mumford3 and performs comparably on Mumford2.  NEA's networks have 1\% lower average cost on Mumford3 than EA's, but the average costs of NEA and EA are within one standard deviation of each other.

\begin{table*}[]
	\centering
	\begin{tabular}{ccccccc}
		\toprule
		$\alpha$ & Environment & LC-100 & EA & NEA \\
		\midrule
		0.0 & Mandl    & 0.697 $\pm$ 0.011 & \bf 0.687 $\pm$ 0.016 & \bf 0.687 $\pm$ 0.016 \\
		& Mumford0 & 0.842 $\pm$ 0.021 & \bf 0.770 $\pm$ 0.022 & 0.771 $\pm$ 0.028 \\
		& Mumford1 & 1.795 $\pm$ 0.057 & 1.672 $\pm$ 0.039 & \bf 1.373 $\pm$ 0.063 \\
		& Mumford2 & 1.375 $\pm$ 0.103 & 1.118 $\pm$ 0.068 & \bf 0.950 $\pm$ 0.063 \\
		& Mumford3 & 1.405 $\pm$ 0.102 & 1.103 $\pm$ 0.070 & \bf 0.955 $\pm$ 0.062 \\
		\midrule
		0.5 & Mandl    & 0.560 $\pm$ 0.003 & 0.548 $\pm$ 0.007 & \bf 0.543 $\pm$ 0.009 \\
		& Mumford0 & 0.926 $\pm$ 0.005 & 0.854 $\pm$ 0.015 & \bf 0.853 $\pm$ 0.018 \\
		& Mumford1 & 1.307 $\pm$ 0.027 & 1.243 $\pm$ 0.032 & \bf 1.101 $\pm$ 0.016 \\
		& Mumford2 & 1.054 $\pm$ 0.033 & 0.917 $\pm$ 0.043 & \bf 0.855 $\pm$ 0.023 \\
		& Mumford3 & 1.046 $\pm$ 0.038 & 0.885 $\pm$ 0.027 & \bf 0.844 $\pm$ 0.027 \\
		\midrule
		1.0 & Mandl    & 0.334 $\pm$ 0.008 & \bf 0.317 $\pm$ 0.002 & 0.318 $\pm$ 0.003 \\
		& Mumford0 & 0.749 $\pm$ 0.021 & \bf 0.614 $\pm$ 0.007 & 0.619 $\pm$ 0.011 \\
		& Mumford1 & 0.601 $\pm$ 0.006 & 0.557 $\pm$ 0.006 & \bf 0.549 $\pm$ 0.005 \\
		& Mumford2 & 0.535 $\pm$ 0.012 & \bf 0.507 $\pm$ 0.003 & 0.507 $\pm$ 0.007 \\
		& Mumford3 & 0.508 $\pm$ 0.009 & 0.491 $\pm$ 0.006 & \bf 0.485 $\pm$ 0.007 \\
		\bottomrule
	\end{tabular}

	\caption{Final cost $C(\mathcal{G}, \mathcal{R})$ achieved baseline experiments for three different $\alpha$ values.  Values are averaged over ten random seeds; the $\pm$ value is the standard deviation of $C(\mathcal{G}, \mathcal{R})$ over the seeds.}
	\label{tab:baseline}
\end{table*}

\begin{figure*}
	\centering
	\includegraphics[width=0.85\textwidth]{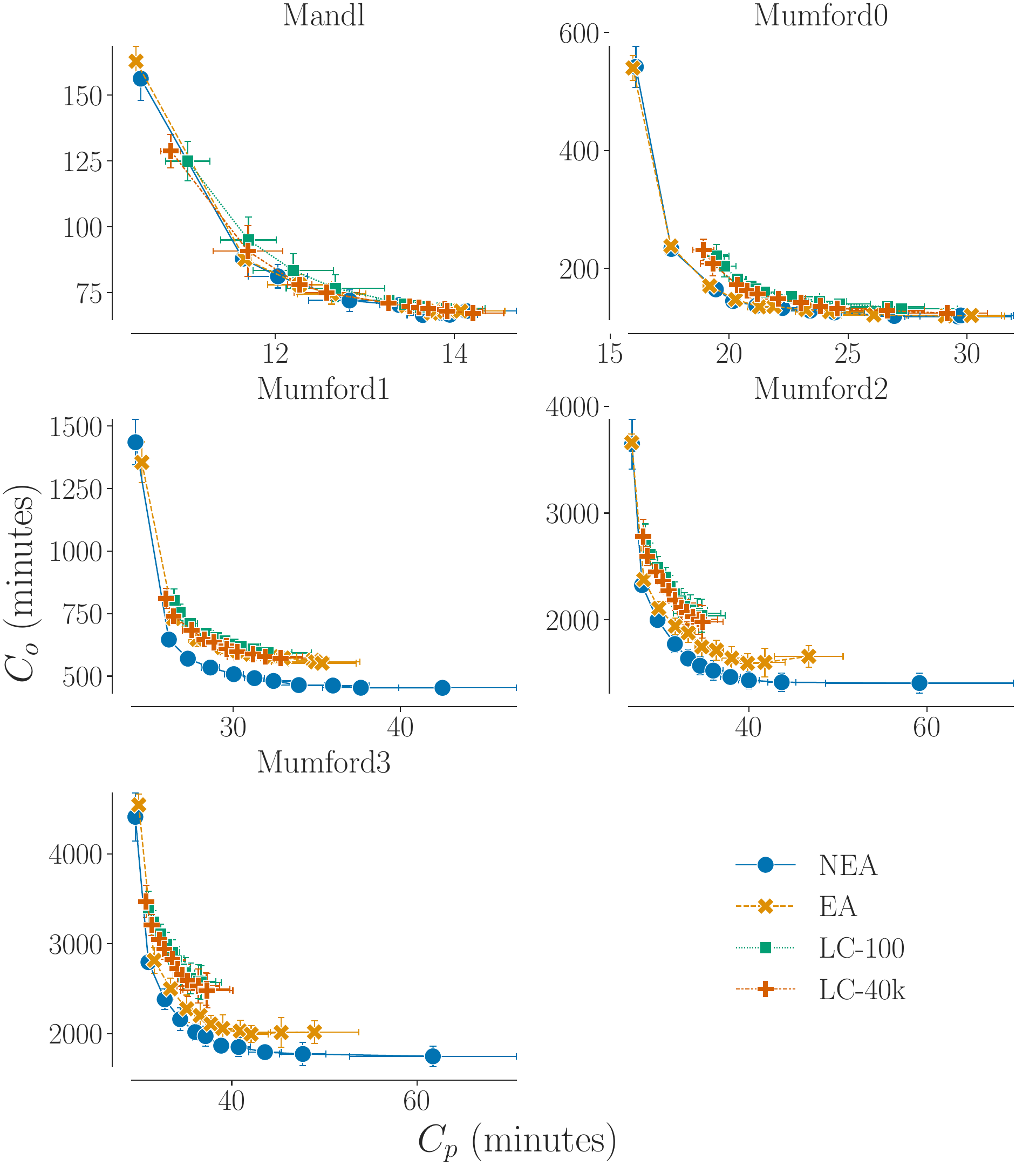}
	\caption{Values of average trip time $C_p$ (on the x-axis) and total route time $C_o$ (on the y-axis) achieved by NEA, EA, LC-100, and LC-40k, across values of $\alpha$ from $0$ at lower-right to $1$ at upper-left, in increments of $0.1$.  Each point is the mean over 10 random seeds for one value of $\alpha$, and bars around each point indicate one standard deviation.  Lines link pairs of points with consecutive $\alpha$ values.}
	\label{fig:40k}
\end{figure*}

\autoref{fig:40k} displays the average trip time $C_p$ and total route time $C_o$ achieved by each algorithm on the three largest Mumford cities, the three which are each based on a real city's statistics, over eleven $\alpha$ values evenly spaced over the range $[0, 1]$ in increments of 0.1.  There is a necessary trade-off between $C_p$ and $C_o$, and as we would expect, as $\alpha$ increases, $C_o$ increases and $C_p$ decreases for each algorithm's output.  We also observe that for intermediate values of $\alpha$, NEA pushes $C_o$ much lower than LC-100, at the cost of increases in $C_p$.  This behaviour is desirable, because as the figure shows, NEA achieves an overall larger range of outcomes than LC-100 - for any network $\mathcal{R}$ from LC-100, there is some value of $\alpha$ for which NEA produces a network $\mathcal{R}'$ which dominates $\mathcal{R}$.  \autoref{fig:40k} also shows results for an additional algorithm, LC-40k, which we discuss in \autoref{sec:ablations}.

We observe a common pattern on all Mumford1, 2, and 3: EA and NEA perform very similarly at $\alpha=1.0$ (the leftmost point on each curve), but a significant performance gap forms as $\alpha$ decreases - consistent with what we see in \autoref{tab:baseline}.  We also observe that LC-100 favours reducing $C_p$ over $C_o$, with its points clustered higher and more leftwards than most of the points with corresponding $\alpha$ values on the other curves.  Both EA and NEA achieve only very small decreases in $C_p$ on LC-100's initial networks at $\alpha=1.0$, and they do so by increasing $C_o$ considerably.

\begin{figure*}
	\centering
	\includegraphics[width=\textwidth]{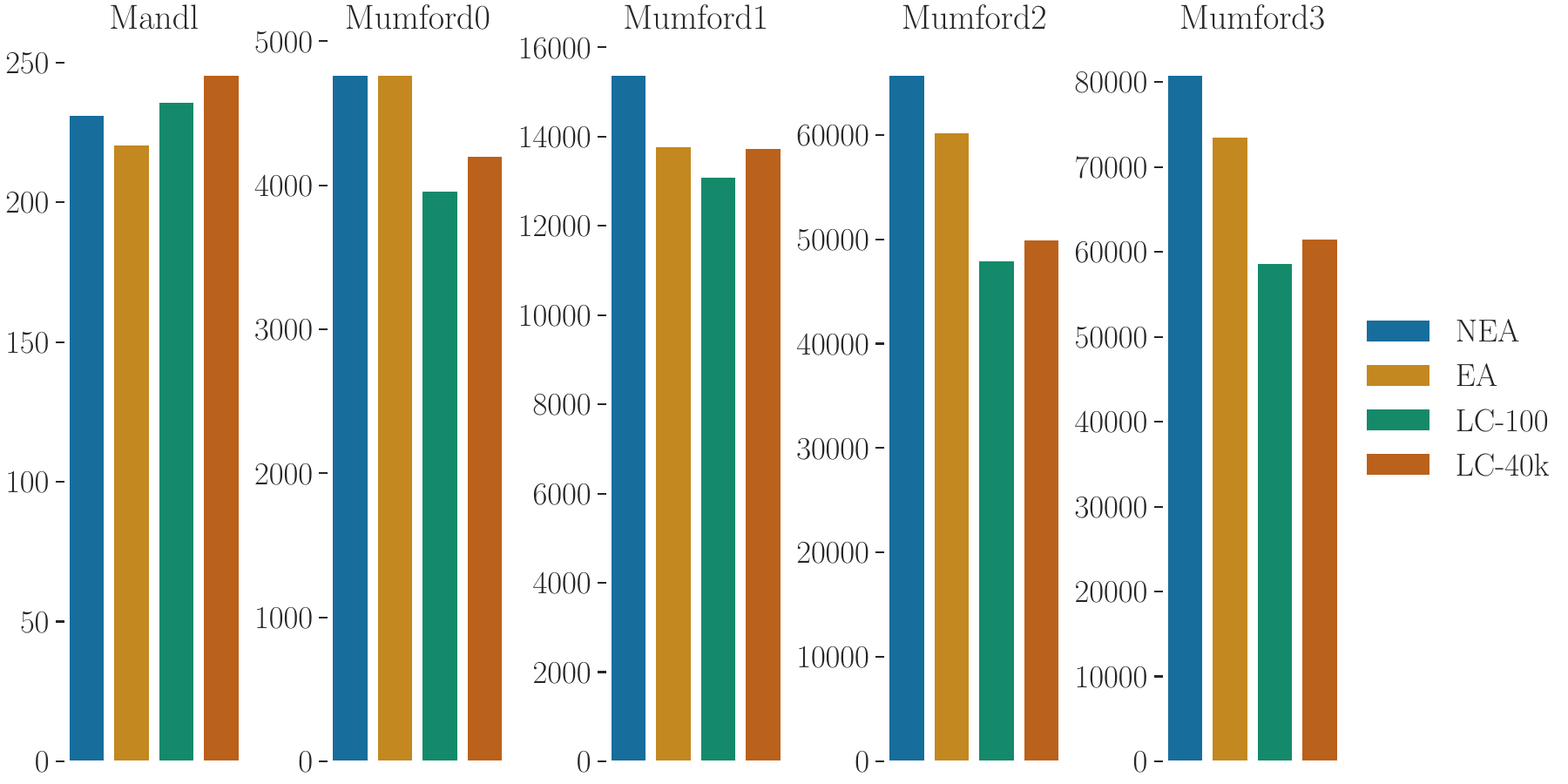}
	\caption{Hypervolumes achieved by LC-100, LC-Greedy, EA, and NEA.}
	\label{fig:40k_hv}
\end{figure*}

\autoref{fig:40k_hv} corroborates what we observe in \autoref{tab:baseline} and \autoref{fig:40k}: introducing the neural mutator improves the performance of the evolutionary algorithm by a significant margin on the three largest benchmark cities, across the range of $\alpha$ values.


\section{Ablation Studies}\label{sec:ablations}

To better understand the contribution of various components of our method, we performed three sets of ablation studies.  These were conducted over the three realistic Mumford cities (1, 2, and 3) and over eleven $\alpha$ values evenly spaced over the range $[0, 1]$ in increments of 0.1.  

\subsection{Effect of number of samples}

We note that over the course of the evolutionary algorithm, with our parameter settings $B=10, I=400, F=10$, a total of $B \times I \times E = 40,000$ different transit networks are considered.  By comparison, LC-100 only considers 100 networks.  It could be that NEA's superiority to LC-100 is only due to its considering more networks.  To test this, we ran LC-40k, in which we sample $40,000$ networks from the learned-construction algorithm, and pick the lowest-cost network.  Comparing LC-100 and LC-40k in \autoref{fig:40k}, we see that across all three cities and all values of $\alpha$, LC-40k performs very similarly to LC-100, improving on it only slightly in comparison with the larger improvements given by EA or NEA.  \autoref{fig:40k_hv} shows further that taking the best of 40,000 samples instead of 100 makes only a slight improvement to the quality of the networks found, in comparison with the improvement of NEA over EA.  From this we conclude that the number of networks considered is not on its own an important factor in EA's and NEA's performance: much more important is the evolutionary algorithm that guides the search of possible networks.

\subsection{Contribution of type-2 mutator}

We next considered the impact of the type-2 mutator heuristic on NEA.  This low-level heuristic is the same between EA and NEA and is not a learned function.  To understand how important it is to NEA's performance, we ran ``all-1 NEA'', a variant of NEA in which only the neural mutator is used, and not the type-2 mutator.  
\begin{figure*}
	\centering
	\includegraphics[width=0.85\textwidth]{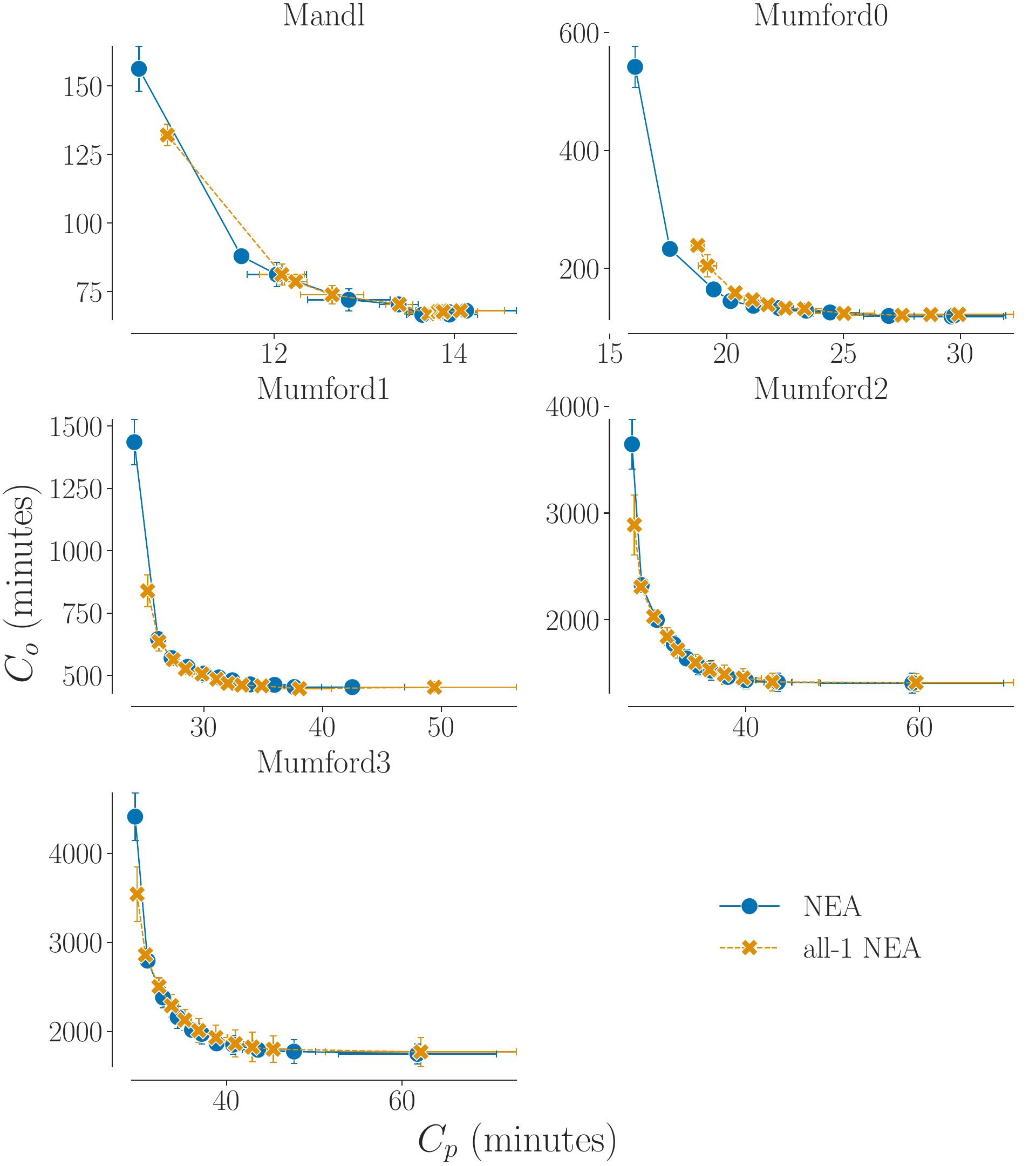}
	\caption{Values of average trip time $C_p$ (on the x-axis) and total route time $C_o$ (on the y-axis) achieved by all-1 NEA, plotted along with EA and NEA (repeated from \autoref{fig:40k}) for comparison.}
	\label{fig:no2}
\end{figure*}

\begin{figure*}
	\centering
	\includegraphics[width=\textwidth]{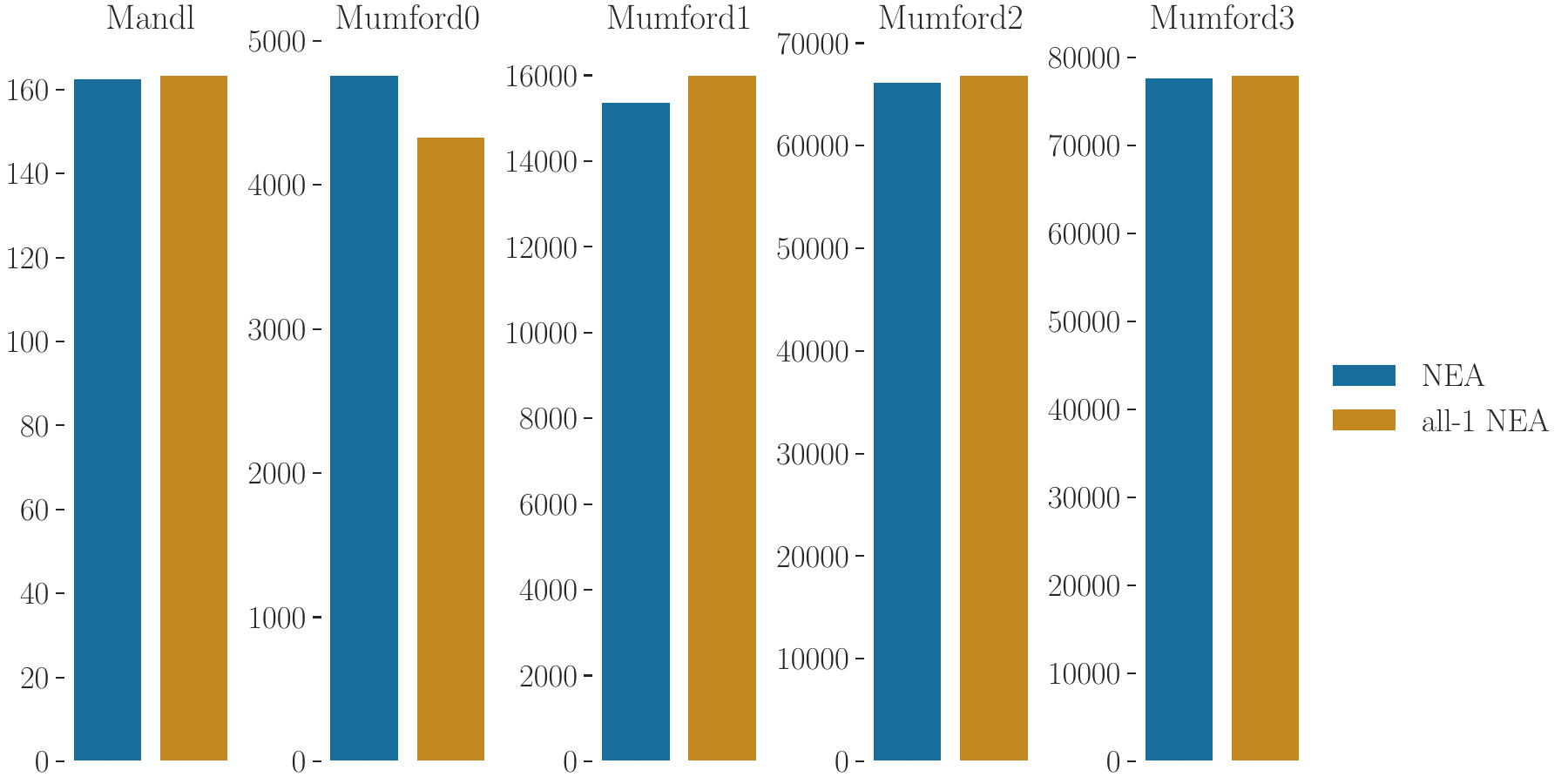}
	\caption{Hypervolumes achieved by NEA vs NEA with only the type-1 mutator.}
	\label{fig:no2_hv}
\end{figure*}

The results are shown in \autoref{fig:no2}, along with the curve for NEA as in \autoref{fig:40k}, for comparison.  It is clear that NEA and all-1 NEA perform very similarly in most scenarios.  The most notable difference is at $\alpha=1.0$, where we see that all-1 NEA underperforms NEA, achieving higher values of $C_p$.  It appears that the minor adjustments to existing routes made by the type-2 mutator are more important at extremes of $\alpha$, where the neural net policy has been pushed to the extremes of its behaviour.

On Mumford3, we observe that the curves do not overlap as cleanly: NEA seems to slightly outperform all-1 NEA at most $\alpha$ values.  So the contribution of the type-2 mutator appears to grow as the city gets larger and more challenging.  This may be because as the city gets larger, the space of possible routes grows larger, so the neural mutator's changes get more dramatic in comparison with the single-node adjustments of the type-2 mutator.  The type-2 mutator's class of changes becomes more distinct from those of the neural mutator, possibly making it more important.  

\autoref{fig:no2_hv} compares the hypervolumes of NEA and all-1 NEA across the five benchmark cities, and confirms that their performance is overall extremely similar, except on Mumford0 where NEA has a more pronounced edge.  This suggests that keeping the type-2 mutator is not a disadvantage on large cities, and may help on smaller ones.  The differences are very small in any case, so we retain this mutator in NEA in future experiments.  This does confirm, however, that between the two mutators, the primary contributor to the NEA's quality is the neural mutator.

\subsection{Importance of learned heuristics}\label{subsec:rcea}

We note that the type-1 mutator used in EA and the neural mutator used in NEA differ in the space of changes each is capable of making.  The type-1 mutator can only add shortest paths as routes, while the neural mutator may compose new routes from multiple shortest paths.  It may be that this structural difference, as opposed to the quality of the heuristics learned by the policy $\pi_\theta$, is part of NEA's advantage over EA.

To test this conjecture, we ran a variant of NEA in which the learned policy $\pi_\theta$ is replaced by the uniformly-random policy $\pi_\text{random}$, previously described in \autoref{chap:neural_policy}.  We call this variant the random-construction evolutionary algorithm (RC-EA).  Since we wanted to gauge the performance of this variant in isolation from the learned policy, we used RC-100 to generate the initial network $\mathcal{R}_0$, instead of LC-100 with $\pi_\theta$.

\begin{figure*}
	\centering
	\includegraphics[width=0.85\textwidth]{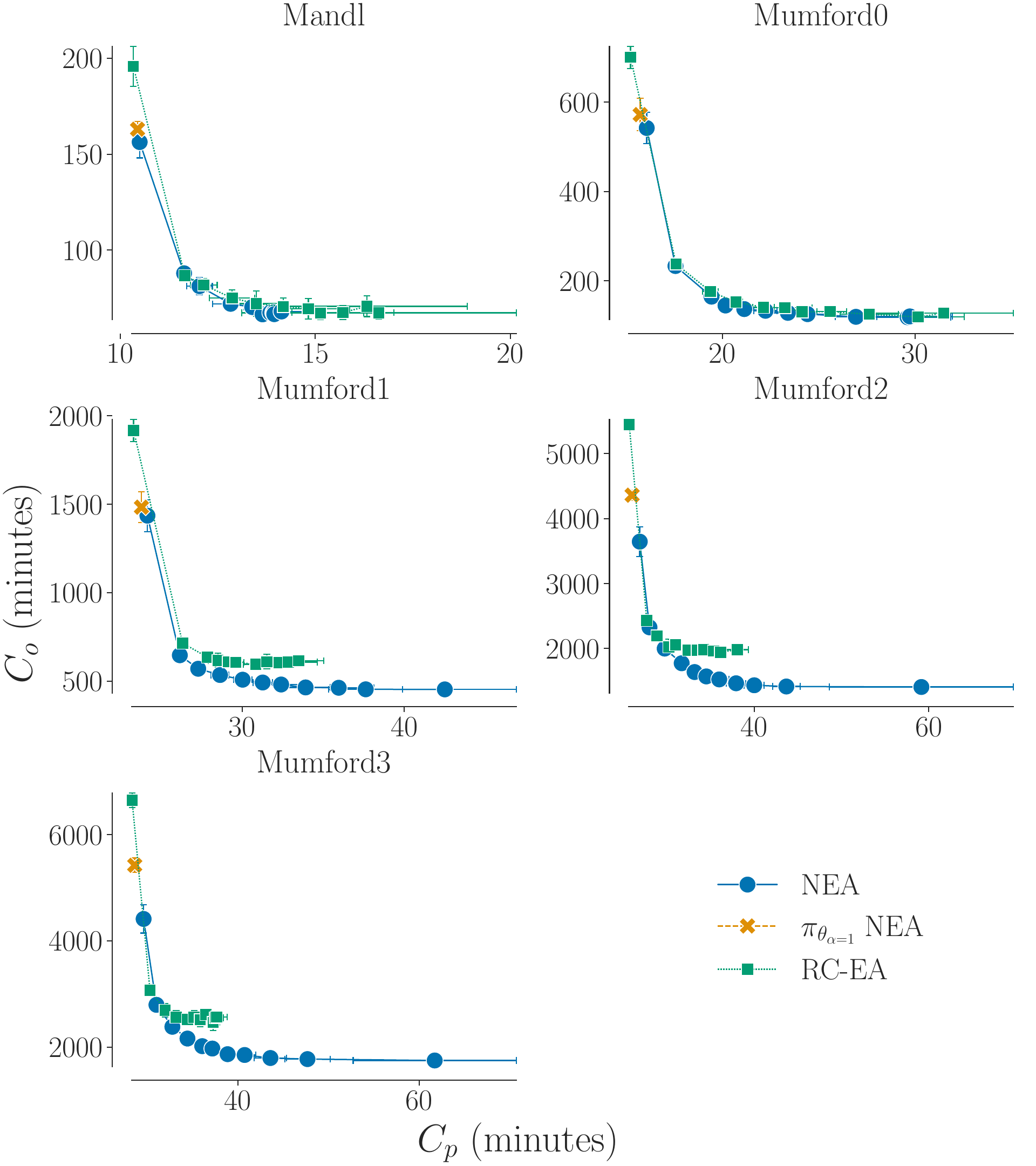}
	\caption{Trade-offs between average trip time $C_p$ (on the x-axis) and total route time $C_o$ (on the y-axis) achieved by RC-EA, plotted along with NEA (repeated from \autoref{fig:40k}) for comparison.}
	\label{fig:random}
\end{figure*}

\begin{figure*}
	\centering
	\includegraphics[width=\textwidth]{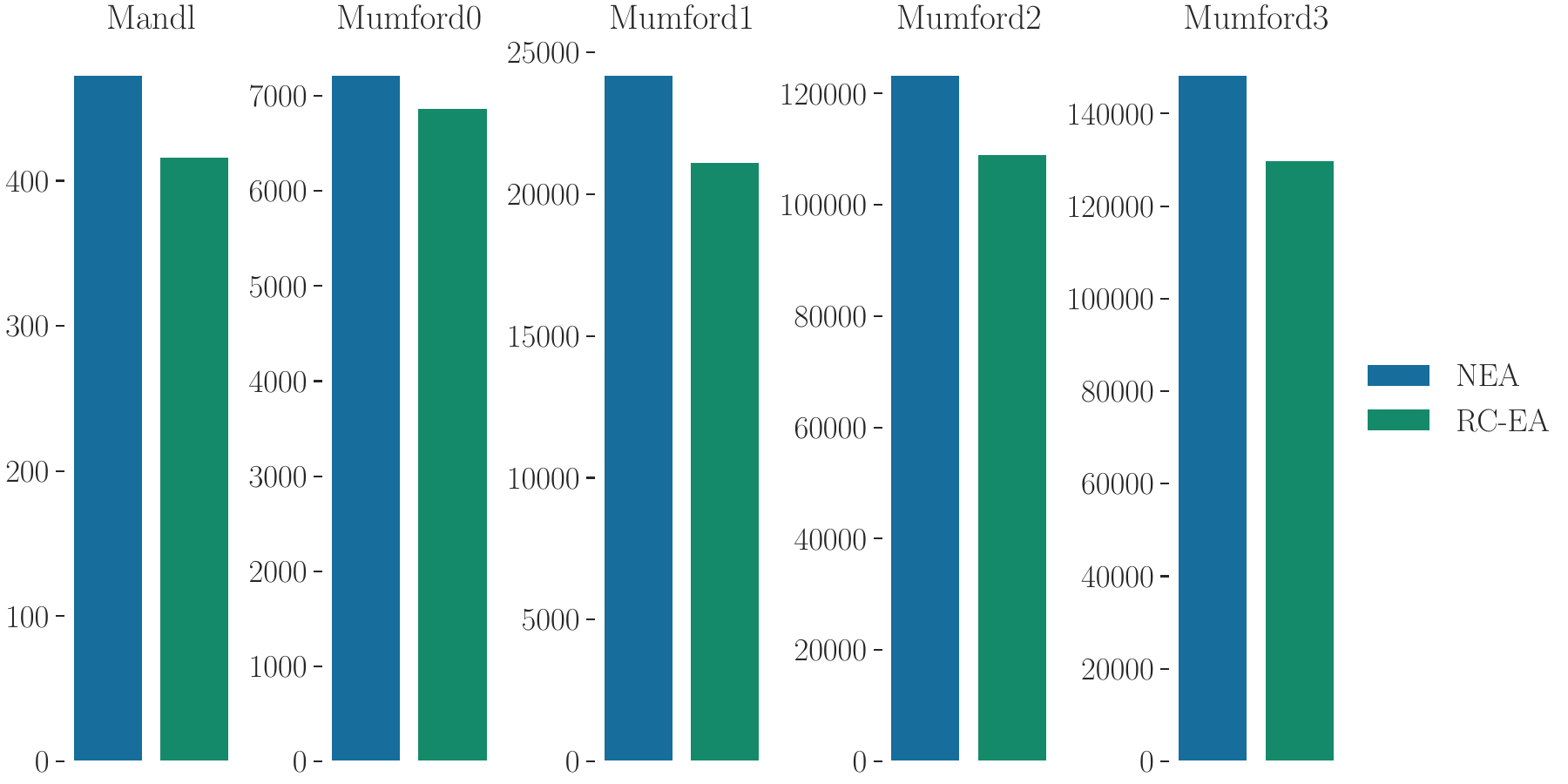}
	\caption{Hypervolumes achieved by NEA and RC-EA with only the type-1 mutator.}
	\label{fig:random_hv}
\end{figure*}

The results are shown in \autoref{fig:random}, along with the same NEA curves as in \autoref{fig:40k}.  For most values of $\alpha$, RC-EA performs less well than NEA, and \autoref{fig:random_hv} confirms that its hypervolume is significantly smaller on each city.  But at the extreme of $\alpha = 1.0$ it performs better than NEA, decreasing average trip time $C_p$ by between one and two minutes versus NEA on each city - an improvement of about 3.7\% on Mumford1 and about 4.5\% on Mumford2 and Mumford3.

RC-EA's poor performance at low $\alpha$ makes sense:  $\pi_\text{random}$, which replaces routes with composites of shortest paths, is biased towards making long routes.  Unlike $\pi_\theta$, which can learn to increase its halting probability with decreasing $\alpha$,  $\pi_\text{random}$'s halting probability is fixed and independent of $\alpha$.  This is an advantage over NEA at $\alpha=1.0$ but a disadvantage at $\alpha$ less than about $0.8$.  It is interesting, though, that RC-EA outperforms NEA at $\alpha=1.0$, as these two share the same space of possible actions.  This calls to mind what we observed in \autoref{subsec:nn_eval}, which was that the superior performance at $\alpha=1.0$ of $\pi_\text{random}$ over $\pi_\theta$ on the transit network construction task.  In that instance, we established that this was because $\pi_\theta$, being trained over a range of $\alpha$ values, had not learned to perform as well as possible at the extreme of $\alpha = 1.0$.  With parameters $\theta_{\alpha=1}$ trained with only $\alpha=1.0$, $\pi_{\theta_{\alpha=1}}$ outperformed $\pi_\text{random}$ at $\alpha = 1.0$.

To test whether the same thing was happening here, we ran another set of NEA experiments, this time using the same policy parameters $\theta_{\alpha=1}$ that were trained in \autoref{subsec:nn_eval}, both for the LC-100 initialization and for NEA's neural mutator.  These experiments were run only at $\alpha=1.0$, as we would not expect $\pi_{\theta_{\alpha=1}}$ to perform well for values of $\alpha$ it never saw during training.  The results are displayed in \autoref{fig:random} alongside those of RC-EA.  We observe that for all four Mumford cities, NEA with $\pi_{\theta_{\alpha=1}}$ at $\alpha=1.0$ achieves lower $C_p$ than NEA with $\pi_\theta$, but not quite as low as RC-EA.  Evidently, $\pi_\text{random}$ still offers advantages over our learned models in the context of this evolutionary algorithm.

In principle, our policy could learn to perform as well as RC-EA by simply learning to choose actions uniformly at random.  Yet that is evidently not the policy it has learned.  In fact, as we observe in \autoref{fig:nn_extremes}, $\pi_{\theta_{\alpha=1}}$ does outperform $\pi_\text{random}$ in the context of a construction process (LC-100 and RC-100), which is nearly the same process as the \gls{mdp} in which $\pi_{\theta_{\alpha=1}}$ was trained.  So $\theta_{\alpha=1}$ learned some policy that was better than a uniformly-random policy in its training context.  But \autoref{fig:random} shows that this superiority does not translate to the evolutionary algorithm, which is quite different from the training context.  This difference in context may explain why NEA $\pi_{\theta_{\alpha=1}}$ does not perform as well as RC-EA at $\alpha=1.0$.  

To test this would require designing a different training process, one where the \gls{mdp} would more closely resemble the improvement process of EA than the construction process of LC-100 and LC-Greedy.  This would be worthwhile but would require a major investment of time, and so lies beyond the scope of this thesis.  Furthermore, as \autoref{fig:random_hv} confirms, NEA with $\pi_\theta$ performs better than RC-EA when considering the whole range of $\alpha$, so we do not think this a critical flaw in our approach.  For now we merely note that composing shortest paths is a good heuristic for route construction at high $\alpha$, but is worse than simply choosing shortest paths at low $\alpha$; so the advantage of NEA comes primarily from what the policy $\pi_\theta$ learns during training.

\section{Comparison with State of the Art Methods}\label{sec:vs_prior}

While conducting the experiments of \autoref{sec:baseline}, we observed that after 400 iterations the cost of NEA's best-so-far network $\mathcal{R}_\text{best}$ was still decreasing from one iteration to the next.  In order to make a fairer comparison with other methods from the literature on the Mandl and Mumford benchmarks, we performed a further set of experiments where we ran NEA with $I=4,000$ instead of $400$, for the operator perspective ($\alpha=0.0$) and the passenger perspective ($\alpha=1.0$).  In addition, because we observed better performance from RC-EA for $\alpha$ near $1.0$, we also ran RC-EA with $I=4,000$ for $\alpha=1.0$, and we include these results in the comparison.

\nomenclature{$d_0$}{The percentage of passenger trips which are made with no transfers between routes, given a city graph and a transit network.}
\nomenclature{$d_1$}{The percentage of passenger trips which are made with 1 transfer between routes, given a city graph and a transit network.}
\nomenclature{$d_2$}{The percentage of passenger trips which are made with 2 transfers between routes, given a city graph and a transit network.}
\nomenclature{$d_{un}$}{The percentage of passenger trips which are made with more than 2 transfers between routes, given a city graph and a transit network.}

\autoref{tab:benchmark_pp} and \autoref{tab:benchmark_op} present the results of these experiments on the Mandl and Mumford benchmarks, alongside results reported on these benchmarks in comparable recent work.  As elsewhere, the results we report for NEA and RC-EA are averaged over ten runs with different random seeds; the same set of ten trained policies used in the experiments of \autoref{sec:baseline} and \autoref{sec:ablations} are used here in NEA.  Results from other work are as reported in that work.  In addition to the average trip time $C_p$ and total route time $C_o$, the tables present metrics of how many transfers between routes were required by the networks.  These are labelled $d_0$, $d_1$, $d_2$, and $d_{un}$.  $d_i$ with $i \in \{0, 1, 2\}$ is the percentage of all passenger trips that required $i$ transfers between routes; while $d_{un}$ is the percentage of trips that require more than 2 transfers ($d_{un} = 100 - (d_0 + d_1 + d_2)$).


Before discussing these results, a brief word should be said about computation time.  On a desktop computer with a 2.4 GHz Intel i9-12900F processor and an NVIDIA RTX 3090 graphics processing unit (used to accelerate our neural net computations), NEA takes about 10 hours for each 4,000-iteration run on Mumford3, the largest environment, while the RC-EA runs take about 6 hours.  By comparison, \cite{john2014routing} and \cite{husselmann2023improved} both use a variant of NSGA-II, a genetic algorithm, with a population of 200 networks, which in both cases takes more than two days to run on Mumford3. \cite{husselmann2023improved}'s DBMOSA variant, meanwhile, takes 7 hours and 52 minutes to run on Mumford3.  \cite{kilic2014demand} report that their procedure takes eight hours just to construct the initial network for Mumford3, and don't report the running time for the subsequent optimization.

These reported running times 
are mainly a function of the metaheuristic algorithm used, rather than the low-level heuristics used in the algorithm.  Genetic algorithms like NSGA-II, with large populations as used in some of these methods, are very time-consuming because of the large number of networks they must modify and evaluate at each step.  But their search of solution space is correspondingly more exhaustive than single-solution methods such as simulated annealing, or an evolutionary algorithm with a small population ($B=10$) as we use in our own experiments.  It is therefore to be expected that they would achieve lower final costs in exchange for their greater run-time.  We are interested here in the quality of our low-level heuristics, rather than that of the metaheuristic algorithm, and this must be kept in mind as we discuss the results of this section.

\subsubsection{Caveat for Ahmed et al. (2019)}

In these comparisons, we include results from~\cite{ahmed2019hyperheuristic}, but we wish to state a caveat to these.  While they reported results that set a new state of the art on these benchmarks, they do not provide the values used for the two parameters to their algorithm.  In our earlier work~\cite{holliday2023augmenting}, theirs was one of several methods we implemented in order to test our proposed initialization scheme.  We discovered parameter values on our own that gave results comparable to other work up to 2019, but after much effort and correspondence with one of the authors of \cite{ahmed2019hyperheuristic}, we were unable to replicate their state-of-the-art results.

\begin{table*}
	\centering
	\resizebox{0.9\textwidth}{!}{	
	\begin{tabular}{llllllll}
		\toprule
		City & Method & $C_p \downarrow$ & $C_o$ &  $d_0 \uparrow$ &  $d_1$ &  $d_2$ & $d_{un} \downarrow$ \\
		\midrule
		Mandl & \cite{mumford2013new} &  10.27 &   221 &  95.38 &   4.56 &   0.06 & \bf 0 \\
		& \cite{john2014routing} &  10.25 &   212 &      - &      - &      - &        - \\
		& \cite{kilic2014demand} &  10.29 &   216 &   95.5 &    4.5 &      0 &        \bf 0 \\
		& \cite{ahmed2019hyperheuristic} & \bf 10.18  & 212  & 97.17  & 2.82   & 0.00   & \bf 0 \\
		& \cite{husselmann2023improved} DBMOSA &  10.27 &   179 &  95.94 &   3.93 &   0.13 & \bf 0 \\
		& \cite{husselmann2023improved} NSGA-II &  10.19 &   197 &  \bf 97.36 &   2.64 &      0 & \bf 0 \\
		& NEA &  10.47 &   157 &  91.89 &   7.55 &   0.52 & 0.04 \\
		& RC-EA &  10.32 &   194 &  94.15 &   5.74 &   0.11 & \bf 0 \\
		\midrule
		Mumford0 & \cite{mumford2013new} &  16.05 &   759 &   63.2 &  35.82 &   0.98 & \bf 0 \\
		& \cite{john2014routing} &   15.4 &   745 &      - &      - &      - &  - \\
		& \cite{kilic2014demand} &  14.99 &   707 &  69.73 &  30.03 &   0.24 & \bf 0 \\
		& \cite{ahmed2019hyperheuristic} & \bf 14.09  & 722  & \bf 88.74  & 11.25   & 0   & \bf 0  \\
		& \cite{husselmann2023improved} DBMOSA &  15.48 &   431 &   65.5 &   34.5 & 0 & \bf 0 \\
		& \cite{husselmann2023improved} NSGA-II & 14.34 &   635 & 86.94 &  13.06 & 0 & \bf 0 \\
		& NEA &  16.00 &   550 &  49.74 &  43.85 &   6.38 & 0.02 \\
		& RC-EA &  14.96 &   722 &  63.21 &  36.36 &   0.43 & \bf 0 \\
		\midrule
		Mumford1 & \cite{mumford2013new} &  24.79 &  2038 &   36.6 &  52.42 &  10.71 &     0.26 \\
		& \cite{john2014routing} &  23.91 &  1861 &      - &      - &      - &        - \\
		& \cite{kilic2014demand} &  23.25 &  1956 &   45.1 &  49.08 &   5.76 &     0.06 \\
		& \cite{ahmed2019hyperheuristic} & \bf 21.69  & 1956  & \bf 65.75  & 34.18   & 0.07   & \bf 0  \\
		& \cite{husselmann2023improved} DBMOSA &  22.31 &  1359 &  57.14 &  42.63 &   0.23 & \bf 0 \\
		& \cite{husselmann2023improved} NSGA-II & 21.94 &  1851 &  62.11 &  37.84 &   0.05 & \bf 0 \\
		& NEA &  24.07 &  1450 &  33.04 &  45.64 &  18.89 & 2.43 \\
		& RC-EA &  23.01 &  1924 &  39.57 &  49.66 &  10.46 &     0.32 \\
		\midrule
		Mumford2 & \cite{mumford2013new} &  28.65 &  5632 &  30.92 &  51.29 &  16.36 &     1.44 \\
		& \cite{john2014routing} &  27.02 &  5461 &      - &      - &      - &        - \\
		& \cite{kilic2014demand} &  26.82 &  5027 &  33.88 &  57.18 &   8.77 &     0.17 \\
		& \cite{ahmed2019hyperheuristic} & \bf 25.19  & 5257  & \bf 56.68  & 43.26   & 0.05   & \bf 0 \\
		& \cite{husselmann2023improved} DBMOSA &  25.65 &  3583 &  48.07 &  51.29 &   0.64 & \bf 0 \\
		& \cite{husselmann2023improved} NSGA-II & 25.31 &  4171 &  52.56 &  47.33 &   0.11 & \bf 0 \\
		& NEA &  26.52 &  4017 &  32.74 &  49.94 &  16.42 & 0.90 \\
		& RC-EA &  25.45 &  5536 &  41.18 &  52.96 &   5.84 &     0.02 \\
		\midrule
		Mumford3 & \cite{mumford2013new} &  31.44 &  6665 &  27.46 &  50.97 &  18.79 &     2.81 \\
		& \cite{john2014routing} &   29.5 &  6320 &      - &      - &      - &        - \\
		& \cite{kilic2014demand} &  30.41 &  5834 &  27.56 &  53.25 &  17.51 &     1.68 \\
		& \cite{ahmed2019hyperheuristic} & 28.05  & 6119  & \bf 50.41  & 48.81   & 0.77   & \bf 0 \\
		& \cite{husselmann2023improved} DBMOSA &  28.22 &  4060 &  45.07 &  54.37 &   0.56 & \bf 0 \\
		& \cite{husselmann2023improved} NSGA-II & \bf 28.03 &  5018 &  48.71 &   51.1 &   0.19 & \bf 0 \\
		& NEA &  29.18 &  5122 &  30.70 &  52.27 &  16.17 & 0.87 \\
		& RC-EA &  28.09 &  6830 &   38.6 &  57.02 &   4.35 &     0.03 \\
		\bottomrule
	\end{tabular}
	}
	\caption{Passenger-perspective results.  $C_p$ is the average passenger trip time. $C_o$ is the total route time.  $d_i$ is the percentage of trips satisfied with number of transfers $i$, while $d_{un}$ is the percentage of trips satisfied with 3 or more transfers.  Arrows next to each quantity indicate which of increase or decrease is desirable.  Bolded values in $C_p$, $d_0$, and $d_{un}$ columns are the best on that environment.}
	\label{tab:benchmark_pp}
\end{table*}

\subsubsection{Passenger-perspective results}

\autoref{tab:benchmark_pp} shows the passenger perspective results alongside results from other work.  On each city except Mumford3, the hyper-heuristic method of \cite{ahmed2019hyperheuristic} performs best, while on Mumford3 the NSGA-II variant of~\cite{husselmann2023improved} performs best.  RC-EA and NEA both perform poorly on the smallest two cities, Mandl and Mumford0, but their relative performance improves as the size of the city increases.  On both Mumford2 and Mumford3, RC-EA's performance is very close to that of~\cite{husselmann2023improved}'s and \cite{ahmed2019hyperheuristic}'s, and better than all other methods listed, despite the relatively under-powered metaheuristic that drives it.  

RC-EA also outperforms \cite{husselmann2023improved}'s DBMOSA, which uses the same low-level heuristics as their NSGA-II with a faster but less-exhaustive metaheuristic.  That RC-EA exceeds DBMOSA's performance on Mumford2 and Mumford3 is evidence that the low-level heuristics used in RC-EA are better for the passenger perspective than those proposed by \cite{husselmann2023improved}.  This is especially the case given that DBMOSA is still a more sophisticated metaheuristic algorithm than ours, with ten different low-level heuristics and a hyper-heuristic that adapts the rate at which each low-level heuristic is applied over the run - though we acknowledge that DBMOSA is a multi-objective optimization algorithm, which may account for its relative weakness here.

NEA does not perform as well as RC-EA on the passenger perspective (aligning with what we observed in \autoref{sec:ablations}), but still shows good performance on Mumford2 and 3, outperforming all of these methods that were published prior to 2019.

\subsection{Operator-perspective results}

\autoref{tab:benchmark_op} shows the operator-perspective results alongside results from other work.  \cite{kilic2014demand} do not report results for the operator perspective and as such we do not include their work in \autoref{tab:benchmark_op}.  Similarly to the passenger-perspective results, our methods underperform on the smallest cities (Mandl and Mumford0) but perform well on larger ones.  Strikingly, 
NEA achieves the lowest value of total route time $C_o$ out of all methods on Mumford3, improving even on \cite{ahmed2019hyperheuristic}, and outperforms all other methods on Mumford1 and Mumford2 as well.  Evidently, the learned policy functions very well as a low-level heuristic at low values of $\alpha$, where the premium is on keeping routes short.

\begin{table*}
	\centering
	\resizebox{0.9\textwidth}{!}{	
	\begin{tabular}{llllllll}
		\toprule
		City & Method & $C_o \downarrow$ & $C_p$ & $d_0 \uparrow$ & $d_1$ & $d_2$ & $d_{un} \downarrow$ \\
		\midrule
		Mandl & \cite{mumford2013new} & \bf 63 & 15.13 & 70.91 & 25.5 & 2.95 & 0.64 \\
		& \cite{john2014routing} & \bf 63 & 13.48 & - & - & - & - \\
		& \cite{ahmed2019hyperheuristic} & \bf 63 & 14.28  &  62.23 & 27.16  & 9.57  & 1.03 \\
		& \cite{husselmann2023improved} DBMOSA & \bf 63 & 13.55 & 70.99 & 24.44 & 4.00 & \bf 0.58 \\
		& \cite{husselmann2023improved} NSGA-II & \bf 63 & 13.49 & \bf 71.18 & 25.21 & 2.97 & 0.64 \\
		& NEA & 68 & 14.13 & 57.28 & 30.56 & 11.84 & \bf 0.31 \\
		\midrule
		Mumford0 & \cite{mumford2013new} & 111 & 32.4 & 18.42 & 23.4 & 20.78 & 37.40 \\
		& \cite{john2014routing} & 95 & 32.78 & - & - & - & - \\
		& \cite{ahmed2019hyperheuristic} & \bf 94  & 26.32  & 14.61  & 31.59   & 36.41  & 17.37 \\
		& \cite{husselmann2023improved} DBMOSA & 98 & 27.61 & 22.39 & 31.27 & 18.82 & 27.51 \\
		& \cite{husselmann2023improved} NSGA-II & \bf 94 & 27.17 & \bf 24.71 & 38.31 & 26.77 & \bf 10.22 \\
		& NEA & 120 & 29.73 & 15.25 & 31.14 & 28.58 & 25.03 \\
		\midrule
		Mumford1 & \cite{mumford2013new} & 568 & 34.69 & 16.53 & 29.06 & 29.93 & 24.66 \\
		& \cite{john2014routing} & 462 & 39.98 & - & - & - & - \\
		& \cite{ahmed2019hyperheuristic} & \bf 408  & 39.45  & 18.02  & 29.88  & 31.9  & 20.19 \\
		& \cite{husselmann2023improved} DBMOSA & 511 & 26.48 & \bf 25.17 & 59.33 & 14.54 & \bf 0.96 \\
		& \cite{husselmann2023improved} NSGA-II & 465 & 31.26 & 19.70 & 42.09 & 33.87 & 4.33 \\
		& NEA & 437 & 49.37 & 13.92 & 21.51 & 22.42 & 42.15 \\
		\midrule
		Mumford2 & \cite{mumford2013new} & 2244 & 36.54 & 13.76 & 27.69 & 29.53 & 29.02 \\
		& \cite{john2014routing} & 1875 & 32.33 & - & - & - & - \\
		& \cite{ahmed2019hyperheuristic} & \bf 1330  & 46.86  & 13.63  & 23.58   & 23.94  & 38.82 \\ 
		& \cite{husselmann2023improved} DBMOSA & 1979 & 29.91 & \bf 22.77 & 58.65 & 18.01 & \bf 0.57 \\
		& \cite{husselmann2023improved} NSGA-II & 1545 & 37.52 & 13.48 & 36.79 & 34.33 & 15.39 \\
		& NEA & 1365 & 61.67 & 8.34 & 15.33 & 18.71 & 57.62 \\
		\midrule
		Mumford3 & \cite{mumford2013new} & 2830 & 36.92 & 16.71 & 33.69 & 33.69 & 20.42 \\
		& \cite{john2014routing} & 2301 & 36.12 & - & - & - & - \\
		& \cite{ahmed2019hyperheuristic} & 1746  & 46.05  & 16.28  & 24.87   & 26.34  & 32.44 \\      
		& \cite{husselmann2023improved} DBMOSA & 2682 & 32.33 & \bf 23.55 & 58.05 & 17.18 & \bf 1.23 \\
		& \cite{husselmann2023improved} NSGA-II & 2043 & 35.97 & 15.02 & 48.66 & 31.83 & 4.49 \\
		& NEA & \bf 1697 & 66.04 & 8.26 & 13.46 & 16.66 & 61.63 \\
		\bottomrule
	\end{tabular}
	}
	\caption{Operator perspective results.  $C_p$ is the average passenger trip time. $C_o$ is the total route time.  $d_i$ is the percentage of trips satisfied with number of transfers $i$, while $d_{un}$ is the percentage of trips satsified with 3 or more transfers.  Arrows next to each quantity indicate which of increase or decrease is desirable.  Bolded values in $C_o$, $d_0$ and $d_{un}$ columns are the best on that environment.}
	\label{tab:benchmark_op}
\end{table*}

\subsection{Transfer Statistics}

The metrics $d_0$ to $d_{un}$ reveal that both RC-EA and NEA favour higher numbers of transfers relative to most of the other methods, particularly \cite{husselmann2023improved}.  This is true for both the passenger and operator perspectives.  For the operator perspective, this matches our intuitions: shorter routes will deliver fewer passengers directly to their destinations and will require more transfers.  But we were surprised to observe this in the passenger perspective case as well.  

Compared to \cite{husselmann2023improved}'s DBMOSA, RC-EA achieves lower average trip time $C_p$ on Mumford2 and Mumford3 for the passenger perspective, despite the fact that it requires more passenger trips to make 1 and 2 transfers.  Still, RC-EA's networks require fewer transfers overall than NEA's, which matches our intuition that fewer transfers should result in lower average trip time $C_p$.  It seems that the process of compositing shortest paths to form routes biases those routes towards directness and efficiency, to a degree that outweighs the cost imposed by having more transfers.  Each transfer is less impactful on riders if the routes they're transferring between are more direct.  

Meanwhile, since both the learned heuristic of NEA and the unlearned heuristic of RC-EA suffer from high transfer counts, it seems the difference between these and the other results from the literature may be a result of the evolutionary algorithm itself.  At any rate, these results are not out of line with those of \cite{mumford2013new} and \cite{kilic2014demand} on the largest cities, so we do not consider this a major strike against our learned heuristics.  Indeed, we expect that these results could be improved by explicitly including $d_0$ in the reward function used to train the policy $\pi_\theta$, and this would reduce the number of transfers required by NEA's networks.

\section{Summary}

In this chapter, we proposed a low-level heuristic that uses the learned neural policies described in \autoref{chap:neural_policy} to modify an existing transit network, in a way that exploits the construction heuristics learned by the policy to prefer better neighbourhood moves to worse ones.  We showed that an evolutionary algorithm that used this heuristic as one of its mutators, which we call a Neural Evolutionary Algorithm, could achieve performance competitive with, and in some cases better than, state-of-the-art metaheuristic improvement methods for the \gls{ndp}, while running in approximately 80\% less wall-clock time.  And we performed several ablation studies to understand the contribution of different parts of the Neural Evolutionary Algorithm, finding that its performance versus a variant without the neural mutator was indeed due mainly to the learned policy.

Up to now, we have evaluated these methods on synthetic cities, whose only connection to real cities is in coarse statistics like the number of nodes and number of required routes.  In the next chapter, we discuss the use of this neural evolutionary algorithm in a more realistic scenario, with geography and travel demand closely based on real data from an existing city.

\chapter{Application to a Realistic Scenario}\label{chap:real_world}

Over the past three chapters, we developed a hybrid method composed of an evolutionary algorithm that uses a learned neural policy both to initialize the algorithm and to select some of the neighbourhood moves made during each mutation step.  We have evaluated our method's performance on a set of synthetic benchmark cities.  While some of these synthetic cities have statistics $n, S, m_\text{min}$, and $m_\text{max}$ based on those of a real-world city, all the particulars - the spatial layout of the nodes $\mathcal{N}$, the road network $\mathcal{E}_s$ that connects them, and the travel demands $D$ - are entirely synthetically generated, with no reference to the actual spatial and demand characteristics of their corresponding real cities.  In this chapter, we describe how we then applied our method to a more realistic instance of the \gls{ndp}, based closely on real data from the city of Laval in Quebec, Canada.  

Laval is a suburb of the major city of Montreal, located on an island north of the island of Montreal.  As of the 2021 census, it had a total population of 429,555~\citep{canadaCensus2021}.  Public transit in Laval is provided primarily by the bus network of the Soci\'et\'e de Transport de Laval; additionally, one line of the Montreal underground metro system has three stops in Laval.

\begin{figure}
    \centering
    \includegraphics[width=\columnwidth]{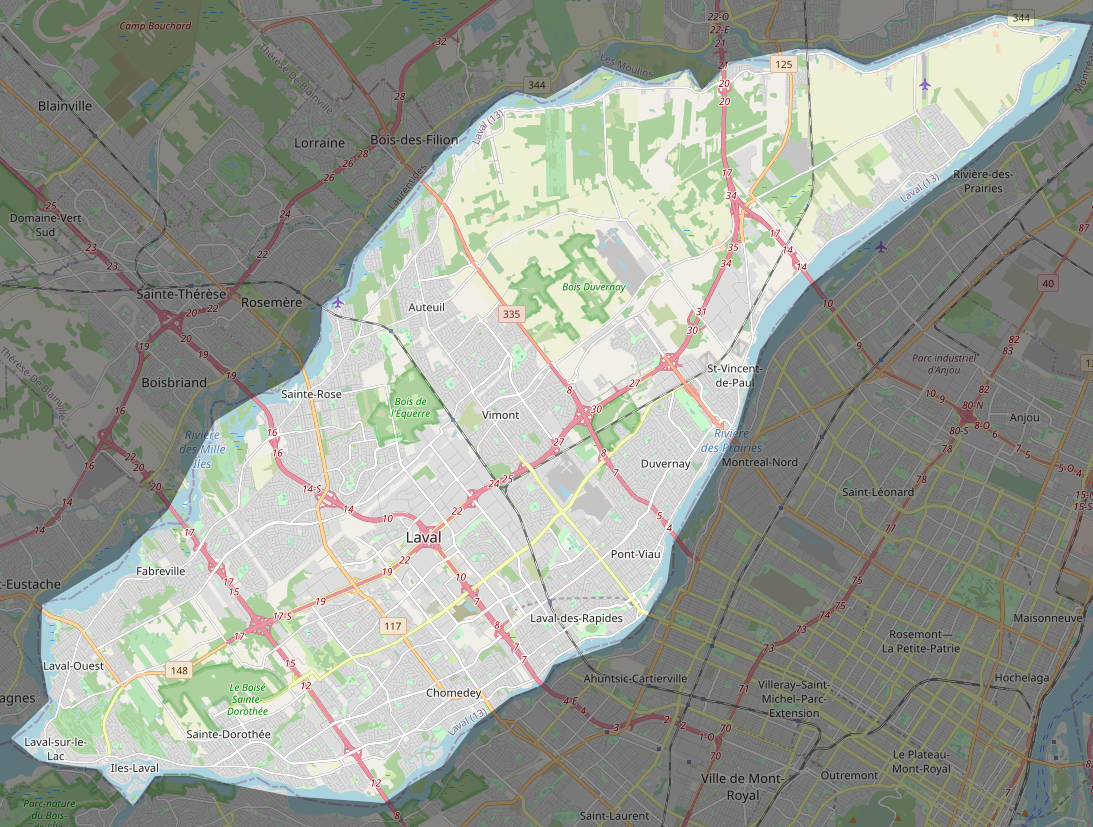}
    \caption{A map of the city of Laval, Quebec, taken from \cite{googleMapsLaval}.}
    \label{fig:laval_google_map}
\end{figure}

\section{Representing Laval}

To apply our learned policy $\pi_\theta$ to Laval, we first had to assemble a city graph $\mathcal{G} = (\mathcal{N}, \mathcal{E}_s, D)$ to represent it using the data sources that were available to us.  Our model of the Laval is based on several sources: geographic data on \glspl{cda} for 2021 from Statistics Canada~\citep{disseminationAreas}, a GIS representation of the road network of Laval provided to us by the Soci\'et\'e de Transport de Laval, an \gls{od} dataset~\citep{od_data} provided by Quebec's Agence M\'etropolitaine de Transport, and publicly-available GTFS data from 2013~\citep{laval_gtfs} that describes Laval's existing transit system.

\subsection{Street graph}\label{subsec:laval_streetgraph}

The real city of Laval contains 2,618 unique transit stop facilities.  Taking each of these to be a node yielded a very large graph.  The experiments in this chapter were run on a compute cluster node with an NVidia A100 GPU with 40 GB of VRAM; our neural policies needed more memory than this to run on the entire 2,618-node graph, creating a technical hurdle.  It is also likely that such a representation would be more fine-grained than necessary: the demand for travel within the service areas of nearby transit stops is likely to be very similar.  For these reasons, we instead made use of a coarsened graph, where the nodes correspond to \glspl{cda}.  \glspl{cda} are designed so as to enclose populations that are roughly homogeneous (among other factors), making them a sensible granularity at which to analyze travel demands.  Laval contains 632 distinct \glspl{cda}, and a graph size of $n=632$ makes it feasible for us to apply our neural policies with the hardware we have available.

The street graph $(\mathcal{N}, \mathcal{E}_s)$ was derived from the 2021 census data by taking the centroid of each \gls{cda} within Laval to be a node in $\mathcal{N}$, and adding a street edge $(i,j,\tau_{ij})$ for node pair $(i,j)$ if their corresponding \glspl{cda} shared a border.  To compute drive times $\tau_{ij}$, we found all points in the road network within the \glspl{cda} of $i$ and $j$, computed the shortest-path driving time over the road network from each of $i$'s road-network points to each of $j$'s, and set $\tau'_{ij}$ as the median of these driving times.  We did this so that $\tau'_{ij}$ would reflect the real drive times given the existing road network.  By this approach, $\tau'_{ij} \neq \tau'_{ji}$ in general, but as our method treats cities as undirected graphs, it expects that $\tau_{ij} = \tau_{ji} \; \forall \; i,j \in \mathcal{N}$.  To enforce this, as a final step we set:
\begin{align}
    \tau_{ij} = \tau_{ji} = \max(\tau'_{ij}, \tau'_{ji}) \; \forall \; i,j \in \mathcal{N}
\end{align}

\subsection{Existing transit}

The existing transit network in Laval has 43 bus routes that operate during morning rush hour from 7 to 9 AM.  To compare networks from our algorithm to the this network, we had to translate it to a network that runs over $(\mathcal{N}, \mathcal{E}_s)$ as defined in \autoref{subsec:laval_streetgraph}.  To do this, we mapped each stop on an existing bus route to the node in $\mathcal{N}$ of the \gls{cda} that contained the stop.  We will refer to this translated transit system as the \gls{stl} network, after the city's transit agency, the Soci\'et\'e de Transport de Laval.

The numbers of stops on the routes in the \gls{stl} network range from 2 to 52.  To ensure a fair comparison, we set $m_\text{min}=2$ and $m_\text{max}=52$ when running our algorithm.  Unlike the routes in our algorithm's networks, many of \gls{stl}'s routes are not symmetric (that is, they follow a different path in each direction between terminals), and several are unidirectional (they go only one way between terminals).  We maintained the constraints of symmetry and bidirectionality on our own algorithm since this was how our policies are trained, so relaxing the constraints might harm their performance.  But we did not enforce these constraints upon the \gls{stl} network, as doing so would require major changes to it, making it a much less realistic point of comparison.  Given that uni-directional streets are a common feature in real cities, it would be good to train future policies without these two constraints, but we leave this for future work. 

Since these constraints were not obeyed by \gls{stl}, when reporting total route time $C_o$ in this section, we calculated it for bidirectional routes by summing the travel times of both directions, instead of just one direction as in the preceding sections.  This allows a fair comparison of $C_o$ between \gls{stl} and other networks.

In addition, there is an underground metro line, the Montreal Orange Line, which has three stops in Laval.  As with the existing bus lines, we mapped these three stops to their containing \glspl{cda} and treat them as forming an additional route.  We added this metro route to both the \gls{stl} network and the networks $\mathcal{R}$ produced by our algorithm when evaluating them, to reflect that unlike the bus routes, it is not feasible to change the metro line and so it represents a constant of the city.

\subsection{Demand matrix}

\nomenclature{$\mathbf{l}_o$}{The (latitude,longitude) coordinates of the origin point of a trip in an \gls{od} dataset.}
\nomenclature{$\mathbf{l}_d$}{The (latitude,longitude) coordinates of the destination point of a trip in an \gls{od} dataset.}
\nomenclature{$expf$}{The expansion factor of a trip in an \gls{od} dataset, which estimates how many real trips correspond to that trip reported in the survey.}

Finally we assembled the demand matrix $D$ from the \gls{od} dataset.  The entries in the \gls{od} dataset correspond to trips reported by residents of Laval who were surveyed about their recent travel behaviour.  Each entry has latitude and longitude coordinates $\mathbf{l}_o$ and $\mathbf{l}_d$ for the trip's approximate origin and destination, as well as an ``expansion factor'' $expf$ giving the estimated number of actual trips corresponding to that surveyed trip.  It also indicates the mode of travel, such as car, bicycle, or public transit, that was used to make the trip.

Many entries in the \gls{od} dataset refer to trips that begin or end in Montreal  For our purposes, we ``redirected'' all trips made by public transit that enter or leave Laval to one of several ``crossover points'' that we defined.  These crossover points are the locations of the three Orange Line metro stations in Laval, and the locations of the last or first stop in Laval of each existing bus route that goes between Laval and Montreal  

For each trip between Laval and Montreal, we identified the crossover point that has the shortest distance to either of $mathbf{l}_o$ or $\mathbf{l}_d$, and overwrote the Montreal end-point of the trip by this crossover point.  This process was automated with a simple computer script.  The idea is that if the Laval end of the trip is close to a transit stop that provides access to Montreal, the rider will choose to cross over to the Montreal transit system as soon as possible, as most locations in Montreal will be easier to access once in the Montreal transit system.  Trips that go between Laval and Montreal by means other than public transit were not included in $D$, as we judge that most such trips cannot be induced to switch modes to transit.  

Having done this remapping, we initialized $D$ with all entries set to 0.  Then for each entry in the \gls{od} dataset, we found the \glspl{cda} that contain $mathbf{l}_o$ and $\mathbf{l}_d$, associated them with the matching nodes $i$ and $j$ in $\mathcal{N}$, and updated $D_{ij} \leftarrow D_{ij} + expf$.  $D$ then had the estimated demand between every pair of \glspl{cda}.  To enforce symmetric demand, we then assigned $D_{ij} \leftarrow \max(D_{ij}, D_{ji}) \; \forall \; i,j \in \mathcal{N}$.  The resulting demand matrix $D$ contains 548,159 trips, of which 63,104 are trips between Laval and Montreal that we redirected.

We then applied a final filtering step.  The \gls{stl} network does not provide a path between all node pairs $i,j$ for which $D_{ij} > 0$, and so it violates the connectedness constraint of \autoref{subsec:constraints}.  This is because $D$ at this point includes all trips that were made within Laval by any mode of transport, including to and from areas that are not served by the \gls{stl} network.  These areas may be unserved because they are populated by car owners unlikely to use transit if it were available.  To ensure a fair comparison between the \gls{stl} network and our algorithm's networks, we set to 0 all entries of $D$ for which the \gls{stl} network does not provide a path.  We expected this to cause our system to output transit networks that are ``closer'' to the existing transit network, in that they satisfy the same travel demand as before.  This will reduce the scope of the changes required to go from the existing network to a new one proposed by our system.

\autoref{tab:laval_stats} contains the statistics of, and parameters used for, the Laval scenario.

\begin{table*}[]
    \centering
    \resizebox{\textwidth}{!}{
    \caption{Statistics and parameters of the Laval scenario}\label{tab:laval_stats}
    \begin{tabular}{c c c c c c c}
    	\toprule
        \# nodes $n$ & \# street edges $|\mathcal{E}_s|$ & \# demand trips & \# routes $S$ & $m_\text{min}$ & $m_\text{max}$ & Area (km$^2$) \\
        \midrule
        632 & 4,544 & 548,159 & 43 & 2 & 52 & 520.1 \\
        \bottomrule
    \end{tabular}
	}
\end{table*}

\begin{figure*}[]
  \centering
  \begin{subfigure}[t]{0.45\columnwidth} 
    \includegraphics[width=\linewidth]{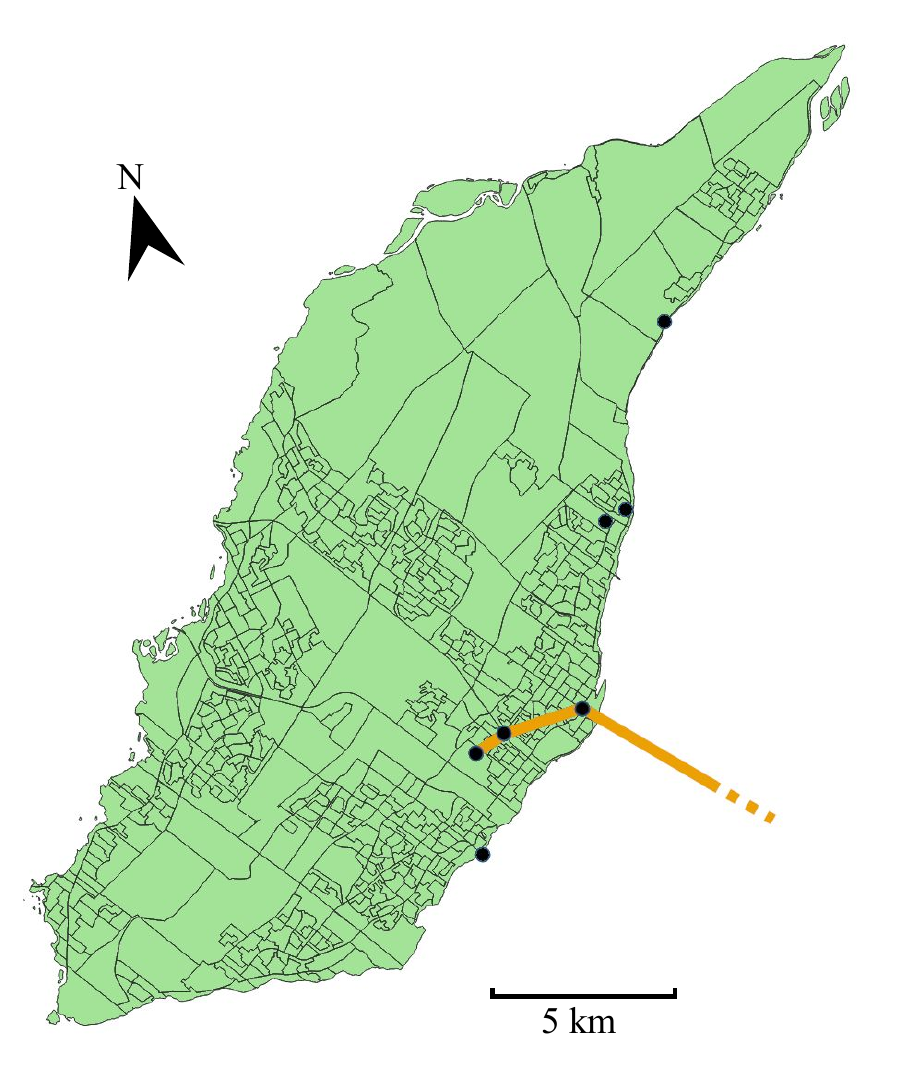}  
    \caption{Census dissemination areas}
    \label{subfig:cda_map}
  \end{subfigure}
  \hfill
  \begin{subfigure}[t]{0.45\columnwidth} 
    \includegraphics[width=\linewidth]{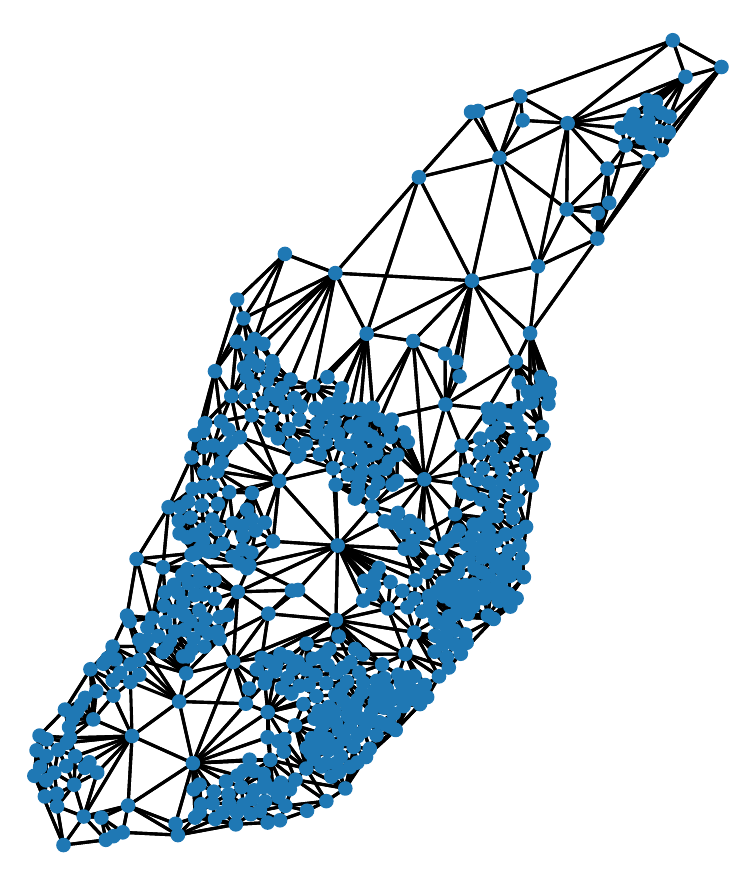}  
    \caption{Street graph}
    \label{subfig:laval_graph}
  \end{subfigure}

  \caption{\autoref{subfig:cda_map} shows a map of the \acrlong{cda}s of the city of Laval, with ``crossover points'' used to remap inter-city demands shown as black circles, and the Montreal Metro Orange Line shown in orange.  \autoref{subfig:laval_graph} shows the street graph constructed from these dissemination areas.}
  \label{fig:laval_maps}
\end{figure*}

\begin{figure}
	\includegraphics[width=0.8\linewidth]{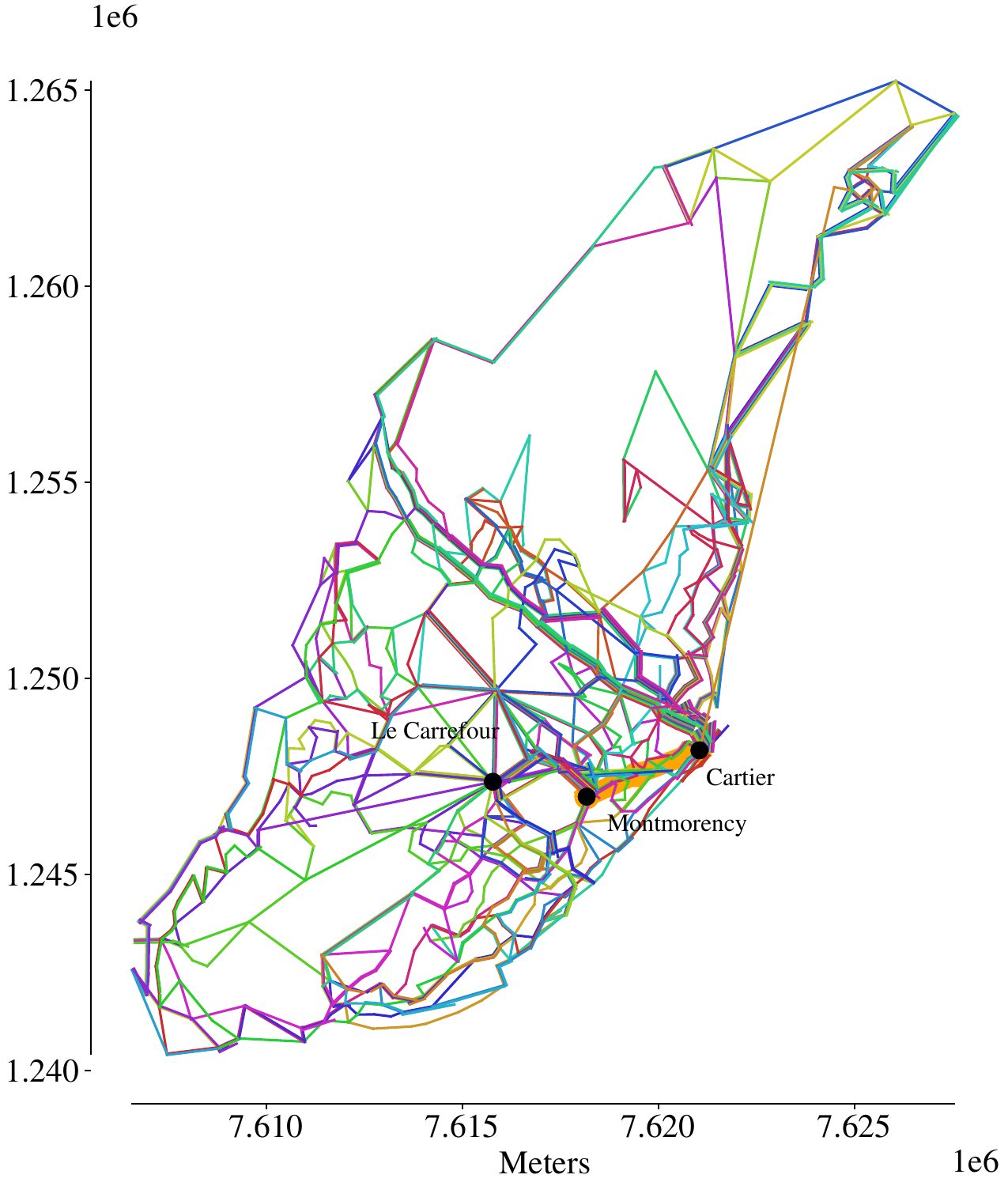}
	\caption{The routes of our \gls{stl} network.  Routes that go along a single edge are placed next to each other, so thicker ``rainbow'' edges show multiple parallel routes.  The thick orange line shows the Orange Line subway of the Montreal Metro, with its stations as orange circles, and the black circles show the nodes of the \glspl{cda} containing Laval's three major bus terminals: Le Carrefour, Montmorency, and Cartier.}
	\label{fig:stl_routes}
\end{figure}

The final set of routes in our \gls{stl} network is shown in \autoref{fig:stl_routes}.  We observe that the three major bus terminals, shown in black, are highly connected, with many discrete routes visiting them, although a large fraction of the bus routes connected to Le Carrefour and Montmorency connect them quickly to each other.  The Montmorency bus terminal is located at the Montmorency metro station on the orange line, so it is likely that these routes are meant to provide access to and from the orange line.  The same is true of the Cartier terminal, which is located at the Cartier orange line station: many routes connect to this node from diverse locations around the Laval graph, probably because of its co-location with the metro station, which is also the closest metro station to Montreal itself.  Connecting Laval's residents to Montreal via the orange line is clearly a priority for the \gls{stl}'s planners.

\section{Enforcing constraint satisfaction}

In the experiments on the Mandl and Mumford benchmark cities described in \autoref{chap:neighbourhood_moves}, the EA and NEA algorithms never produced networks that violated any of the constraints on networks outlined in \autoref{subsec:constraints}.  However, Laval's city graph, with 632 nodes, is considerably larger than even the largest Mumford city, which has only 127 nodes.  In our initial experiments on Laval, we found that both EA and NEA consistently produced networks that violated the connectedness constraint, providing no transit path for some node pairs $(i,j)$ for which $D_{ij} > 0$.  

To remedy this, we modified the \gls{mdp} described in \autoref{sec:mdp} to enforce reduction of unsatisfied demand.  Let $c_1(\mathcal{G}, \mathcal{R})$ be the number of node pairs $i,j$ with $D_{ij} > 0$ for which network $\mathcal{R}$ provides no transit path, and let $\mathcal{R}'_t = \mathcal{R}_t \cup \{ r_t \}$, the network formed from the finished routes and the in-progress route $r_t$.  We applied the following changes to the action space $\mathcal{A}_t$ whenever $c_1(\mathcal{R}_t) > 0$:  
\begin{itemize}
    \item When the timestep $t$ is even and $\mathcal{A}_t$ would otherwise be $\{\textup{continue}, \textup{halt}\}$, the $\textup{halt}$ action is removed from $\mathcal{A}_t$.  This means that if $c_1(\mathcal{G}, \mathcal{R}'_t) > 0$, the current route must be extended if it is possible to do so without violating another constraint.
    \item When $t$ is odd, if $\mathcal{A}_t$ contains any paths that would reduce $c_1(\mathcal{G}, \mathcal{R}'_t)$ if added to $r_t$, then all paths that would {\it not} reduce $c_1(\mathcal{G}, \mathcal{R}'_t)$ are removed from $\mathcal{A}_t$.  This means that if it is {\it possible} to connect some unconnected trips by extending $r_t$, then {\it not} doing so is forbidden.
\end{itemize}


With these changes to the \gls{mdp}, we found that both LC-100 and NEA produced networks that connected all desired passenger trips, with no constraint violations - even when the policy $\pi_\theta$ was trained without these changes.  The results reported in this section are obtained using this variant of the \gls{mdp}.

\section{Results}\label{sec:laval_results}

We performed experiments evaluating both the NEA and RC-EA algorithms on the Laval city graph.  These experiments were conducted on different hardware than those in the previous chapters: specifically, they were run on an academic compute cluster node, using one NVidia A100 GPU with 40 GB of GPU RAM.

Three sets of NEA experiments were performed with different $\alpha$ values representing different perspectives: the operator perspective ($\alpha=0.0$), the passenger perspective ($\alpha=1.0$), and a balanced perspective ($\alpha=0.5$).  RC-EA experiments are only run at $\alpha=1.0$ since we observed in \autoref{subsec:rcea} that RC-EA did not perform well at other $\alpha$.  As before, we perform ten runs of each experiment with ten different random seeds and ten different learned policies $\pi_\theta$, and report statistics over these ten runs; the same per-seed parameters $\theta$ are used as in the experiments of \autoref{chap:neighbourhood_moves}.  We also run the \gls{stl} network on the Laval city graph $\mathcal{G}$ in the same way that we run the networks from our other algorithms to measure their performance, but since the \gls{stl} network is a single, pre-determined network that does not depend on $\alpha$, we run it only once.

The parameters of NEA were the same as in the Mumford experiments in \autoref{sec:baseline}: $I=400$, $B=10$, $F=10$, and transfer penalty $p_T = 300$s.

\begin{figure*}
    \centering
    \includegraphics[width=0.5\columnwidth]{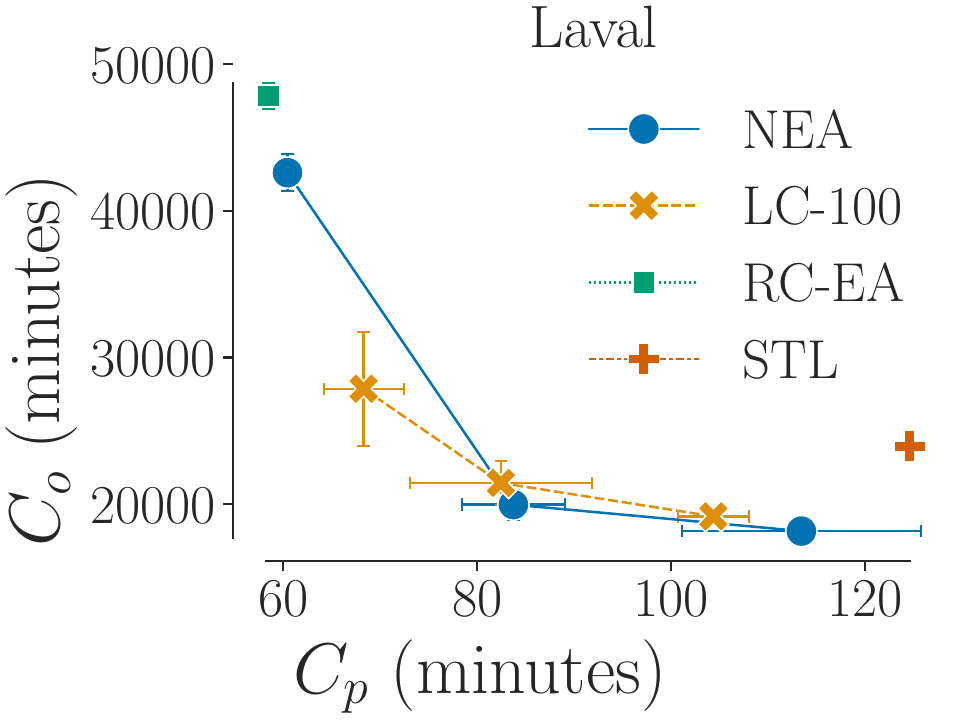}
    \caption{Average trip time $C_p$ versus total route time $C_o$ achieved by LC-100 and NEA at $\alpha=0.0$ (bottom-right), 0.5 (middle), and 1.0 (top-left), as well as for RC-EA at $\alpha=1.0$ and the \gls{stl} network itself (independent of $\alpha$).  Except for \gls{stl}, each point is a mean value over 10 random seeds, and bars show one standard deviation.}
    \label{fig:laval_pareto}
\end{figure*}

The results of the Laval experiments are shown in \autoref{tab:laval_results}.  The results for the initial networks produced by LC-100 are presented as well, to show how much improvement the NEA makes over its initial networks.  \autoref{fig:laval_pareto} highlights the trade-offs each method achieves between average trip time $C_p$ and total route time $C_o$, and how they compare to the \gls{stl} network.  

\begin{table*}[]
    \centering
    \resizebox{\textwidth}{!}{
    \begin{tabular}{clcccccc}
	\toprule
	$\alpha$ & Method & $C_p \downarrow$ & $C_o \downarrow$ & $d_0 \uparrow$ & $d_1$ & $d_2$ & $d_{un} \downarrow$ \\
	\midrule
	N/A  & \gls{stl}   &   124.61 &   23954 &        14.48 &       22.8 &      20.88 &         41.83 \\
	\midrule
	& LC-100 & \bf 104.36 $\pm$ 3.69 & 19175 $\pm$ 316 & 14.25 $\pm$ 0.64 & 21.38 $\pm$ 1.15 & 21.81 $\pm$ 1.23 & 42.57 $\pm$ 2.49 \\
	0.0 & NEA & 113.44 $\pm$ 12.34 & \bf 18171 $\pm$ 486 & \bf 14.70 $\pm$ 1.14 & 20.62 $\pm$ 2.10 & 22.17 $\pm$ 1.72 & 42.52 $\pm$ 3.38 \\
	\midrule
	& LC-100 & \bf 82.44 $\pm$ 9.38 & 21452 $\pm$ 1482 & 13.56 $\pm$ 0.61 & 24.23 $\pm$ 2.73 & 25.01 $\pm$ 3.13 & 37.20 $\pm$ 5.63 \\
	0.5 & NEA & 83.74 $\pm$ 5.32 & \bf 19978 $\pm$ 1015 & 13.68 $\pm$ 0.80 & 24.01 $\pm$ 1.88 & 25.23 $\pm$ 1.85 & \bf 37.08 $\pm$ 3.48 \\
	\midrule
	& LC-100 & 68.29 $\pm$ 4.13 & \bf 27871 $\pm$ 3871 & 15.89 $\pm$ 1.41 & 29.70 $\pm$ 3.74 & 27.60 $\pm$ 1.16 & 26.81 $\pm$ 5.77 \\
	1.0 & NEA & 60.43 $\pm$ 1.03 & 42600 $\pm$ 1269 & 20.29 $\pm$ 0.82 & 33.82 $\pm$ 2.24 & 27.51 $\pm$ 0.93 & 18.39 $\pm$ 3.16 \\
	& RC-EA  & \bf 58.46 $\pm$ 0.24 &  47810 $\pm$ 900 & \bf 21.26 $\pm$ 0.46 & 38.98 $\pm$ 0.86 & 26.87 $\pm$ 0.50 & \bf 12.88 $\pm$ 0.74 \\
	\bottomrule
\end{tabular}
}
    \caption{Performance of networks from LC-100 and NEA at three $\alpha$ values, RC-EA at $\alpha=1.0$, and the \gls{stl} network.  Bolded values are best for the corresponding $\alpha$, except where \gls{stl} performs better than all methods at that $\alpha$, in which case no values are bolded.  $C_p$ is the average passenger trip time. $C_o$ is the total route time.  $d_i$ is the percentage of trips satisfied with number of transfers $i$, while $d_{un}$ is the percentage of trips satisfied with 3 or more transfers.
	All values are averaged over ten random seeds, with one standard deviation following ``$\pm$'', except for those in the \gls{stl} row.}
    \label{tab:laval_results}
\end{table*}

We see that for each value of $\alpha$, NEA's network outperforms the \gls{stl} network: it achieves 52\% lower $C_p$ at $\alpha=1.0$, 24\% lower $C_o$ at $\alpha=0.0$, and at $\alpha=0.5$, 33\% lower $C_p$ and 17\% lower $C_o$.  At both $\alpha=0.0$ and $\alpha=0.5$, NEA strictly dominates the existing transit system, achieving both lower $C_p$ and lower $C_o$.

Comparing NEA's results to LC-100, we see that at $\alpha=0.0$, NEA decreases $C_o$ and increases $C_p$ relative to LC-100; and for $\alpha=1.0$, the reverse is true.  In both cases, NEA improves over the initial network on the objective being optimized, at the cost of the other, as we would expect.  In the balanced case ($\alpha=0.5$), we see that NEA decreases $C_o$ by 7\% while increasing $C_p$ by 1.6\% relative to LC-100.  This matches what we observed in \autoref{fig:40k}, as discussed in \autoref{sec:baseline}.  NEA improves on the \gls{stl} network by a considerable margin at all three $\alpha$ values, even though RC-EA outperforms NEA at $\alpha=1.0$.

Looking at the transfer percentages $d_0$, $d_1$, $d_2$, and $d_{un}$, we see that as $\alpha$ increases, the overall number of transfers shrinks, with $d_0$, $d_1$ and $d_2$ growing and $d_{un}$ shrinking.  This is as expected, since fewer transfers implies reduced passenger travel times.  By the same token, at $\alpha=1.0$ RC-EA performs best on these metrics, as it does on average trip time $C_p$.  At $\alpha=1.0$, NEA outperforms the \gls{stl} network by these metrics as well.  At $\alpha=0.5$, surprisingly, NEA's $d_0$ is lower than at the other $\alpha$ values, and lower than the \gls{stl} network's $d_0$ by 0.8\% of trips.  But NEA's $d_{un}$ is lower than \gls{stl}'s by 4.8\% of trips, while $d_1$ and $d_2$ are both higher, meaning that NEA's networks let more passengers reach their destination in 2 or fewer transfers than the \gls{stl} network.  It is only at $\alpha=0.0$ that \gls{stl}'s network requires fewer transfers overall than NEA's networks, and this is reasonable given that at $\alpha=0.0$ the goal is strictly to minimize route length regardless of transfers.

\subsection{Visual Inspection}

\begin{figure*}
	\includegraphics[width=0.8\linewidth]{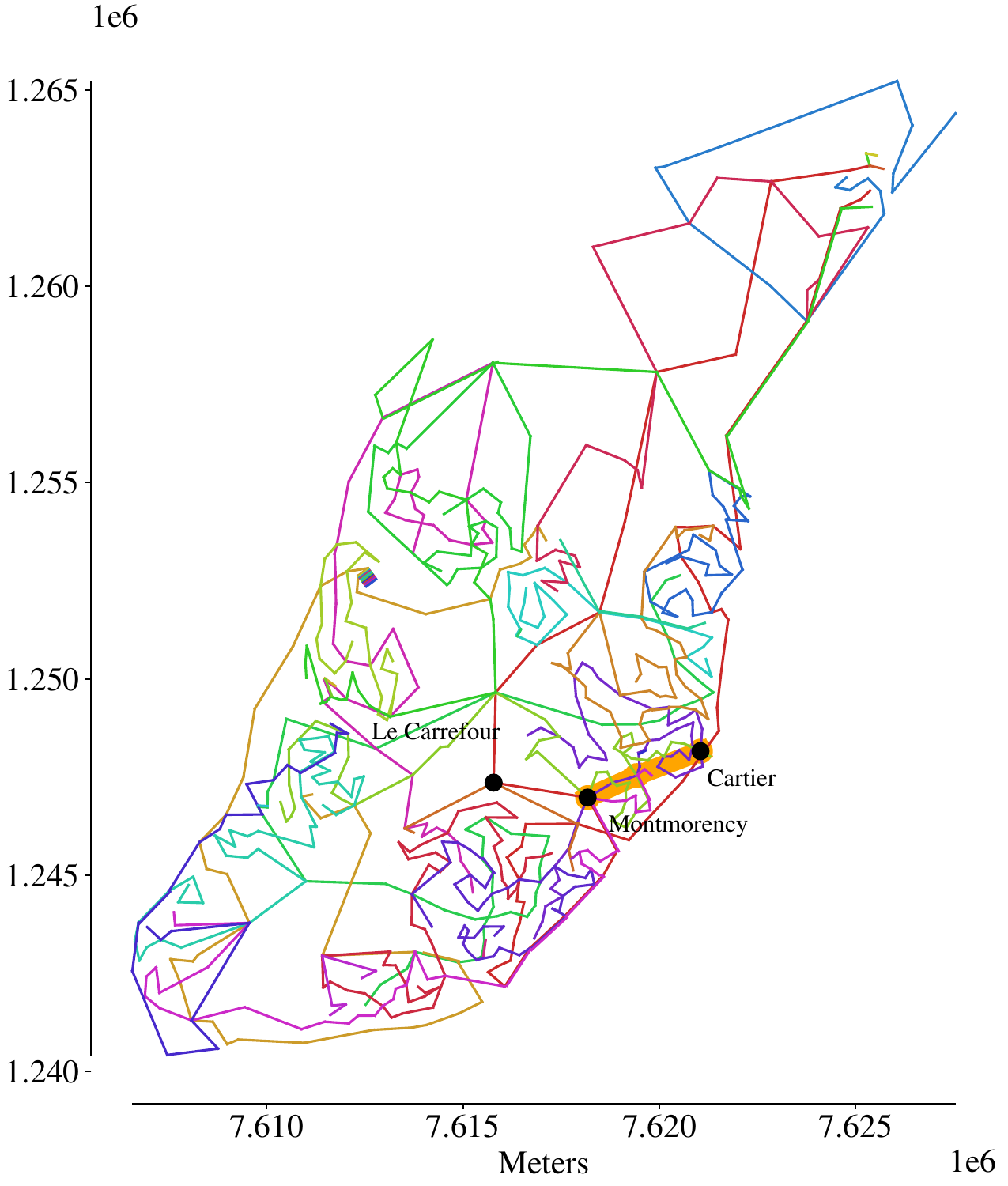}
	\caption{The routes of one of NEA's generated networks at $\alpha = 0.0$.  The heavy orange line and its circles represent the orange line metro and its stops, while the black dots represent existing bus terminals.}
	\label{fig:a0_routes}
\end{figure*}

\begin{figure*}
	\includegraphics[width=0.8\linewidth]{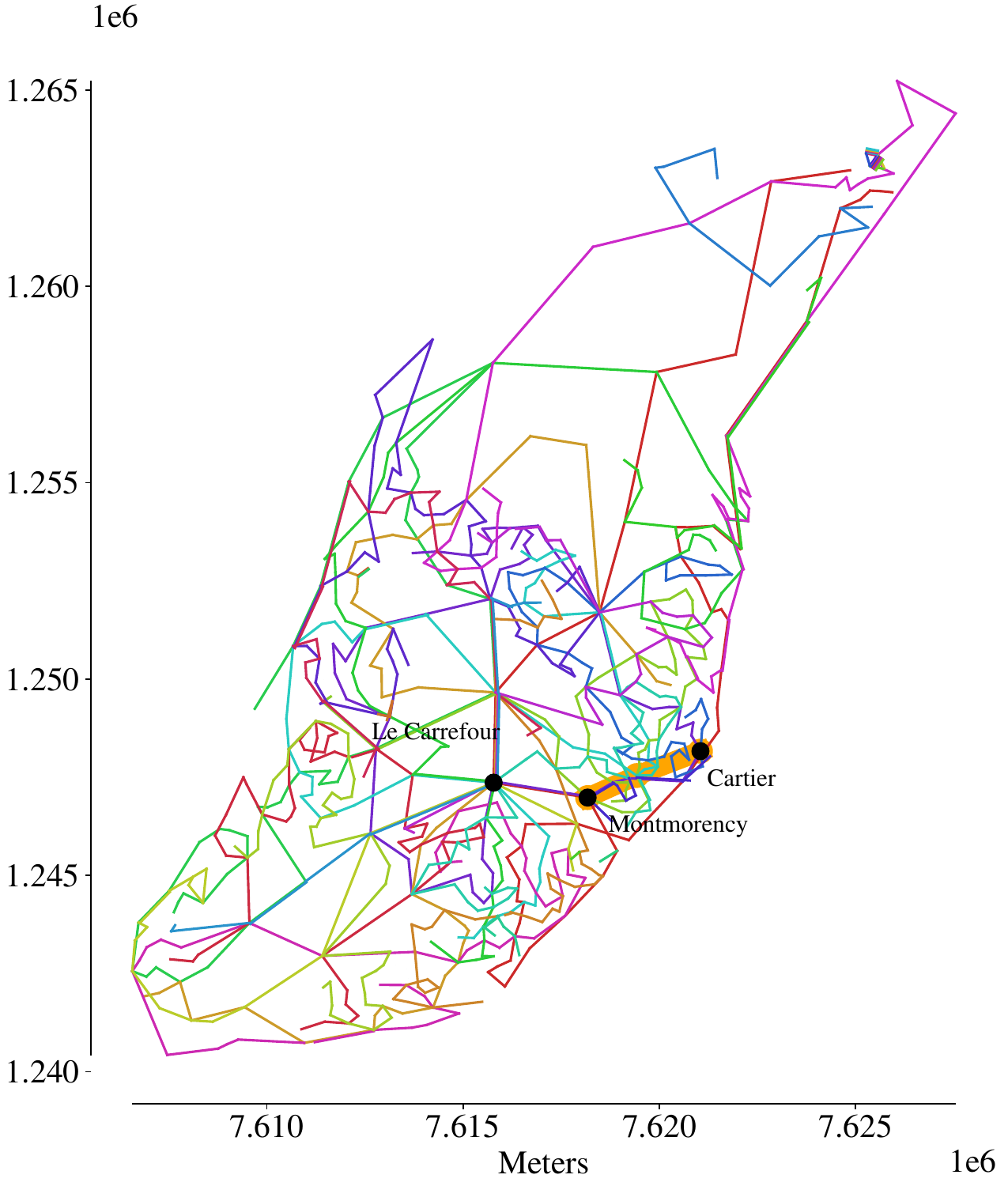}
	\caption{The routes of one of NEA's generated networks at $\alpha = 0.5$.  The heavy orange line and its circles represent the orange line metro and its stops, while the black dots represent existing bus terminals.}
	\label{fig:a0p5_routes}
\end{figure*}

\begin{figure*}
	\includegraphics[width=0.8\linewidth]{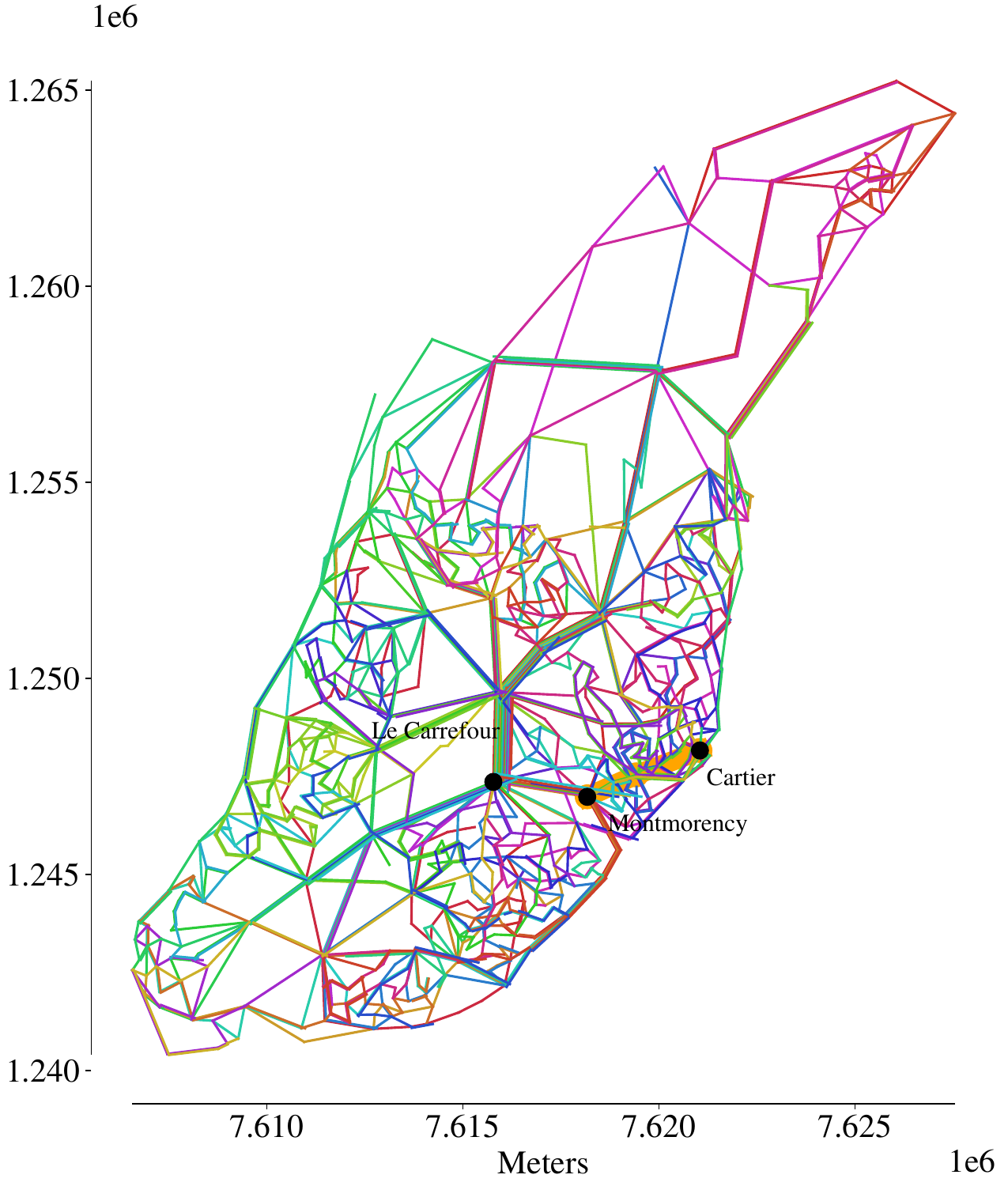}
	\caption{The routes of one of NEA's generated networks at $\alpha = 1.0$.  The heavy orange line and its circles represent the orange line metro and its stops, while the black dots represent existing bus terminals.}
	\label{fig:a1_routes}
\end{figure*}

\autoref{fig:a0_routes}, \autoref{fig:a0p5_routes}, and \autoref{fig:a1_routes} show one of the networks produced by NEA for each of $\alpha=0.0$, $\alpha=0.5$, and $\alpha=1.0$.  These networks were chosen among the ten networks for each $\alpha$ value based on their $(C_p, C_o)$ being closest to the mean $(C_p, C_o)$ of the ten networks, so that the network displayed would be representative of the algorithm's performance.

\begin{table*}
	\centering
    \resizebox{\textwidth}{!}{	
    \begin{tabular}{cccccc}
    	\toprule
    	Terminal & Has metro station & \# routes in: \gls{stl} & NEA $\alpha=0.0$ & NEA $\alpha=0.5$ & NEA $\alpha=1.0$ \\
    	\midrule
    	Le Carrefour & No            & 25			    & 5.0              & 6.8              & 21.0 \\
    	Montmorency	 & Yes			 & 29			    & 2.9              & 3.9              & 9.2 \\
    	Cartier	     & Yes 			 & 38			    & 1.2              & 1.1              & 4.1 \\
    	\midrule
    	Total		 &   			 & 92				& 9.1             & 11.7             & 34.3 \\
    	\bottomrule
    \end{tabular}
	}
	\caption{Counts of the number of routes that visit each bus terminal in each network.  For NEA, counts are the average over the ten networks computed at that $\alpha$ value.}
	\label{tab:n_routes_visiting}
\end{table*}

One thing we can observe in these figures is that the NEA $\alpha=0.0$ and NEA $\alpha=0.5$ networks have many fewer routes that connect directly to the three large bus terminals.  To confirm this, we found the number of routes that stop at each terminal node for \gls{stl} and NEA each $\alpha$ value, averaged over the ten random seeds.  We present these in \autoref{tab:n_routes_visiting}.  As the maps suggest, at $\alpha=0.0$ and $\alpha=0.5$, the number of routes that reach the terminals is very small compared to the \gls{stl} network.  At $\alpha=1.0$, there are many more than at $\alpha=0.0$ and $\alpha=0.5$, but there are still less than half as many stops at bus terminals as in the \gls{stl} network.  This may suggest that a ``hub and spoke'' network where a few large terminals are served by many routes is less efficient than a more ``distributed'' network where routes provide direct trips between more diverse pairs of nodes.  The fact that the number of transfers in NEA's $\alpha=0.0$ and $\alpha=0.5$ networks is lower than in the \gls{stl} network indicates that a distributed network need not require more transfers overall.  The fact that the number of connections to these terminals increases with $\alpha$ is what we would expect: at low $\alpha$, we want routes to be as short as possible.  Connecting more nodes via transfers becomes more attractive as this allows routes to be shorter.

It is also interesting to note that the ordering of terminals by the number of routes that visit is reversed between the \gls{stl} network and NEA's networks: in \gls{stl}, Cartier is the most-visited and Le Carrefour the least, while the reverse is true in the NEA networks on average.  This means that NEA de-emphasizes direct connections to the orange line metro.  We do note, however, that in each of the NEA maps, there is at least one direct line connecting Le Carrefour to Montmorency (the first orange line stop), with no bus stops in between - something that the \gls{stl} network lacks.  It appears that NEA uses Le Carrefour as a major point of connection for Laval, perhaps due to its relative centrality compared with Montmorency and Cartier, and then uses a route from Le Carrefour to connect people to the orange line.  This may partly explain why NEA's networks have higher numbers of one-transfer trips ($d_1$) and lower numbers of direct trips ($d_0$) compared with the \gls{stl} network.

Inspecting \autoref{fig:a0_routes}, we note that at $\alpha=0.0$, there are twelve very short, two-node routes that appear along the same short edge.  This is visible as a small multi-coloured block at about $7.6125e6$ on the x-axis and $1.2525e6$ on the y-axis.  An inspection of the other $\alpha=0.0$ networks from NEA reveals similar blocks in most of them, though not always in the same place.  \autoref{fig:a0p5_routes} shows a similar cluster near the upper-right tip of Laval.  

This is most likely the response of NEA to the need to plan a fixed number of routes ($S=43$ for Laval), when it has already found some number $s < S$ of routes that adequately connect the city.  The best way to minimize $C_o$ is then to make the remaining $S-s$ routes as short as possible.  Since we have set the minimum number of stops $m_\text{min}=2$, this is best done by placing all remaining routes along the shortest edge in the graph.  Supporting this is as the likely cause is that in \autoref{fig:a1_routes}, where $\alpha = 1.0$, we notice no obvious two-node routes, nor any ``blocks'' of these along a single edge - in this case, the lengths of routes doesn't matter, so it is never beneficial to make these very short routes.

\begin{figure*}
	\begin{subfigure}[b]{0.43\textwidth}
		\centering
		\includegraphics[width=\textwidth]{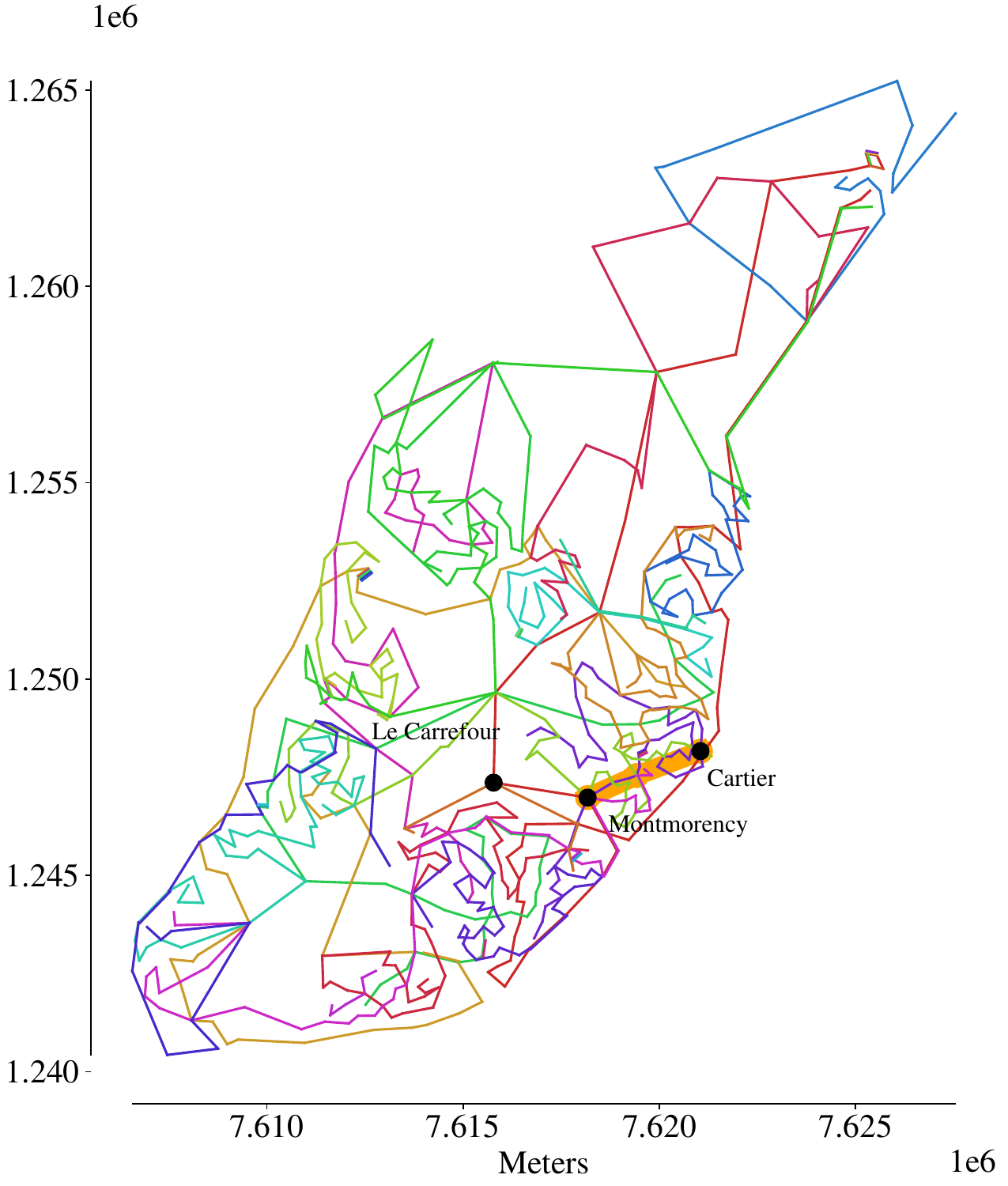}
		\caption{$\alpha=0.0$}
	\end{subfigure}
	\begin{subfigure}[b]{0.43\textwidth}
		\centering
		\includegraphics[width=\textwidth]{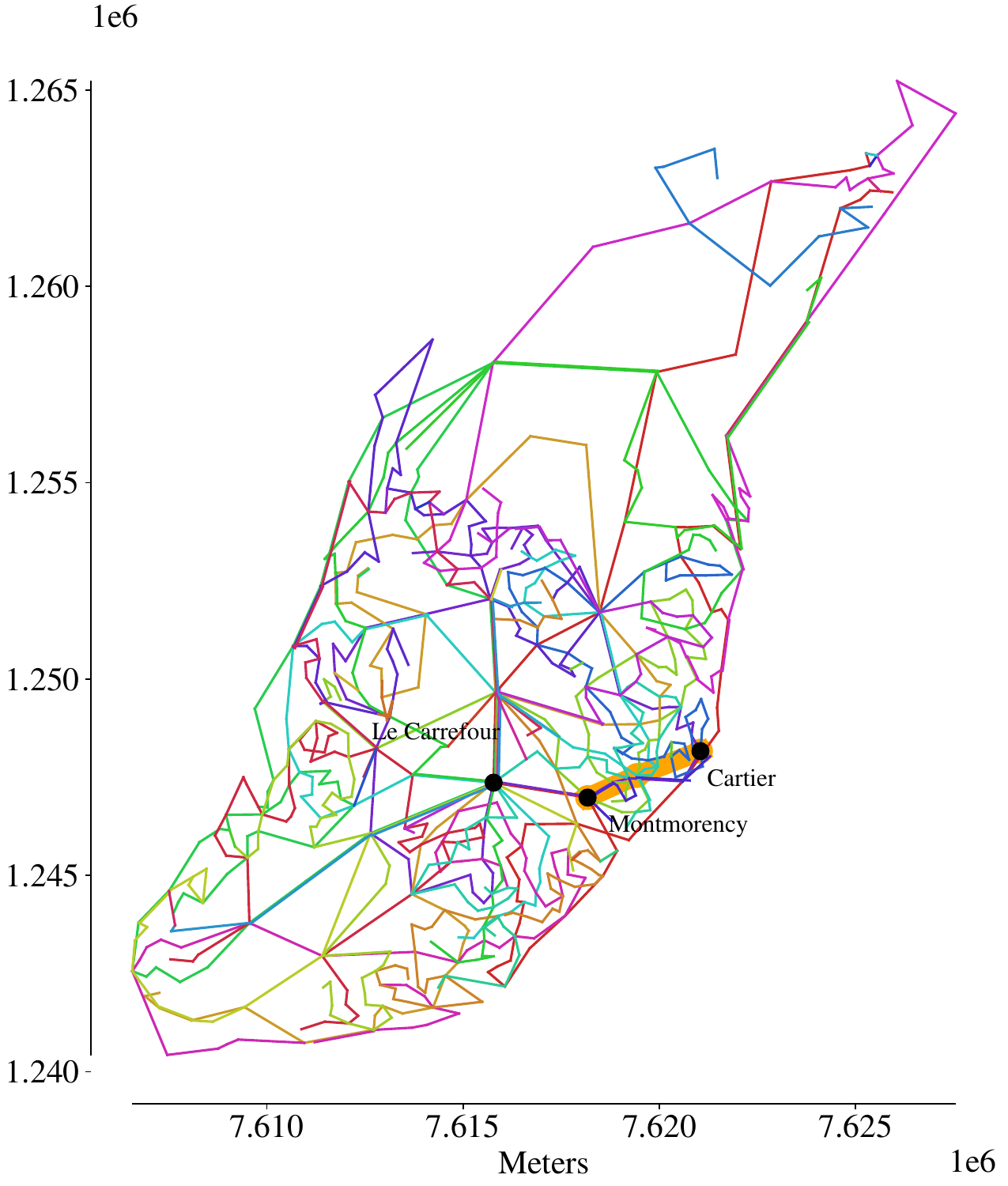}
		\caption{$\alpha=0.5$}	
	\end{subfigure} \\
	\begin{subfigure}[b]{0.43\textwidth}
		\centering
		\includegraphics[width=\textwidth]{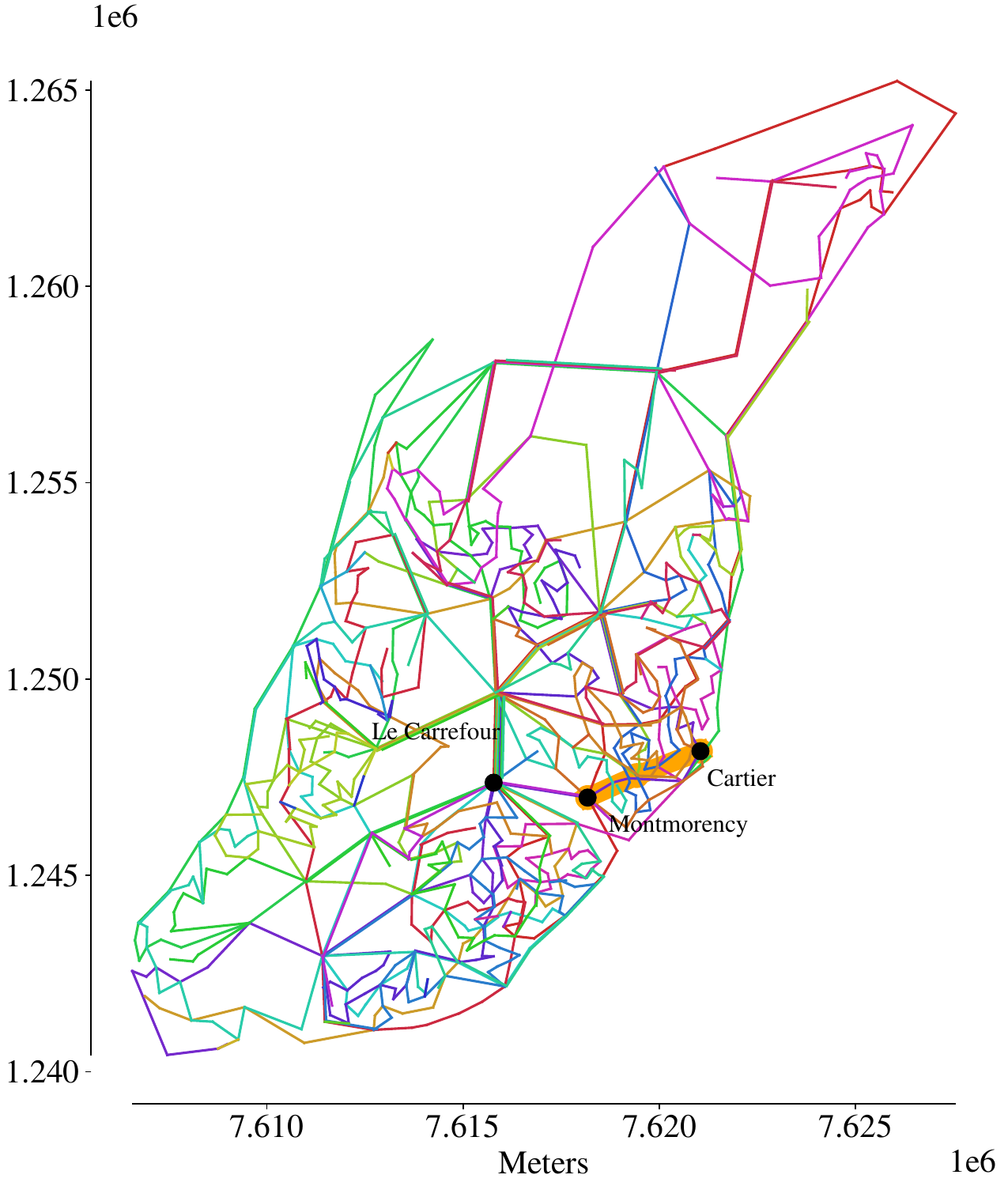}
		\caption{$\alpha=1.0$}	
	\end{subfigure}
	\caption{LC-100 initial networks for the three NEA runs shown in Figures \ref{fig:a0_routes}, \ref{fig:a0p5_routes}, and \ref{fig:a1_routes}.}
	\label{fig:laval_init_maps}
\end{figure*}

We inspected the initial network, produced by LC-100, that we used for the NEA run that produced the network in \autoref{fig:a0_routes} (shown in \autoref{fig:laval_init_maps}.  We found that the same edge had six two-node routes along it in the LC-100 network.  The other six were placed there by NEA.  We also performed a run of plain EA with the same initial network, $\alpha$, and parameters that we used for NEA on Laval elsewhere; this run ended with ten two-node routes along this edge.  So the neural mutator was not the main cause of these routes being added during the improvement method's run.  We thus conclude that this strategy of filling up the ``quota'' $S$ of required routes with parallel very short routes was both learned by the neural policy $\pi_\theta$ and stumbled upon by the un-learned evolutionary algorithm, and both components contributed equally to this outcome.  

In a real situation where we were only concerned with $C_o$, the appropriate thing to do when applying this plan to the real world would simply be to discard such routes, and treat this as a suggestion from the algorithm that $S$ is set too high and ought to be $S-s$ instead.

The eastern end of the island of Laval (the upper-right quadrant of each map) is sparsely-populated compared to the rest of the city, as \autoref{fig:laval_maps} reveals by the relatively small number and large size of the census dissemination areas there.  Directing our attention to this region of \autoref{fig:stl_routes}, we see that the \gls{stl} network has several routes that together wrap around the outside of this part of Laval, as well as others that pass through the more densely-populated areas near Laval's eastern tip, and which are especially dense in a neighbourhood on the south side of that tip.  Meanwhile, NEA's routes are more uniformly distributed throughout this area.  At no setting of $\alpha$ do these routes wrap fully around the outside of the island's eastern tip; instead, NEA sends its routes through the middle of this region, connecting nodes at the periphery of the graph directly to nodes in the interior.  The routes in this region grow more dense as $\alpha$ increases, as we would expect, but remain more evenly distributed throughout this area than the tightly-clustered routes of the \gls{stl} network.  Apparently, this more uniform coverage is more efficient than the \gls{stl} network's.

This seems to characterize the difference in other parts of Laval as well; the \gls{stl}'s routes are straight and run in parallel more often than NEA's, and NEA's routes are more meandering and tightly-wound upon themselves.  At $\alpha=1.0$, these paths together overlap cross often enough to create the appearance of a triangular mesh over large parts of the city.  These meandering routes seem to stem from the learned policy: \autoref{fig:laval_init_maps} shows the initial networks used for NEA to get the networks shown in \autoref{fig:a0_routes}, \autoref{fig:a0p5_routes}, and \autoref{fig:a1_routes}, and these networks exhibit the same qualitative ``wiggliness'' as the routes produced by NEA.

\section{Practical takeaways}

The likely aim of Laval's transit agency is to reduce costs while improving service quality (or at least not harming it too much).  The ``balanced'' case, where $\alpha = 0.5$, is more relevant to that aim than the $\alpha=0.0$ or $\alpha=1.0$ cases.  In this balanced case, we find that NEA proposes networks that reduce total route time by 17\% versus the existing \gls{stl} transit network.  Some simple calculations can help understand the savings this represents.  

The approximate cost of operating a Laval city bus is 200 Canadian dollars (CAD) per hour.  Suppose that each route has the same headway (time between bus arrivals) $H$.  Then, the number of buses $N_r$ required by route $r$ is the ceiling of the route's driving time $\tau_r$ divided by the headway, $\left \lceil \frac{\tau_r}{H} \right \rceil$.  The total number of buses $N_\mathcal{R}$ required for a transit network $\mathcal{R}$ is then:
\begin{equation*}
N_\mathcal{R} = \sum_{r \in \mathcal{R}} N_r \approx \frac{1}{H} \sum_{r \in \mathcal{R}} \tau_r = \frac{C_o}{H}
\end{equation*}

The total operating cost of the system is $N_\mathcal{R} * 200$ CAD.  Assuming a headway of 15 minutes on all routes, that gives a per-hour operating cost of 319,387 CAD for the \gls{stl} network, while the average operating cost of networks from NEA with $\alpha = 0.5$ is 266,373 CAD.  So, under these simplifying assumptions, NEA could save the Laval transit agency on the order of 53,000 CAD per hour, or about 17\% of operating costs.  This is only a rough estimate: in practice, headways differ between routes, and there may be additional costs involved in altering the routes from the current system, such as from moving or building bus stops.  But it shows that NEA, and neural heuristics more broadly, may offer practical savings to transit agencies.

In addition, NEA's balanced-case networks reduce the average passenger trip time by 33\% versus the \gls{stl} network, and reduce the number of transfers that passengers have to make, especially reducing the percentage of trips that require three or more transfers from 41.83\% to 37.08\%.  Trips requiring three or more transfers are widely regarded as being so unattractive that riders will not make them (but note that these trips are still included in the calculation of $C_p$).  So NEA's networks may even increase overall transit ridership.  This not only makes the system more useful to the city's residents, but also increases the agency's revenue from fares.

We also note that at $\alpha=1.0$, RC-EA achieves only 3\% lower $C_p$ than NEA.  This supports our conclusions in \autoref{subsec:rcea} and \autoref{sec:vs_prior} that the learned heuristics of the neural net policies $\pi_\theta$ are the main driver of NEA's performance across $\alpha$, rather than simply the heuristic of assembling routes from shortest paths.

In these experiments, the nodes of the graph represent \acrlong{cda} instead of existing bus stops.  In order to be used in reality, the proposed routes would need to be fitted to the existing locations of bus stops in the \glspl{cda}, likely making multiple stops within each \glspl{cda}.  To minimize the cost of the network redesign, it would be necessary to keep stops at most or all stop locations that have shelters, as constructing or moving these may cost tens of thousands of dollars, while stops with only a signpost at their locations can be moved at the cost of only a few thousand dollars.  An algorithm for translating our census-level routes to real-stop-level routes, in a way that minimizes cost, is an important next step if the methods developed here are to be applied in real cities.  We leave this for future work.

\section{Summary}

We have shown that our learned graph net policies are useful for transit planning in large, realistic cities.  The learned policies can construct transit networks that outperform a model of the city's existing transit, in terms of both estimated operating costs and passenger travel times.  In combination with an evolutionary algorithm as described in \autoref{chap:neighbourhood_moves}, the resulting transit network can be further improved - in the case of the city of Laval, the neural-evolutionary algorithm reduces operating costs by 17\% and passenger travel times by 33\% versus the existing transit network.

Our learned policies are trained on small synthetic cities of only twenty nodes, but our results show that they generalize well even to Laval's graph of more than six hundred nodes.  This shows that there exist good heuristics for the \gls{ndp} which are simple enough to be evident in very small graphs but which generalize to much larger graphs.  It is likely that there are more heuristics which could be learned only from experience on larger graphs, and that policies trained on a range of graph sizes could outperform those considered here.  

These results also demonstrate that our method is capable of incorporating fixed, pre-existing routes in its planning, in this case the Orange Line subway in Laval.  This is an important capability for solving real instances of the \gls{ndp}, as rail and underground transit routes are common in real cities and must be integrated with more flexible forms of transit like buses.

A number of challenges must still be overcome to make these methods for the \gls{ndp} directly applicable to the design of transit networks for real cities, such as vehicle scheduling and accounting for time-varying demand.  But finding good solutions to the \gls{ndp} is the first major challenge, and the results of this chapter show that our learning-based methods can capably address it.

\chapter{Conclusion}\label{chap:conclusion}



In this thesis, we explored various ways of using \acrfull{rl} with \acrlong{gat}s to solve the \acrlong{ndp}.  In \autoref{chap:neural_policy}, we develop a neural policy architecture and an \acrfull{rl} algorithm to train these policies to construct transit networks.  In \autoref{chap:initialization} we showed that such a policy can generate the initial transit networks for metaheuristic improvement methods to then optimize, and that this improves the quality of the final networks found by those methods, in comparison with the heuristic algorithms typically used to find initial solutions.  And in \autoref{chap:neighbourhood_moves}, we integrated the policy more tightly with an improvement method - specifically, an evolutionary algorithm - by using it to propose neighbourhood moves at each step of improvement, finding that this improved the quality of networks found versus improvement methods that used unlearned heuristics to select neighbourhood moves.  We refer to this hybrid method as a Neural-Evolutionary Algorithm (NEA).  Finally, in \autoref{chap:real_world}, we applied our combination of neural policy and evolutionary algorithm to a large real-world problem instance, and showed that it was able to improve on the existing transit on multiple metrics.

\section{Future Work}

In conducting this research, many more directions presented themselves to us than we were able to pursue.  We present some of these here, in the hopes that we will inspire some of our readers to pursue them in our stead.

\subsection{Learned Heuristics in Other Metaheuristic Algorithms}
The evolutionary algorithm in which we use our learned heuristic was chosen for its speed and simplicity, which aided in implementation and rapid experimentation.  But state-of-the-art algorithms, like NSGA-II, which are more computationally costly but more powerful, may allow still better performance to be gained from learned heuristics.  Future work should attempt to use learned policies as low-level heuristics in a wider variety of metaheuristic algorithms.  

It would also be interesting to use multiple different learned policies, perhaps trained under different conditions such as over certain limited ranges of $\alpha$, on different values of $n$ or $S$, or on training data generated by different processes, as different low-level heuristics within the same metaheuristic algorithm.  This would coordinate a diverse set of policies with potentially diverse heuristics for the \gls{ndp} to solve a given problem instance, in a way similar to ensemble methods, which are common in applications of machine learning.

\subsection{Training Policies Differently, and Training Different Policies}
Our method for learning policies has several limitations.  One is that the construction \gls{mdp} on which our neural policies are trained differs from the evolutionary algorithm - an improvement process - in which they is deployed.  Another is that it learns just one type of policy, that for constructing a route from shortest paths.  Natural next steps would be to learn a policies for a more diverse space of actions, perhaps including route-lengthening and -shortening operators, and to train our neural net policies directly in the context of an improvement process, which may yield learned heuristics that are better-suited to that process.

Another promising idea is to learn hyper-heuristics for a metaheuristic algorithm.  A typical hyper-heuristic, as used by \cite{ahmed2019hyperheuristic} and \cite{husselmann2023improved}, dynamically adapts the probability of using each low-level heuristic based on its performance over the search so far.  A neural net policy could be trained to act as a hyper-heuristic to select among low-level heuristics, given details about the scenario and previous steps taken during the algorithm.  This could complement the neural policies we have here used as low-level heuristics.


\subsection{Interactions with Mobility-On-Demand}
As discussed in \autoref{subsec:motivation}, mobility-on-demand services and self-driving automobiles on their own are more likely to worsen urban mobility than to improve it.  But they may still be beneficial if used as one component of a mass-transit-centric system for urban transport.  A fully-autonomous urban transit system will need to make use of both point-to-point autonomous ride-sharing and fixed autonomous mass transit routes to be as useful and efficient as possible~\citep{alonso2018potential, leich2019should}.  There is an existing literature on the planning of such \gls{amod} systems, in which the main problem is planning where empty autonomous taxis should be sent to anticipate demand as it arises over the course of the day~\citep{pavone2012robotic, spieser2014amodDesign, bischoff2017integrating, vakayil2017integrating, ruch2018amodeus, enders2023hybrid}.

The methods developed in this thesis could serve as part of an algorithm for planning such a combined autonomous transit system.  Future work could explore how our NEA could be combined with techniques from the literature on autonomous mobility-on-demand planning to minimize travel times, operating costs, and congestion.

\subsection{Going From Simulation to Reality}

As discussed in \autoref{sec:laval_results}, in order to apply  in a real-world city like Laval, many additional problems need to be solved.  The routes ou
r NEA proposes are at the level of \acrlong{cda}s, of which there are less than seven hundred, while the actual number of transit stops in Laval is more than two thousand.  To make use of our NEA in reality, one would either have to derive real-stop-level networks from the \gls{cda}-level networks it presently produces, or attempt to apply our NEA directly at the level of real transit stops.  Either of these would be an interesting challenge: the former would require a sensible algorithm be developed, while the latter would require us to extend our NEA to the asymmetric \gls{ndp}, since one-way streets are part of the real street network of Laval.

Another problem that would have to be addressed is developing a schedule for the routes on the network.  When applied to a given transit network, the \acrfull{fsp} is relatively simple, and many techniques for it have been proposed~\citep{guihaire2008transitReview, cederBook}.  But most of these rely on the chunking of time into fixed increments, usually of an hour, with demand assumed to be static within each chunk.  Treating demand as a continuous function of time could allow for better results, as could solving the \gls{fsp} in tandem with the \gls{ndp}.  \cite{darwish2020optimising} and \cite{yoo2023reinforcement} both apply \gls{rl} to solve the \gls{ndp} and \gls{fsp} jointly, though only for the very small Mandl city.  It would be worthwhile to explore ways that our NEA could be extended to do the same.

\section{Concluding Statement}

Public transit is essential service to the health of modern cities, and one that has been too much neglected.  But there is a tension at its heart.  As one researcher put it to me, public transit must take you from where you aren't to where you don't want to go.  To mitigate this drawback enough that city-dwellers will still find transit appealing, sound design of transit networks is needed.  

The motivation for this thesis came from seeing that the methods of machine learning with deep neural nets, which in the past decade have underlain dramatic progress in many fields, had been neglected in the study of this complex design problem.  The work we present here clearly shows the potential of these methods to improve the transit networks that tie together the world's cities.  We hope that the community of researchers studying the \acrlong{ndp} will be inspired by it to further develop data-driven learning methods in this area, and that some day, practitioners may put them to use in reality, to tie the cities of the future more closely together.

{

\bibliography{references}

\begin{thebibliography}{118}
\providecommand{\natexlab}[1]{#1}
\providecommand{\url}[1]{\texttt{#1}}
\expandafter\ifx\csname urlstyle\endcsname\relax
  \providecommand{\doi}[1]{doi: #1}\else
  \providecommand{\doi}{doi: \begingroup \urlstyle{rm}\Url}\fi

\bibitem[Abduljabbar et~al.(2019)Abduljabbar, Dia, Liyanage, and
  Bagloee]{abduljabbar2019applications}
Rusul Abduljabbar, Hussein Dia, Sohani Liyanage, and Saeed~Asadi Bagloee.
\newblock Applications of artificial intelligence in transport: An overview.
\newblock \emph{Sustainability}, 11\penalty0 (1):\penalty0 189, 2019.

\bibitem[{Agence métropolitaine de transport}(2013)]{od_data}
{Agence métropolitaine de transport}.
\newblock Enquête origine-destination 2013, 2013.
\newblock Montreal, QC.

\bibitem[Ahmed et~al.(2019)Ahmed, Mumford, and Kheiri]{ahmed2019hyperheuristic}
Leena Ahmed, Christine Mumford, and Ahmed Kheiri.
\newblock Solving urban transit route design problem using selection
  hyper-heuristics.
\newblock \emph{European Journal of Operational Research}, 274\penalty0
  (2):\penalty0 545--559, 2019.

\bibitem[Ai et~al.(2022)Ai, Zuo, Chen, and Wu]{ai2022deep}
Guanqun Ai, Xingquan Zuo, Gang Chen, and Binglin Wu.
\newblock Deep reinforcement learning based dynamic optimization of bus
  timetable.
\newblock \emph{Applied Soft Computing}, 131:\penalty0 109752, 2022.

\bibitem[Akiba et~al.(2019)Akiba, Sano, Yanase, Ohta, and Koyama]{optuna_2019}
Takuya Akiba, Shotaro Sano, Toshihiko Yanase, Takeru Ohta, and Masanori Koyama.
\newblock Optuna: A next-generation hyperparameter optimization framework.
\newblock In \emph{Proceedings of the 25rd {ACM} {SIGKDD} International
  Conference on Knowledge Discovery and Data Mining}, 2019.

\bibitem[Alonso-Gonz{\'a}lez et~al.(2018)Alonso-Gonz{\'a}lez, Liu, Cats,
  Van~Oort, and Hoogendoorn]{alonso2018potential}
Mar{\'\i}a~J Alonso-Gonz{\'a}lez, Theo Liu, Oded Cats, Niels Van~Oort, and
  Serge Hoogendoorn.
\newblock The potential of demand-responsive transport as a complement to
  public transport: An assessment framework and an empirical evaluation.
\newblock \emph{Transportation Research Record}, 2672\penalty0 (8):\penalty0
  879--889, 2018.

\bibitem[Applegate et~al.(2001)Applegate, Bixby, Chvátal, and
  Cook]{concordeTspSolver}
David Applegate, Robert~E. Bixby, Vašek Chvátal, and William~J. Cook.
\newblock Concorde tsp solver, 2001.
\newblock URL \url{https://www.math.uwaterloo.ca/tsp/concorde/index.html}.

\bibitem[Battaglia et~al.(2018)Battaglia, Hamrick, Bapst, Sanchez-Gonzalez,
  Zambaldi, Malinowski, Tacchetti, Raposo, Santoro, Faulkner,
  et~al.]{battaglia2018relational}
Peter~W Battaglia, Jessica~B Hamrick, Victor Bapst, Alvaro Sanchez-Gonzalez,
  Vinicius Zambaldi, Mateusz Malinowski, Andrea Tacchetti, David Raposo, Adam
  Santoro, Ryan Faulkner, et~al.
\newblock Relational inductive biases, deep learning, and graph networks.
\newblock \emph{arXiv preprint arXiv:1806.01261}, 2018.

\bibitem[Beene(2018)]{bloombergJacksonville}
Ryan Beene.
\newblock A {Florida} monorail makes way for the robot bus of tomorrow.
\newblock \emph{Bloomberg}, 2 2018.

\bibitem[Bellman(1958)]{bellman1958dynamic}
Richard Bellman.
\newblock Dynamic programming and stochastic control processes.
\newblock \emph{Information and control}, 1\penalty0 (3):\penalty0 228--239,
  1958.

\bibitem[Bengio et~al.(2021)Bengio, Lodi, and Prouvost]{bengio2021machine}
Yoshua Bengio, Andrea Lodi, and Antoine Prouvost.
\newblock Machine learning for combinatorial optimization: a methodological
  tour d’horizon.
\newblock \emph{European Journal of Operational Research}, 290\penalty0
  (2):\penalty0 405--421, 2021.

\bibitem[Bergstra et~al.(2013)Bergstra, Yamins, and Cox]{bergstra2013making}
James Bergstra, Daniel Yamins, and David Cox.
\newblock Making a science of model search: Hyperparameter optimization in
  hundreds of dimensions for vision architectures.
\newblock In \emph{International conference on machine learning}, pages
  115--123. PMLR, 2013.

\bibitem[Bischoff et~al.(2017)Bischoff, Kaddoura, Maciejewski, and
  Nagel]{bischoff2017integrating}
Joschka Bischoff, Ihab Kaddoura, Michal Maciejewski, and Kai Nagel.
\newblock Re-defining the role of public transport in a world of shared
  autonomous vehicles.
\newblock In \emph{Symposium of the European Association for Research in
  Transportation (hEART)}, 2017.

\bibitem[Brody et~al.(2021)Brody, Alon, and Yahav]{gatv2conv}
Shaked Brody, Uri Alon, and Eran Yahav.
\newblock How attentive are graph attention networks?, 2021.
\newblock URL \url{https://arxiv.org/abs/2105.14491}.

\bibitem[Bruna et~al.(2013)Bruna, Zaremba, Szlam, and LeCun]{bruna2013spectral}
Joan Bruna, Wojciech Zaremba, Arthur Szlam, and Yann LeCun.
\newblock Spectral networks and locally connected networks on graphs.
\newblock \emph{arXiv preprint arXiv:1312.6203}, 2013.

\bibitem[Caro(1974)]{powerbroker}
Robert~A. Caro.
\newblock \emph{The Power Broker}.
\newblock Random House, New York, 1974.

\bibitem[Ceder(2016)]{cederBook}
Avishai Ceder.
\newblock \emph{Public Transit Planning and Operation: Modeling, Practice and
  Behaviour}.
\newblock CRC Press, 2016.

\bibitem[Chakrabarti(2017)]{Chakrabarti2017getPeopleOutOfTheirCars}
Sandip Chakrabarti.
\newblock How can public transit get people out of their cars? {An} analysis of
  transit mode choice for commute trips in los angeles.
\newblock \emph{Transport Policy}, 54\penalty0 (October 2016):\penalty0 80--89,
  2017.
\newblock ISSN 1879310X.
\newblock \doi{10.1016/j.tranpol.2016.11.005}.
\newblock URL \url{http://dx.doi.org/10.1016/j.tranpol.2016.11.005}.

\bibitem[Chen and Tian(2019)]{chen2019learning}
Xinyun Chen and Yuandong Tian.
\newblock Learning to perform local rewriting for combinatorial optimization.
\newblock \emph{Advances in Neural Information Processing Systems}, 32, 2019.

\bibitem[Chien et~al.(2002)Chien, Ding, and Wei]{chien2002dynamic}
Steven I-Jy Chien, Yuqing Ding, and Chienhung Wei.
\newblock Dynamic bus arrival time prediction with artificial neural networks.
\newblock \emph{Journal of transportation engineering}, 128\penalty0
  (5):\penalty0 429--438, 2002.

\bibitem[Choo et~al.(2022)Choo, Kwon, Kim, Jae, Hottung, Tierney, and
  Gwon]{choo2022simulation}
Jinho Choo, Yeong-Dae Kwon, Jihoon Kim, Jeongwoo Jae, Andr{\'e} Hottung, Kevin
  Tierney, and Youngjune Gwon.
\newblock Simulation-guided beam search for neural combinatorial optimization.
\newblock \emph{Advances in Neural Information Processing Systems},
  35:\penalty0 8760--8772, 2022.

\bibitem[Christophides(1976)]{christophides1976worst}
N~Christophides.
\newblock Worst-case analysis of a new heuristic for the traveling salesman
  problem.
\newblock In \emph{Proc. Symposium on New Directions and Recent Results in
  Algorithms and Complexity}, 1976.

\bibitem[da~Costa et~al.(2020)da~Costa, Rhuggenaath, Zhang, and
  Akcay]{d2020learning}
Paulo da~Costa, Jason Rhuggenaath, Yingqian Zhang, and Alp Akcay.
\newblock Learning 2-opt heuristics for the traveling salesman problem via deep
  reinforcement learning.
\newblock In \emph{Asian conference on machine learning}, pages 465--480. PMLR,
  2020.

\bibitem[Dai et~al.(2017)Dai, Khalil, Zhang, Dilkina, and
  Song]{dai2017learningCombinatorial}
Hanjun Dai, Elias~B Khalil, Yuyu Zhang, Bistra Dilkina, and Le~Song.
\newblock Learning combinatorial optimization algorithms over graphs.
\newblock \emph{arXiv preprint arXiv:1704.01665}, 2017.

\bibitem[Darwish et~al.(2020)Darwish, Khalil, and
  Badawi]{darwish2020optimising}
Ahmed Darwish, Momen Khalil, and Karim Badawi.
\newblock Optimising public bus transit networks using deep reinforcement
  learning.
\newblock In \emph{2020 IEEE 23rd International Conference on Intelligent
  Transportation Systems (ITSC)}, pages 1--7. IEEE, 2020.

\bibitem[DeClerq(2020)]{ctv2020congestion}
Katherine DeClerq.
\newblock Drivers in toronto lose 142 hours on the roads during rush hour,
  report finds.
\newblock
  \url{https://toronto.ctvnews.ca/drivers-in-toronto-lose-142-hours-on-the-roads-during-rush-hour-report-finds-1.4790478},
  1 2020.
\newblock Accessed: 2024-09-27.

\bibitem[Defferrard et~al.(2016)Defferrard, Bresson, and
  Vandergheynst]{defferrard2016spectral}
Micha{\"{e}}l Defferrard, Xavier Bresson, and Pierre Vandergheynst.
\newblock Convolutional neural networks on graphs with fast localized spectral
  filtering.
\newblock \emph{CoRR}, abs/1606.09375, 2016.
\newblock URL \url{http://arxiv.org/abs/1606.09375}.

\bibitem[Dur{\'a}n-Micco and Vansteenwegen(2022)]{duran2022survey}
Javier Dur{\'a}n-Micco and Pieter Vansteenwegen.
\newblock A survey on the transit network design and frequency setting problem.
\newblock \emph{Public Transport}, 14\penalty0 (1):\penalty0 155--190, 2022.

\bibitem[Duranton and Turner(2011)]{duranton2011fundamental}
Gilles Duranton and Matthew~A Turner.
\newblock The fundamental law of road congestion: Evidence from us cities.
\newblock \emph{American Economic Review}, 101\penalty0 (6):\penalty0 2616--52,
  2011.

\bibitem[Duvenaud et~al.(2015)Duvenaud, Maclaurin, Iparraguirre, Bombarell,
  Hirzel, Aspuru-Guzik, and Adams]{duvenaud2015convolutional}
David~K Duvenaud, Dougal Maclaurin, Jorge Iparraguirre, Rafael Bombarell,
  Timothy Hirzel, Al{\'a}n Aspuru-Guzik, and Ryan~P Adams.
\newblock Convolutional networks on graphs for learning molecular fingerprints.
\newblock \emph{Advances in neural information processing systems}, 28, 2015.

\bibitem[Enders et~al.(2023)Enders, Harrison, Pavone, and
  Schiffer]{enders2023hybrid}
Tobias Enders, James Harrison, Marco Pavone, and Maximilian Schiffer.
\newblock Hybrid multi-agent deep reinforcement learning for autonomous
  mobility on demand systems.
\newblock In \emph{Learning for Dynamics and Control Conference}, pages
  1284--1296. PMLR, 2023.

\bibitem[Fan and Mumford(2010)]{fan2010metaheuristic}
Lang Fan and Christine~L Mumford.
\newblock A metaheuristic approach to the urban transit routing problem.
\newblock \emph{Journal of Heuristics}, 16:\penalty0 353--372, 2010.

\bibitem[Fortune(1995)]{fortune1995voronoi}
Steven Fortune.
\newblock Voronoi diagrams and delaunay triangulations.
\newblock \emph{Computing in Euclidean geometry}, pages 225--265, 1995.

\bibitem[France-Presse(2021)]{malagaAutonomousBus}
Agence France-Presse.
\newblock Driverless electric bus hits the road in {Spanish} city of
  {M{\'a}laga}.
\newblock
  \url{https://www.theguardian.com/world/2021/feb/25/driverless-electric-bus-hits-the-road-in-spanish-city-of-malaga},
  2 2021.
\newblock Accessed: 2021-03-09.

\bibitem[Frank(2020)]{selfDrivingReducesCongestion}
B~Alex Frank.
\newblock How self-driving cars will reduce traffic congestion.
\newblock
  \url{https://medium.com/swlh/how-self-driving-cars-will-reduce-traffic-congestion-8bad5594c5d0},
  1 2020.
\newblock Accessed: 2021-03-26.

\bibitem[Freeman(2001)]{freeman2001effects}
Lance Freeman.
\newblock The effects of sprawl on neighborhood social ties: An explanatory
  analysis.
\newblock \emph{Journal of the American Planning Association}, 67\penalty0
  (1):\penalty0 69--77, 2001.

\bibitem[Fu et~al.(2021)Fu, Qiu, and Zha]{fu2021generalize}
Zhang-Hua Fu, Kai-Bin Qiu, and Hongyuan Zha.
\newblock Generalize a small pre-trained model to arbitrarily large tsp
  instances.
\newblock In \emph{Proceedings of the AAAI conference on artificial
  intelligence}, volume~35, pages 7474--7482, 2021.

\bibitem[Fukushima(1969)]{fukushima1969visual}
Kunihiko Fukushima.
\newblock Visual feature extraction by a multilayered network of analog
  threshold elements.
\newblock \emph{IEEE Transactions on Systems Science and Cybernetics},
  5\penalty0 (4):\penalty0 322--333, 1969.

\bibitem[Gilmer et~al.(2017)Gilmer, Schoenholz, Riley, Vinyals, and
  Dahl]{gilmer2017quantum}
Justin Gilmer, Samuel~S. Schoenholz, Patrick~F. Riley, Oriol Vinyals, and
  George~E. Dahl.
\newblock Neural message passing for quantum chemistry.
\newblock In Doina Precup and Yee~Whye Teh, editors, \emph{Proceedings of the
  34th International Conference on Machine Learning}, volume~70 of
  \emph{Proceedings of Machine Learning Research}, pages 1263--1272. PMLR,
  06--11 Aug 2017.
\newblock URL \url{https://proceedings.mlr.press/v70/gilmer17a.html}.

\bibitem[Goodfellow et~al.(2016)Goodfellow, Bengio, and
  Courville]{goodfellow2016}
Ian Goodfellow, Yoshua Bengio, and Aaron Courville.
\newblock \emph{Deep Learning}.
\newblock MIT Press, 2016.
\newblock \url{http://www.deeplearningbook.org}.

\bibitem[Google(2024)]{googleMapsLaval}
Google.
\newblock Laval, quebec.
\newblock
  https://www.google.com/maps/place/Laval,+QC/@45.6100977,-73.8063121,11.61z,
  2024.
\newblock Online; accessed 12 April 2024.

\bibitem[Guan et~al.(2006)Guan, Yang, and
  Wirasinghe]{guan2006AnalyticRoutePlanning}
J.F. Guan, Hai Yang, and S.C. Wirasinghe.
\newblock Simultaneous optimization of transit line configuration and passenger
  line assignment.
\newblock \emph{Transportation Research Part B: Methodological}, 40:\penalty0
  885--902, 12 2006.
\newblock \doi{10.1016/j.trb.2005.12.003}.

\bibitem[Guihaire and Hao(2008)]{guihaire2008transitReview}
Val{\'e}rie Guihaire and Jin-Kao Hao.
\newblock Transit network design and scheduling: A global review.
\newblock \emph{Transportation Research Part A: Policy and Practice},
  42\penalty0 (10):\penalty0 1251--1273, 2008.

\bibitem[Harb et~al.(2018)Harb, Xiao, Circella, Mokhtarian, and
  Walker]{harb2018projecting}
Mustapha Harb, Yu~Xiao, Giovanni Circella, Patricia~L Mokhtarian, and Joan~L
  Walker.
\newblock Projecting travelers into a world of self-driving vehicles:
  estimating travel behavior implications via a naturalistic experiment.
\newblock \emph{Transportation}, 45\penalty0 (6):\penalty0 1671--1685, 2018.

\bibitem[Hendrycks and Gimpel(2016)]{hendrycks2016gaussian}
Dan Hendrycks and Kevin Gimpel.
\newblock Gaussian error linear units (gelus).
\newblock \emph{arXiv preprint arXiv:1606.08415}, 2016.

\bibitem[Holliday and Dudek(2023)]{holliday2023augmenting}
Andrew Holliday and Gregory Dudek.
\newblock Augmenting transit network design algorithms with deep learning.
\newblock In \emph{2023 26th IEEE International Conference on Intelligent
  Transportation Systems (ITSC)}. IEEE, 2023.

\bibitem[Holliday and Dudek(2024)]{holliday2024autonomous}
Andrew Holliday and Gregory Dudek.
\newblock A neural-evolutionary algorithm for autonomous transit network
  design.
\newblock In \emph{presented at 2024 IEEE International Conference on Robotics
  and Automation (ICRA)}. IEEE, 2024.

\bibitem[Hottung and Tierney(2019)]{hottung2019neural}
Andr{\'e} Hottung and Kevin Tierney.
\newblock Neural large neighborhood search for the capacitated vehicle routing
  problem.
\newblock \emph{arXiv preprint arXiv:1911.09539}, 2019.

\bibitem[H{\"u}sselmann et~al.(2023)H{\"u}sselmann, van Vuuren, and
  Andersen]{husselmann2023improved}
G{\"u}nther H{\"u}sselmann, Jan~Harm van Vuuren, and Simen~Johann Andersen.
\newblock An improved solution methodology for the urban transit routing
  problem.
\newblock \emph{Computers \& Operations Research}, page 106481, 2023.

\bibitem[Islam et~al.(2019)Islam, Moosa, Mobin, Nayeem, and
  Rahman]{islam2019heuristic}
Kazi~Ashik Islam, Ibraheem~Muhammad Moosa, Jaiaid Mobin, Muhammad~Ali Nayeem,
  and M~Sohel Rahman.
\newblock A heuristic aided stochastic beam search algorithm for solving the
  transit network design problem.
\newblock \emph{Swarm and Evolutionary Computation}, 46:\penalty0 154--170,
  2019.

\bibitem[Jeong and Rilett(2004)]{jeong2004bus}
Ranhee Jeong and R~Rilett.
\newblock Bus arrival time prediction using artificial neural network model.
\newblock In \emph{Proceedings. The 7th international IEEE conference on
  intelligent transportation systems (IEEE Cat. No. 04TH8749)}, pages 988--993.
  IEEE, 2004.

\bibitem[Jiang et~al.(2018)Jiang, Fan, Liu, Zhu, and
  Gu]{jiang2018passengerInflow}
Zhibin Jiang, Wei Fan, Wei Liu, Bingqin Zhu, and Jinjing Gu.
\newblock Reinforcement learning approach for coordinated passenger inflow
  control of urban rail transit in peak hours.
\newblock \emph{Transportation Research Part C: Emerging Technologies},
  88:\penalty0 1--16, 2018.
\newblock ISSN 0968-090X.
\newblock \doi{https://doi.org/10.1016/j.trc.2018.01.008}.
\newblock URL
  \url{https://www.sciencedirect.com/science/article/pii/S0968090X18300111}.

\bibitem[John(2016)]{john2016thesis}
Matthew~P John.
\newblock \emph{Metaheuristics for designing efficient routes \& schedules for
  urban transportation networks}.
\newblock PhD thesis, Cardiff University, 2016.

\bibitem[John et~al.(2014)John, Mumford, and Lewis]{john2014routing}
Matthew~P. John, Christine~L. Mumford, and Rhyd Lewis.
\newblock An improved multi-objective algorithm for the urban transit routing
  problem.
\newblock In Christian Blum and Gabriela Ochoa, editors, \emph{Evolutionary
  Computation in Combinatorial Optimisation}, pages 49--60, Berlin, Heidelberg,
  2014. Springer Berlin Heidelberg.
\newblock ISBN 978-3-662-44320-0.

\bibitem[Kar et~al.(2022)Kar, Carrel, Miller, and Le]{kar2022public}
Armita Kar, Andre~L Carrel, Harvey~J Miller, and Huyen~TK Le.
\newblock Public transit cuts during covid-19 compound social vulnerability in
  22 us cities.
\newblock \emph{Transportation Research Part D: Transport and Environment},
  110:\penalty0 103435, 2022.

\bibitem[Kautz et~al.(1992)Kautz, Selman, et~al.]{kautz1992planning}
Henry~A Kautz, Bart Selman, et~al.
\newblock Planning as satisfiability.
\newblock In \emph{ECAI}, volume~92, pages 359--363. Citeseer, 1992.

\bibitem[Kepaptsoglou and Karlaftis(2009)]{kepaptsoglou2009transitReview}
Konstantinos Kepaptsoglou and Matthew Karlaftis.
\newblock Transit route network design problem: Review.
\newblock \emph{Journal of Transportation Engineering}, 135\penalty0
  (8):\penalty0 491--505, 2009.
\newblock \doi{10.1061/(ASCE)0733-947X(2009)135:8(491)}.

\bibitem[Kim et~al.(2021)Kim, Park, et~al.]{kim2021learning}
Minsu Kim, Jinkyoo Park, et~al.
\newblock Learning collaborative policies to solve np-hard routing problems.
\newblock \emph{Advances in Neural Information Processing Systems},
  34:\penalty0 10418--10430, 2021.

\bibitem[Kingma and Ba(2015)]{kingma2015adam}
Diederik~P. Kingma and Jimmy Ba.
\newblock Adam: A method for stochastic optimization.
\newblock In \emph{ICLR}, 2015.

\bibitem[Kipf and Welling(2016)]{kipf2016semi}
Thomas~N Kipf and Max Welling.
\newblock Semi-supervised classification with graph convolutional networks.
\newblock \emph{arXiv preprint arXiv:1609.02907}, 2016.

\bibitem[Kool et~al.(2019)Kool, Hoof, and Welling]{Kool2019AttentionLT}
W.~Kool, H.~V. Hoof, and M.~Welling.
\newblock Attention, learn to solve routing problems!
\newblock In \emph{ICLR}, 2019.

\bibitem[Kılıç and Gök(2014)]{kilic2014demand}
Fatih Kılıç and Mustafa Gök.
\newblock A demand based route generation algorithm for public transit network
  design.
\newblock \emph{Computers \& Operations Research}, 51:\penalty0 21--29, 2014.
\newblock ISSN 0305-0548.
\newblock \doi{https://doi.org/10.1016/j.cor.2014.05.001}.
\newblock URL
  \url{https://www.sciencedirect.com/science/article/pii/S0305054814001300}.

\bibitem[{Le Corbusier}(1929)]{leCorbusierCityOfTomorrow}
{Le Corbusier}.
\newblock \emph{The City of Tomorrow and Its Planning}.
\newblock Dover Publications, 1929.
\newblock English translation published 1987.

\bibitem[Leich and Bischoff(2019)]{leich2019should}
Gregor Leich and Joschka Bischoff.
\newblock Should autonomous shared taxis replace buses? a simulation study.
\newblock \emph{Transportation Research Procedia}, 41:\penalty0 450--460, 2019.

\bibitem[Leyden(2003)]{leyden2003social}
Kevin~M Leyden.
\newblock Social capital and the built environment: the importance of walkable
  neighborhoods.
\newblock \emph{American journal of public health}, 93\penalty0 (9):\penalty0
  1546--1551, 2003.

\bibitem[Li et~al.(2020)Li, Bai, Liu, Yao, and Waller]{li2020graph}
Can Li, Lei Bai, Wei Liu, Lina Yao, and S~Travis Waller.
\newblock Graph neural network for robust public transit demand prediction.
\newblock \emph{IEEE Transactions on Intelligent Transportation Systems}, 2020.

\bibitem[Lin and Tang(2022)]{lin2022analysis}
Haifeng Lin and Chengpei Tang.
\newblock Analysis and optimization of urban public transport lines based on
  multiobjective adaptive particle swarm optimization.
\newblock \emph{IEEE Transactions on Intelligent Transportation Systems},
  23\penalty0 (9):\penalty0 16786--16798, 2022.
\newblock \doi{10.1109/TITS.2021.3086808}.

\bibitem[Litman(2012)]{litman2012evaluating}
Todd Litman.
\newblock \emph{Evaluating public transportation health benefits}.
\newblock Victoria Transport Policy Institute, 2012.

\bibitem[Liu et~al.(2020)Liu, Miller, and Scheff]{liu2020impacts}
Luyu Liu, Harvey~J Miller, and Jonathan Scheff.
\newblock The impacts of covid-19 pandemic on public transit demand in the
  united states.
\newblock \emph{Plos one}, 15\penalty0 (11):\penalty0 e0242476, 2020.

\bibitem[Ma et~al.(2021)Ma, Li, Cao, Song, Zhang, Chen, and
  Tang]{ma2021learning}
Yining Ma, Jingwen Li, Zhiguang Cao, Wen Song, Le~Zhang, Zhenghua Chen, and
  Jing Tang.
\newblock Learning to iteratively solve routing problems with dual-aspect
  collaborative transformer.
\newblock \emph{Advances in Neural Information Processing Systems},
  34:\penalty0 11096--11107, 2021.

\bibitem[Madden(2017)]{selfDrivingCouldSupplementPublicTransit}
Jenifer~Joy Madden.
\newblock Self-driving vehicles will improve our cities, if they don't ruin
  them.
\newblock
  \url{https://ggwash.org/view/63147/self-driving-vehicles-will-improve-our-cities-if-they-dont-ruin-them},
  5 2017.

\bibitem[Mandl(1980)]{mandl1980evaluation}
Christoph~E Mandl.
\newblock Evaluation and optimization of urban public transportation networks.
\newblock \emph{European Journal of Operational Research}, 5\penalty0
  (6):\penalty0 396--404, 1980.

\bibitem[Mirhoseini et~al.(2021)Mirhoseini, Goldie, Yazgan, Jiang, Songhori,
  Wang, Lee, Johnson, Pathak, Nazi, et~al.]{mirhoseini2021graph}
Azalia Mirhoseini, Anna Goldie, Mustafa Yazgan, Joe~Wenjie Jiang, Ebrahim
  Songhori, Shen Wang, Young-Joon Lee, Eric Johnson, Omkar Pathak, Azade Nazi,
  et~al.
\newblock A graph placement methodology for fast chip design.
\newblock \emph{Nature}, 594\penalty0 (7862):\penalty0 207--212, 2021.

\bibitem[Montgomery(2013)]{happycity}
Charles Montgomery.
\newblock \emph{Happy City}.
\newblock Doubleday Canada, New York, 2013.

\bibitem[Mumford(2004)]{mumford2004simple}
Christine~L Mumford.
\newblock Simple population replacement strategies for a steady-state
  multi-objective evolutionary algorithm.
\newblock In \emph{Genetic and Evolutionary Computation Conference}, pages
  1389--1400. Springer, 2004.

\bibitem[Mumford(2013{\natexlab{a}})]{mumford2013dataset}
Christine~L Mumford.
\newblock Supplementary material for: New heuristic and evolutionary operators
  for the multi-objective urban transit routing problem, cec 2013.
\newblock
  \url{https://users.cs.cf.ac.uk/C.L.Mumford/Research\%20Topics/UTRP/CEC2013Supp.zip},
  2013{\natexlab{a}}.
\newblock Accessed: 2023-03-24.

\bibitem[Mumford(2013{\natexlab{b}})]{mumford2013new}
Christine~L Mumford.
\newblock New heuristic and evolutionary operators for the multi-objective
  urban transit routing problem.
\newblock In \emph{2013 IEEE congress on evolutionary computation}, pages
  939--946. IEEE, 2013{\natexlab{b}}.

\bibitem[Mundhenk et~al.(2021)Mundhenk, Landajuela, Glatt, Santiago, Faissol,
  and Petersen]{mundhenk2021symbolic}
T~Nathan Mundhenk, Mikel Landajuela, Ruben Glatt, Claudio~P Santiago, Daniel~M
  Faissol, and Brenden~K Petersen.
\newblock Symbolic regression via neural-guided genetic programming population
  seeding.
\newblock \emph{arXiv preprint arXiv:2111.00053}, 2021.

\bibitem[Nikoli{\'c} and Teodorovi{\'c}(2013)]{nikolic2013transit}
Milo{\v{s}} Nikoli{\'c} and Du{\v{s}}an Teodorovi{\'c}.
\newblock Transit network design by bee colony optimization.
\newblock \emph{Expert Systems with Applications}, 40\penalty0 (15):\penalty0
  5945--5955, 2013.

\bibitem[Oh et~al.(2020)Oh, Seshadri, Le, Zegras, and
  Ben-Akiva]{oh2020evaluating}
Simon Oh, Ravi Seshadri, Diem-Trinh Le, P~Christopher Zegras, and Moshe~E
  Ben-Akiva.
\newblock Evaluating automated demand responsive transit using microsimulation.
\newblock \emph{IEEE Access}, 8:\penalty0 82551--82561, 2020.

\bibitem[Pavone et~al.(2012)Pavone, Smith, Frazzoli, and
  Rus]{pavone2012robotic}
Marco Pavone, Stephen~L Smith, Emilio Frazzoli, and Daniela Rus.
\newblock Robotic load balancing for mobility-on-demand systems.
\newblock \emph{The International Journal of Robotics Research}, 31\penalty0
  (7):\penalty0 839--854, 2012.

\bibitem[Plautz(2018)]{smartcitiesdriveDetroit}
Jason Plautz.
\newblock Autonomous shuttles launch in {Detroit}.
\newblock
  \url{https://www.smartcitiesdive.com/news/autonomous-shuttles-launch-in-detroit/526999/},
  7 2018.
\newblock Accessed: 2019-11-27.

\bibitem[Rich et~al.(2023)Rich, Seshadri, Jomeh, and Clausen]{rich2023fixed}
Jeppe Rich, Ravi Seshadri, Ali~Jamal Jomeh, and Sofus~Rasmus Clausen.
\newblock Fixed routing or demand-responsive? agent-based modelling of
  autonomous first and last mile services in light-rail systems.
\newblock \emph{Transportation Research Part A: Policy and Practice},
  173:\penalty0 103676, 2023.

\bibitem[Rodrigue(1997)]{rodrigue1997NNsForLandUseAndTransport}
Jean-Paul Rodrigue.
\newblock Parallel modelling and neural networks: An overview for
  transportation/land use systems.
\newblock \emph{Transportation Research Part C: Emerging Technologies},
  5\penalty0 (5):\penalty0 259--271, 1997.
\newblock ISSN 0968-090X.
\newblock \doi{https://doi.org/10.1016/S0968-090X(97)00014-4}.
\newblock URL
  \url{https://www.sciencedirect.com/science/article/pii/S0968090X97000144}.

\bibitem[Roughgarden and Tardos(2002)]{roughgarden2002bad}
Tim Roughgarden and {\'E}va Tardos.
\newblock How bad is selfish routing?
\newblock \emph{Journal of the ACM (JACM)}, 49\penalty0 (2):\penalty0 236--259,
  2002.

\bibitem[Ruch et~al.(2018)Ruch, H{\"o}rl, and Frazzoli]{ruch2018amodeus}
Claudio Ruch, Sebastian H{\"o}rl, and Emilio Frazzoli.
\newblock Amodeus, a simulation-based testbed for autonomous mobility-on-demand
  systems.
\newblock In \emph{2018 21st International Conference on Intelligent
  Transportation Systems (ITSC)}, pages 3639--3644. IEEE, 2018.

\bibitem[Sandow(2011)]{sandow2011road}
Erika Sandow.
\newblock \emph{On the road: Social aspects of commuting long distances to
  work}.
\newblock PhD thesis, Kulturgeografiska institutionen, Ume{\aa} universitet,
  2011.

\bibitem[{Schaller Consulting}(2018)]{SchallerConsulting2018}
{Schaller Consulting}.
\newblock The new automobility: {Lyft}, {Uber} and the future of {American}
  cities.
\newblock \emph{Journal of Urban Economics}, 2018.
\newblock URL
  \url{www.schallerconsult.com/rideservices/automobility.pdf{\%}0Ahttp://www.schallerconsult.com/rideservices/automobility.htm}.

\bibitem[Schulman et~al.(2017)Schulman, Wolski, Dhariwal, Radford, and
  Klimov]{schulman2017PPO}
John Schulman, Filip Wolski, Prafulla Dhariwal, Alec Radford, and Oleg Klimov.
\newblock Proximal policy optimization algorithms.
\newblock \emph{arXiv preprint arXiv:1707.06347}, 2017.

\bibitem[Shrikant(2018)]{spielerInterview2018}
Aditi Shrikant.
\newblock The best and worst cities in america for public transportation,
  according to an urban planner.
\newblock
  \url{https://www.vox.com/the-goods/2018/12/7/18131132/public-transportation-bus-subway-america-us},
  2018.
\newblock Accessed: 2024-09-25.

\bibitem[{Soci\'et\'e de Transport de Laval}()]{laval_gtfs}
{Soci\'et\'e de Transport de Laval}.
\newblock {GTFS data for the Soci\'et\'e de Transport de Laval bus network}.
\newblock URL
  \url{https://transitfeeds.com/p/societe-de-transport-de-laval/38/1383528159}.
\newblock Retreived 2020.

\bibitem[S{\"o}rensen(2015)]{sorensen2015metaheuristics}
Kenneth S{\"o}rensen.
\newblock Metaheuristics—the metaphor exposed.
\newblock \emph{International Transactions in Operational Research},
  22\penalty0 (1):\penalty0 3--18, 2015.

\bibitem[S{\"o}rensen et~al.(2018)S{\"o}rensen, Sevaux, and
  Glover]{sorensen2018history}
Kenneth S{\"o}rensen, Marc Sevaux, and Fred Glover.
\newblock A history of metaheuristics.
\newblock In \emph{Handbook of heuristics}, pages 791--808. Springer, 2018.

\bibitem[Spieser et~al.(2014)Spieser, Treleaven, Zhang, Frazzoli, Morton, and
  Pavone]{spieser2014amodDesign}
Kevin Spieser, Kyle Treleaven, Rick Zhang, Emilio Frazzoli, Daniel Morton, and
  Marco Pavone.
\newblock Toward a systematic approach to the design and evaluation of
  automated mobility-on-demand systems: A case study in singapore.
\newblock In \emph{Road vehicle automation}, pages 229--245. Springer, 2014.

\bibitem[{Statistics Canada}(2021)]{disseminationAreas}
{Statistics Canada}.
\newblock Census dissemination area boundary files, 2021.
\newblock URL \url{https://www150.statcan.gc.ca/n1/en/catalogue/92-169-X}.
\newblock Accessed: 2023-07-01.

\bibitem[{Statistics Canada}(2023)]{canadaCensus2021}
{Statistics Canada}.
\newblock Census profile, 2021 census of population, 2023.
\newblock URL
  \url{https://www12.statcan.gc.ca/census-recensement/2021/dp-pd/prof/details/page.cfm}.
\newblock Accessed: 2023-10-25.

\bibitem[Stieglitz(1939)]{moses_image}
C.M. Stieglitz.
\newblock Robert {M}oses with {B}attery {B}ridge model, 1939.
\newblock URL \url{http://hdl.loc.gov/loc.pnp/cph.3c36079}.

\bibitem[Sutton and Barto(2018)]{sutton2018reinforcement}
Richard~S Sutton and Andrew~G Barto.
\newblock \emph{Reinforcement learning: An introduction}.
\newblock MIT press, 2018.

\bibitem[Sykora et~al.(2020)Sykora, Ren, and Urtasun]{sykora2020multi}
Quinlan Sykora, Mengye Ren, and Raquel Urtasun.
\newblock Multi-agent routing value iteration network.
\newblock In \emph{International Conference on Machine Learning}, pages
  9300--9310. PMLR, 2020.

\bibitem[Vakayil et~al.(2017)Vakayil, Gruel, and
  Samaranayake]{vakayil2017integrating}
Akhil Vakayil, Wolfgang Gruel, and Samitha Samaranayake.
\newblock Integrating shared-vehicle mobility-on-demand systems with public
  transit.
\newblock Technical report, 2017.

\bibitem[Valenzuela(2002)]{valenzuela2002simple}
Christine~L Valenzuela.
\newblock A simple evolutionary algorithm for multi-objective optimization
  (seamo).
\newblock In \emph{Proceedings of the 2002 Congress on Evolutionary
  Computation. CEC'02 (Cat. No. 02TH8600)}, volume~1, pages 717--722. IEEE,
  2002.

\bibitem[van Bilsen(1964)]{corbusier_image}
Joop van Bilsen.
\newblock Le {C}orbusier, 1964.
\newblock URL
  \url{http://proxy.handle.net/10648/aa7b03b0-d0b4-102d-bcf8-003048976d84}.

\bibitem[van Nes(2003)]{vannes2003AnalyticRouteAndSchedule}
Rob van Nes.
\newblock Multiuser-class urban transit network design.
\newblock \emph{Transportation Research Record}, 1835\penalty0 (1):\penalty0
  25--33, 2003.
\newblock \doi{10.3141/1835-04}.
\newblock URL \url{https://doi.org/10.3141/1835-04}.

\bibitem[Vaswani et~al.(2017)Vaswani, Shazeer, Parmar, Uszkoreit, Jones, Gomez,
  Kaiser, and Polosukhin]{vaswani2017attention}
Ashish Vaswani, Noam Shazeer, Niki Parmar, Jakob Uszkoreit, Llion Jones,
  Aidan~N Gomez, {\L}ukasz Kaiser, and Illia Polosukhin.
\newblock Attention is all you need.
\newblock In \emph{Advances in neural information processing systems}, pages
  5998--6008, 2017.

\bibitem[Veličković et~al.(2018)Veličković, Cucurull, Casanova, Romero,
  Liò, and Bengio]{velickovic2018graphattentionnetworks}
Petar Veličković, Guillem Cucurull, Arantxa Casanova, Adriana Romero, Pietro
  Liò, and Yoshua Bengio.
\newblock Graph attention networks, 2018.
\newblock URL \url{https://arxiv.org/abs/1710.10903}.

\bibitem[Vinyals et~al.(2015)Vinyals, Fortunato, and
  Jaitly]{vinyals2015pointer}
Oriol Vinyals, Meire Fortunato, and Navdeep Jaitly.
\newblock Pointer networks.
\newblock \emph{arXiv preprint arXiv:1506.03134}, 2015.

\bibitem[Wang et~al.(2024)Wang, Zhu, Zhang, Tian, and Zhang]{wang2024large}
Tao Wang, Zhipeng Zhu, Jing Zhang, Junfang Tian, and Wenyi Zhang.
\newblock A large-scale traffic signal control algorithm based on multi-layer
  graph deep reinforcement learning.
\newblock \emph{Transportation Research Part C: Emerging Technologies},
  162:\penalty0 104582, 2024.
\newblock ISSN 0968-090X.
\newblock \doi{https://doi.org/10.1016/j.trc.2024.104582}.
\newblock URL
  \url{https://www.sciencedirect.com/science/article/pii/S0968090X24001037}.

\bibitem[Williams(1992)]{williams1992reinforce}
Ronald~J Williams.
\newblock Simple statistical gradient-following algorithms for connectionist
  reinforcement learning.
\newblock \emph{Machine learning}, 8\penalty0 (3):\penalty0 229--256, 1992.

\bibitem[Wu et~al.(2021)Wu, Song, Cao, Zhang, and Lim]{wu2021learning}
Yaoxin Wu, Wen Song, Zhiguang Cao, Jie Zhang, and Andrew Lim.
\newblock Learning improvement heuristics for solving routing problems.
\newblock \emph{IEEE transactions on neural networks and learning systems},
  33\penalty0 (9):\penalty0 5057--5069, 2021.

\bibitem[Xiong and Schneider(1992)]{xiong1992transportation}
Yihua Xiong and Jerry~B Schneider.
\newblock Transportation network design using a cumulative genetic algorithm
  and neural network.
\newblock \emph{Transportation Research Record}, 1364, 1992.

\bibitem[Yan et~al.(2023)Yan, Cui, Chen, and Ma]{Yan2023DistributedMD}
Haoyang Yan, Zhiyong Cui, Xinqiang Chen, and Xiaolei Ma.
\newblock Distributed multiagent deep reinforcement learning for multiline
  dynamic bus timetable optimization.
\newblock \emph{IEEE Transactions on Industrial Informatics}, 19:\penalty0
  469--479, 2023.

\bibitem[Yang and Jiang(2020)]{yang2020application}
Jie Yang and Yangsheng Jiang.
\newblock Application of modified nsga-ii to the transit network design
  problem.
\newblock \emph{Journal of Advanced Transportation}, 2020:\penalty0 1--24,
  2020.

\bibitem[Ye et~al.(2024)Ye, Wang, Cao, Liang, and Li]{ye2024deepaco}
Haoran Ye, Jiarui Wang, Zhiguang Cao, Helan Liang, and Yong Li.
\newblock Deepaco: neural-enhanced ant systems for combinatorial optimization.
\newblock \emph{Advances in Neural Information Processing Systems}, 36, 2024.

\bibitem[Yen(1970)]{yen1970algorithm}
Jin~Y Yen.
\newblock An algorithm for finding shortest routes from all source nodes to a
  given destination in general networks.
\newblock \emph{Quarterly of applied mathematics}, 27\penalty0 (4):\penalty0
  526--530, 1970.

\bibitem[Ying et~al.(2018)Ying, He, Chen, Eksombatchai, Hamilton, and
  Leskovec]{ying2018webscale}
Rex Ying, Ruining He, Kaifeng Chen, Pong Eksombatchai, William~L. Hamilton, and
  Jure Leskovec.
\newblock Graph convolutional neural networks for web-scale recommender
  systems.
\newblock \emph{CoRR}, abs/1806.01973, 2018.
\newblock URL \url{http://arxiv.org/abs/1806.01973}.

\bibitem[Yoo et~al.(2023)Yoo, Lee, and Han]{yoo2023reinforcement}
Sunhyung Yoo, Jinwoo~Brian Lee, and Hoon Han.
\newblock A reinforcement learning approach for bus network design and
  frequency setting optimisation.
\newblock \emph{Public Transport}, pages 1--32, 2023.

\bibitem[Zou et~al.(2006)Zou, Xu, and Zhu]{zou2006lightrail}
Liang Zou, Jian-min Xu, and Ling-xiang Zhu.
\newblock Light rail intelligent dispatching system based on reinforcement
  learning.
\newblock In \emph{2006 International Conference on Machine Learning and
  Cybernetics}, pages 2493--2496, 2006.
\newblock \doi{10.1109/ICMLC.2006.258785}.

\bibitem[Çodur and Tortum(2009)]{akgungor2009NNsForAccidentPrediction}
Muhammed~Yasin Çodur and Ahmet Tortum.
\newblock An artificial intelligent approach to traffic accident estimation:
  Model development and application.
\newblock \emph{Transport}, 24\penalty0 (2):\penalty0 135--142, 2009.
\newblock \doi{10.3846/1648-4142.2009.24.135-142}.

\end{thebibliography}
\bibliographystyle{plainnat}

}
	
\end{document}